%% file: nips_main.tex
\documentclass{article}

\PassOptionsToPackage{numbers,compress}{natbib}
\usepackage[main, final]{neurips_2026}

\usepackage[utf8]{inputenc}
\usepackage[T1]{fontenc}
\usepackage{hyperref}
\usepackage{url}
\usepackage{booktabs}
\usepackage{amsfonts}
\usepackage{amsmath}
\usepackage{amssymb}
\usepackage{mathtools}
\usepackage{nicefrac}
\usepackage{censor}
\usepackage{microtype}
\usepackage{xcolor}
\usepackage{graphicx}
\usepackage{subcaption}
\usepackage[inline]{enumitem}
\usepackage{amsthm}
\usepackage{comment}
\newtheorem{proposition}{Proposition}

\newtheorem{lemma}{Lemma}
\newtheorem{corollary}{Corollary}
\theoremstyle{definition}
\newtheorem{definition}{Definition}
\theoremstyle{remark}
\newtheorem{remark}{Remark}
\usepackage{adjustbox}
\usepackage{makecell}
\usepackage{array}
\usepackage{titletoc}
\hypersetup{colorlinks=true,linkcolor=blue,citecolor=blue,urlcolor=blue}
\usepackage{tabularx}

\newcommand{\KL}{\mathrm{KL}}
\newcommand{\MMD}{\mathrm{MMD}}

\newcommand{\E}{\mathbb{E}}
\newcommand{\R}{\mathbb{R}}

\newcommand{\Sink}{\operatorname{Sink}}
\DeclareMathOperator{\dist}{dist}

\title{Understanding Multimodality in Generative Behavioral Cloning}

\author{%
Lorenzo Mazza$^{1,\dagger}$\\
\texttt{lorenzo.mazza@nct-dresden.de}
\And
Massimiliano Datres$^{2,\dagger}$\\
\texttt{datres@math.lmu.de}
\And
Ariel Rodriguez$^{1,3}$
\AND
Sebastian Bodenstedt$^{1,4}$
\And
Gitta Kutyniok$^{2,5,6}$
\And
Stefanie Speidel$^{1,3 , 4,7}$
}

\begin{document}

\maketitle
\begingroup
\renewcommand{\thefootnote}{}
\footnotetext{%
$^\dagger$Corresponding authors. 
$^1$NCT/UCC Dresden, UKDD Dresden, TU Dresden, DKFZ Heidelberg. 
$^2$Ludwig-Maximilians-Universit\"at M\"unchen, Munich Center for Machine Learning (MCML). 
$^3$BMFTR Research Hub 6G-Life. 
$^4$Cluster of Excellence CeTI. 
$^5$University of Troms{\o}. 
$^6$German Aerospace Center (DLR). 
$^7$HZDR Dresden.%
}
\endgroup

\begin{abstract}
Behavioral cloning becomes challenging when the same observation admits several valid actions. We study how generative behavioral-cloning policies represent such multimodal expert behavior and identify different bottlenecks across model parameterizations. For latent-variable policies, preserving demonstrated modes requires action-conditioned information in the latent representation. Excessive posterior–prior regularization can suppress this information and prevent the policy from distinguishing demonstrated modes. Weaker or aggregate regularization can preserve mode information, but shifts the challenge to ensuring that the deployment-time prior covers the relevant latent regions. For action-space generative policies, multimodality is constrained by the smoothness of the base-to-action transport: a map with a small Lipschitz constant cannot assign substantial probability to many well-separated modes. Covering many modes therefore requires either sharp transitions in base space or off-support bridge regions in action space. Experiments on synthetic multimodal navigation and a physical-robot bimodal manipulation task support these mechanisms. In contrast, our analysis reveals limited conditional
multimodality in standard robotic simulation benchmarks, where
deterministic regression remains competitive.
\end{abstract}

\input{sections/01_introduction}
\input{sections/02_background}
\input{sections/04_main_results}

\input{sections/05_experiments}
\input{sections/06_conclusion}

\input{sections/07_ack}
\clearpage
\bibliographystyle{plainnat}
\bibliography{references}

\clearpage
\appendix
\startcontents[appendices]
\section*{Appendix Overview}
\printcontents[appendices]{l}{1}{\setcounter{tocdepth}{2}}
\clearpage
\input{sections/A_appendix}
\input{sections/A2_appendix_related_work}
\input{sections/B_appendix}
\input{sections/C_appendix}

\clearpage
\input{checklist}

\end{document}

%% file: sections/01_introduction.tex
\section{Introduction}
\label{sec:intro}

Imitation learning trains a robot
policy directly from expert demonstrations
\citep{pomerleau1988alvinn,ross2011reduction,ho2016generative}.
A standard approach is behavioral cloning (BC), which fits a policy to
observation--action pairs collected under an expert policy, without the
need for online interaction, reward engineering, or exploration.
Due to its practical appeal, BC has become a dominant paradigm for training deep visuomotor policies
in robotic manipulation
\citep{zhao2023learning,chi2025diffusion,team2024octo,kim2024openvla,black2024pi0}.
Recent work on behavioral cloning has highlighted multimodality in the
expert's conditional action distribution as a central challenge
~\cite{florence2021implicit, shafiullah2022bet, chi2025diffusion, lee2024vqbet, sheebaelhamd2025quantization}: for the same observation, the
dataset may contain multiple valid actions. This may arise from different expert styles, equally good choices, or
latent factors absent from the observation space~\citep{li2017infogail}. Under an $L_2$ or unimodal-Gaussian
behavioral-cloning head, the maximum-likelihood solution regresses
to the conditional \emph{mean} of the modes, which is typically not
an element of any mode's
support~\citep{wang2022diffusion, florence2021implicit}. A canonical example is inverse kinematics in robot control\citep{bishop1994mdn}: the same end-effector target may admit disconnected joint-space solutions. Averaging can yield a joint command that lies on none of the inverse-kinematics branches, while filtering to one branch removes feasible alternatives needed under constraints such as obstacles or joint limits.

Contemporary multimodal BC methods typically predict short chunks of
future actions rather than single-step actions, improving temporal
coherence while enabling one-shot non-autoregressive inference over the
chunk
\citep{zhao2023learning,chi2025diffusion}. There are two broad approaches to making such policies multimodal: \textit{latent-variable methods} and \textit{action-space generative methods}. Latent-variable methods introduce an
unobserved code, continuous or discrete, from which different action
chunks can be decoded~\citep{zhao2023learning,lee2024vqbet,wang2025vq}. Action-space generative methods model the
distribution over action chunks directly, using flows or diffusion models~\citep{chi2025diffusion, black2024pi0, intelligence2025pi05}. Both approaches can generate diverse actions, and their multimodality is typically evaluated by rollout performance. 

In both approaches, however, it remains unclear which model- and training-level quantities control the multimodality of the learned policy. Our goal is to fill this gap by providing a precise definition of multimodality. Furthermore, we identify the key quantities that must be controlled to preserve multimodality: the posterior–prior regularization strength in CVAE-based policies, and the Lipschitz constant of the base-to-action map in action-space generators. 

\paragraph{Contribution.}
 We study how posterior-prior regularization affects multimodality in action-chunking behavioral cloning. For latent-variable policies, we show that preserving demonstrated modes requires action-conditioned latent information; see Proposition \ref{prop:mode-info}. We then show that pointwise posterior–prior regularization can suppress this information as its strength increases, giving the multimodality collapse criterion in Corollary \ref{cor:beta-critical}. For action-space generators, we show that a deterministic sampler with a small Lipschitz constant is more likely to assign a small probability to many modes; see Proposition \ref{prop:multflow}, losing multimodality. Therefore, regularization methods that enforce a small Lipschitz constant tend to suppress multimodality. We empirically validate these mechanisms on controlled multimodal navigation and manipulation
tasks and provide practical heuristics. We also introduce a diagnostic
for conditional multimodality in robot demonstration datasets.
Applied to standard robotic simulation benchmarks, it detects
limited conditional multimodality, while our policy evaluations
show that deterministic regression remains competitive.
We release the code used in the experiments%
\footnote{\url{https://github.com/Lorenzo-Mazza/VersatIL}}.
\paragraph{Structure of the paper.}Section~\ref{sec:background} defines
multimodal behavioral cloning over actions and the generative
policy families considered in the paper. Section~\ref{sec:theory}
formalizes mode preservation and proves that latent policies must carry
action-conditioned mode information, leading to a collapse criterion for
pointwise prior regularization, a latent-geometry condition for
aggregate matching, and a Lipschitz lower bound for action-space
generators. Section~\ref{sec:experiments} introduces controlled synthetic
benchmarks with ground-truth mode labels and evaluates representative
methods on synthetic and robotic tasks to test whether these mechanisms explain practical failures. Notation, proofs, related works, and experimental details are described in the Appendices.

%% file: sections/02_background.tex
\section{Background}
\label{sec:background}
\label{sec:bg_bc}
For any notation that may be unclear, we refer the reader to Appendix~\ref{app:notation}.  
\paragraph{Offline imitation learning.}
We consider a finite-horizon Markov decision process \((\mathcal S,\mathcal A,P,r,\rho_0,\mathsf T)\), where \(\mathcal S\) is the state or observation space, \((\mathcal A,d_{\mathcal A})\) is the (metric) action space, \(P(\cdot\mid s,a)\in \mathcal P(\mathcal S)\) is the transition kernel, \(r:\mathcal S\times\mathcal A\to\mathbb R\) is the reward, \(\rho_0\in\mathcal P(\mathcal S)\) is the initial-state distribution, and \(\mathsf T \in \mathbb{N}\) is the horizon. A stochastic policy \(\pi:\mathcal S \to \mathcal P(\mathcal A)\) maps each state $s$ to a probability distribution \(\pi(\cdot\mid s)\) over actions. In the offline-imitation-learning setting, the reward signal $r$ is not observed. Instead, we
are given demonstrations generated by an expert policy \(\pi_E\).  A trajectory realization is $\tau=(s_0,a_0,\ldots,(s_{\mathsf T-1},a_{\mathsf T-1}),s_{\mathsf T})
\in
\mathcal T_{\mathsf T}
:=
(\mathcal S\times\mathcal A)^{\mathsf T}\times\mathcal S$.
The expert policy, together with \(P\) and \(\rho_0\), induces a
trajectory law \(p_E\in\mathcal P(\mathcal T_{\mathsf T})\). When the relevant
densities or probability mass functions exist, this law factorizes as
$ p_E(\tau) = \rho_0(s_0) \prod_{t=0}^{\mathsf T-1} \pi_E(a_t\mid s_t)\,P(s_{t+1}\mid s_t,a_t)$. We denote by \(p_\mathcal{D}\in\mathcal P(\mathcal S\times\mathcal A)\)
the time-marginal expert state-action distribution:
\[
p_\mathcal{D}(s,a)
:=
\frac{1}{\mathsf T}
\sum_{t=0}^{\mathsf T-1}
p_E(S_t=s,A_t=a) ,
\] where \(p_E(S_t=s,A_t=a)\) denotes the \(t\)-step state-action marginal
induced by the trajectory law \(p_E\).
The goal of offline imitation learning is to learn a policy
\(\pi_\theta(\cdot\mid s)\) from samples drawn from this expert state-action law described by $p_{\mathcal D}(s,a)$.
Let \(
 T_1,\ldots, T_N \overset{\mathrm{iid}}{\sim} p_E\)
be independent expert trajectory random variables. The dataset of expert demonstrations is one realization $\tilde{\mathcal D}
=
\{\tau_i\}_{i=1}^N
=
\big\{
(s^i_0,a^i_0,\ldots,s^i_{\mathsf T-1},a^i_{\mathsf T-1},s^i_{\mathsf T})
\big\}_{i=1}^N $.
Using $\tilde{\mathcal D}$, we construct the empirical state-action
distribution $p_{\tilde{\mathcal D}}:= \frac{1}{N\mathsf T} \sum_{i=1}^{N} \sum_{t=0}^{\mathsf T-1} \delta_{(s_t^i,a_t^i)}$.

\remark{In practice, action-chunk BC heads \(a_{t:t+\mathsf H-1}\in\mathcal A^{\mathsf H}\) are commonly used. Action-chunk BC is executed open-loop between observations, to retain temporal dependence absent from \(p_{\mathcal D}\) and reduce rollout error~\citep{zhao2023learning,simchowitz2025pitfalls}. All definitions and results extend by replacing \(\mathcal A\) with \(\mathcal A^{\mathsf H}\), so samples become \((s_t^i,a_{t:t+\mathsf H-1}^i)\).}
\paragraph{Multimodal imitation learning.}
\label{par:multimodal_il}
We formalize \emph{multimodality} as a property of \(p_{\mathcal D}(\cdot\mid s)\). For each observation \(s\in\mathcal S\), the support of the expert conditional action distribution is contained in finitely many well-separated valid mode sets. In other words, for each \(s\in\mathcal S\), there exists a natural number \(\mathsf K(s)\ge 1\) and a family of covering sets \( C_1(s),\ldots, C_{\mathsf K(s)}(s)\subseteq\mathcal A\) of \(\operatorname{supp}p_{\mathcal D}(\cdot\mid s)\) such that
\begin{equation}\label{def:multimodal}
\operatorname{supp} p_{\mathcal D}(\cdot\mid s)
\subseteq
\bigcup_{k=1}^{\mathsf K(s)} C_k(s),
\qquad
\Delta(s)
:=
\min_{k\neq k'}
\operatorname{dist}_{\mathcal A}
\!\bigl(C_k(s),C_{k'}(s)\bigr)
>0 .
\end{equation}
Each set \(C_k(s)\) represents one valid strategy or trajectory
mode for observation \(s\). The mode sets have disjoint closures.
We further define the \emph{mode-assignment} function
\(g:\mathcal A\times\mathcal S\to\mathbb N_0\) as
\begin{equation}
\label{eq:modeassdef}
g(s,a)
=
\sum_{k=1}^{\mathsf K(s)}
k\,\mathbf 1_{\{a\in C_k(s)\}},
\qquad
\mathbf 1_{\{a\in C_k(s)\}}
=
\begin{cases}
1, & a\in C_k(s),\\
0, & \text{otherwise}.
\end{cases}
\end{equation}
The function \(g\) is a mode-level labeling of expert actions used for the analysis. For each realized pair \((s,a)\), \(g(s,a)\) is the deterministic index of the mode containing \(a\). Under the data distribution \((S,A)\sim p_{\mathcal D}\), we denote the induced mode-label random variable by \(M:=g(S,A).\) Here \(\mathsf K(s)=1\) gives the unimodal case, while \(\mathsf K(s)\ge 2\) is the setting of interest.

\paragraph{Behavioral cloning.} 
Behavioral cloning instantiates offline imitation learning as supervised learning on state-action pairs extracted from expert trajectories.
Let \(\pi_\theta(a\mid s)\) be the conditional action distribution given a state $s$ and parametrized by \(\theta\). In practice, \(\pi_\theta\) is represented by a neural network parametrized by \(\theta\), which maps each state \(s\) to a distribution over actions. The parameters \(\theta\) are usually determined by minimizing the standard maximum-likelihood objective
\begin{equation}
\label{eq:lossBC}
\mathcal{L}_{\mathrm{BC}}(\theta)
=
-\E_{(S,A)\sim p_{\mathcal D}}\!
\left[\log \pi_\theta(A\mid S)\right],
\end{equation}
which, in the infinite-data limit, is equivalent \citep{hoffman2013stochastic} up to an additive
constant to minimizing the Kullback-Leibler (KL) divergence between the
expert data and the model distribution $\E_{S\sim p_{\mathcal D,S}}
\KL\!\left(
p_{\mathcal D}(\cdot\mid S)
\,\|\, 
\pi_\theta(\cdot\mid S)
\right),
$
where \(p_{\mathcal D,S}\) denotes the state marginal of
\(p_{\mathcal D}\).

\paragraph{Conditional latent-variable policies.}
Latent-variable BC policies introduce an auxiliary random variable
\(Z\in\mathbb R^{\mathsf d_z}\) to represent action ambiguity not
resolved by the observation. They consist of a posterior encoder
\(q_\phi(\cdot\mid a,s)\), a conditional prior \(p_\psi(\cdot\mid s)\),
and a latent-conditioned decoder policy \(\pi_\theta(\cdot\mid z,s)\).
At deployment, $Z\sim p_\psi(\cdot\mid s),\ A\sim \pi_\theta(\cdot\mid Z,s)$. We write their training objective in the generic regularized form
\begin{equation}
\mathcal{L}^\beta_{\mathrm{var}}(\theta,\phi,\psi)
=
\E_{(S,A)\sim p_{\mathcal D}}\!\left[
-\E_{Z\sim q_\phi(\cdot\mid A,S)}
\log \pi_\theta(A\mid Z,S)
\right]
+
\beta \mathit{D}(q_\phi,p_\psi),
\label{eq:latent_bc_general}
\end{equation}
where \(\mathit D(q_\phi,p_\psi)\) couples the posterior encoder to the
deployment prior and \(\beta>0\) controls its strength.

\paragraph{Action-space generative policies.}
Action-space generative policies sample actions directly by transforming
base noise into actions. For fixed \(s\), we write the full deterministic
sampler as $G_{\theta,s}(u):=G_\theta(s,u),\ 
U\sim p_0,\ 
A=G_{\theta,s}(U).$
The induced conditional policy is therefore the push-forward $\pi_\theta(\cdot\mid s)
=
(G_{\theta,s})_{\#}p_0$.
This abstraction covers deterministic samplers obtained from diffusion
or flow-matching policies.

%% file: sections/04_main_results.tex
\section{Main Results}
\label{sec:theory}
We analyze how multimodality appears in the action-chunking generative policy families.
For latent-variable policies, preserving mode identity requires action information in the latent variable. For action-space
generative policies, the constraint is geometric: separated modes must be
realized by the sampling map from base noise to action chunks. The reason for these two different analyses is that, for CVAEs, the action information is a natural quantity for studying multimodality, whereas for generative policies it is meaningless, since there is no information bottleneck.
\paragraph{Multimodality in latent-variable policies.}
\label{sec:theory-latent}
Latent-variable policies route multimodal action ambiguity through an explicit latent random variable \(Z\). We first quantify the information required to preserve the demonstrated mode. Let
\[
q_{\phi,\mathcal D}(s,a,z)
:=
p_{\mathcal D}(s,a)\,q_\phi(z\mid a,s)
=
p_{\mathcal D,S}(s)\,
p_{\mathcal D}(a\mid s)\,
q_\phi(z\mid a,s)
\]
denote the posterior-encoder-induced joint distribution over
\((S,A,Z)\). Unless stated otherwise, all probabilities, mutual informations, expectations, and
conditional distributions involving \((S,A,Z)\) in this subsection are
taken under \(q_{\phi,\mathcal D}\). We write the
state-conditioned aggregated posterior as $q_{\phi,\mathcal D}(z\mid s)
:=
\mathbb{E}_{A\sim p_{\mathcal D}(a\mid s)}[q_\phi(z\mid A,s)] $.
The conditional mutual information between \(A\) and \(Z\) given \(S\) is denoted by \(I(A;Z\mid S)\). From ~\cite{hoffman2016elbosurgery}, it holds
\[
\E_{(S,A)\sim p_{\mathcal D}}
\KL\!\left(
q_\phi(\cdot\mid A,S)
\,\|\, 
p_\psi(\cdot\mid S)
\right)
=
I(A;Z\mid S)
+
\E_{S\sim p_{\mathcal D, S}}
\KL\!\left(
q_{\phi,\mathcal D}(\cdot\mid S)
\,\|\, 
p_\psi(\cdot\mid S)
\right).
\]
The first term measures how much information \(Z\) must encode about the demonstrated action \(A\) beyond what is already determined by the state \(S\), while the second term measures aggregated posterior--prior mismatch. Thus \(I(A;Z\mid S)\) captures the action information available to preserve the demonstrated mode: if \(S\) already determines \(A\) uniquely, then \(Z\) need not contain additional information about \(A\). 

In the rest of the analysis, we assume that the latent-conditioned decoder policy \(\pi_\theta(\cdot\mid z,s)\) is concentrated on the graph of a neural network \(f_\theta\), that is \(\pi_\theta(\cdot\mid z,s)=\delta_{f_\theta(z,s)}\), with decoded action \(a=f_\theta(z,s)\). Therefore, given a sample \((s,a)\in\operatorname{supp}(p_{\tilde{\mathcal D}})\), the
training-time prediction law is
\(p_{\theta,\phi}(\cdot\mid s,a):=(f_{\theta,s})_{\#}
q_\phi(\cdot\mid a,s)\), where \(f_{\theta,s}(z):=f_\theta(z,s)\).
A realized action prediction \(\hat a\sim p_{\theta,\phi}(\cdot\mid s,a)\)
is correct if and only if \(\hat a\in C_{g(s,a)}(s)\).

For a measurable function \(\rho:\mathcal S\to[0,1]\) representing a state-dependent mode-identification error, we define the Fano-corrected lower bound as
\[
B_\rho
:=
\E_{S\sim p_{\mathcal D,S}}\!\left[H(M\mid S)\right]
-
\E_{S\sim p_{\mathcal D,S}}
\Big[
H_b(\rho(S))+\rho(S)\log(\mathsf K(S)-1)
\Big],
\]
where \(H_b(p)=-p\log p-(1-p)\log(1-p)\) is the entropy of a \(\mathrm{Bernoulli}(p)\) random variable, and the bracketed term is taken to be zero when \(\mathsf K(S)=1\). The quantity \(B_\rho\) measures the portion of the expected conditional mode entropy \(\E_{S\sim p_{\mathcal D,S}}[H(M\mid S)]\) that remains after applying a Fano correction for \(\rho(S)\). A positive value indicates that, under the data distribution, the mode \(M\) cannot be resolved from \(S\) alone up to the specified error \(\rho(S)\), providing a certificate of unresolved conditional multimodality.

The next result lower bounds \(I(A;Z\mid S)\) by the Fano-corrected entropy of the
demonstrated mode label \(M\).

\begin{proposition}[Mode-information lower bound]
\label{prop:mode-info}
Under the setting introduced above, for each \(s\in\mathcal S\), let
\begin{equation}
\label{eq:assprop}
\rho(s)
:=
\Pr\!\left(
f_\theta(Z,s)\notin \mathcal C_M(s)
\,\middle|\,
S=s
\right).
\end{equation}For all \(s\in\operatorname{supp}(p_{\mathcal D,S})\) with
\(\mathsf K(s)>1\), assume
\(\rho(s)\le 1-1/\mathsf K(s)\). Then $I(A;Z\mid S)\ge B_\rho$. Consequently, it holds $\E_{(S,A)\sim p_{\mathcal D}}
\KL\!\left(
q_\phi(\cdot\mid A,S)
\,\|\, 
p_\psi(\cdot\mid S)
\right)
\ge B_\rho $.
\end{proposition}

The detailed proof of Proposition \ref{prop:mode-info} is given in Appendix~\ref{app:proofs}. We note that the assumption of Proposition~\ref{prop:mode-info} ensures that, for all \(s\in\operatorname{supp}(p_{\mathcal D,S})\) with \(\mathsf K(s)>1\), the induced mode-recovery rule is nontrivial: its error probability \(\rho(s)\) is no larger than the error rate \(1-1/\mathsf K(s)\) of uniform random guessing among the \(\mathsf K(s)\) admissible modes. The result shows that accurate mode recovery forces the posterior latent to retain action-conditioned information: the better the decoder preserves the demonstrated mode, the larger the lower bound \(B_\rho\).


\begin{corollary}[Uniform exact-recovery case]
  \label{cor:logk}
Under the setting of Proposition~\ref{prop:mode-info}, if \(\rho(s)=0\) for \(p_{\mathcal D,S}\)-almost every \(s\), and \(M\mid S=s\) is uniform over \(\mathsf K(s)\) modes, then
\[
I(A;Z\mid S)
\ge
\E_{S\sim p_{\mathcal D,S}}\log \mathsf K(S).
\]
\end{corollary}
Corollary~\ref{cor:logk} isolates the maximum-ambiguity case: exact recovery of \(\mathsf K(s)\) equiprobable modes requires the latent to carry on average the cost of the full mode entropy, \(\log \mathsf K(s)\) nats.

We recall that CVAEs are usually trained using a regularized objective \(\mathcal{L}^\beta_{\mathrm{var}}(\theta,\phi,\psi)\) defined in \eqref{eq:latent_bc_general}. A widely used regularizer~\citep{zhao2023learning} is given by
\begin{equation}
\label{eq:regular}
\mathit D(q_{\phi}, p_{\psi})
=
\E_{(S,A)\sim p_{\mathcal D}}
\KL\!\left(
q_\phi(\cdot\mid A,S)
\,\|\,
p_\psi(\cdot\mid S)
\right).
\end{equation}

\begin{corollary}
\label{cor:beta-critical}
Let us fix $\beta > 0, \mathsf b_0>0$. Let us denote with 
\[
(\hat{\theta},\hat{\phi},\hat{\psi})
:=
\arg\min_{\theta,\phi,\psi}
\mathcal L^\beta_{\mathrm{var}}(\theta,\phi,\psi),
\qquad
\hat{\mathcal L}^{\beta}
:=
\mathcal L^\beta_{\mathrm{var}}
(\hat{\theta},\hat{\phi},\hat{\psi})
\]
the solution of the learning problem and the achieved training loss, respectively. Let us assume that: (H1) there exists \((\theta_0,\phi_0,\psi_0)\) such that
    \(\mathit D(q_{\phi_0},p_{\psi_0})=0\), and denote by \(\mathcal L^0=
    \mathcal L^0_{\mathrm{var}}(\theta_0,\phi_0,\psi_0)\)
    the corresponding training loss; (H2) the mode-recovery error $\hat\rho(s)
    :=
    \Pr\!\left(
    f_{\hat\theta}(Z,s)\notin C_M(s)
    \,\middle|\, S=s
    \right)$
    satisfies
    \(\hat\rho(s)\le 1-1/\mathsf K(s)\) for all
    \(s\in\operatorname{supp}(p_{\mathcal D,S})\) with \(\mathsf K(s)>1\). Then there exists a constant \(\mathsf C>0\), independent of \(\beta\), such that \(B_{\hat\rho}\le I(A;Z\mid S) \le \frac{\mathsf C}{\beta}\). Consequently, a necessary condition for achieving a certificate \(B_{\hat\rho}> \mathsf b_0\) is \(\beta < \frac{\mathsf C}{\mathsf b_0}.\)
    
\end{corollary}
\begin{remark}
    We note that the Fano term is what turns the information bound into a mode-recovery statement: \(I(A;Z\mid S)\) may encode any action-dependent variation, whereas \(B_\rho\) is positive only when unresolved mode entropy
\(H(M\mid S)\) is recovered with sufficiently small error \(\rho\).
\end{remark}
Assumption \textit{(H1)} requires \(q_{\phi_0}\) and \(p_{\psi_0}\) to coincide for some parameter values \((\phi_0,\psi_0)\), so that \(\mathit D(q_{\phi_0},p_{\psi_0})=0\). For the pointwise KL regularizer, this means that \(q_{\phi_0}(\cdot\mid a,s)=p_{\psi_0}(\cdot\mid s)\) on the data support, and therefore \(Z\) does not encode additional information about \(A\) beyond \(S\). This collapse behavior is observed in practice; see, for instance,~\cite{bowman2016generating,chen2016variational}. Assumption \textit{(H2)} ensures that the learned mode-recovery rule is nontrivial, since its error probability is no larger than that of uniform random guessing among the \(\mathsf K(s)\) admissible modes.

Corollary~\ref{cor:beta-critical} shows that the strength of the variational regularization directly limits the amount of certified conditional multimodality that the learned latent representation can retain. In particular, as \(\beta\) increases, the conditional mutual information \(I(A;Z\mid S)\) is forced to decrease at rate \(1/\beta\). Since \(B_{\hat\rho}\) is upper-bounded by this information term, strong regularization can prevent the latent variable from encoding the action-dependent mode information needed to represent multiple plausible actions at the same state. Therefore, obtaining a nontrivial multimodality certificate \(B_{\hat\rho}>\mathsf b_0\) requires the regularization strength to be sufficiently small, namely \(\beta<\mathsf C/\mathsf b_0\).

\begin{remark}
Corollary~\ref{cor:beta-critical} is specific to regularizers of the form \eqref{eq:regular}. For other regularization strategies, an analogous relation between the regularization strength $\beta$ and the certified multimodality does not follow in general. Aggregate matching objectives provide a counterexample to this behavior; see Appendix~\ref{app:proofs-aggregate} for details.
\end{remark}

\paragraph{Multimodality in action-space generative policies.}
\label{par:action-space-theory}
Action-space generative policies remove the explicit posterior latent, but not the geometric problem of routing samples to separated action modes. In this paragraph, we investigate when multimodality is preserved by such generative policies.

We consider a generative policy defined as follows. Let us fix \(s\in\mathcal S\), $d\ge 1$ a positive integer and let \(p_0:=\mathcal N(0,I_d)\),
\(U\sim p_0\), where $I_d\in\R^{d\times d}$ denotes the identity matrix. The transport map, or deterministic sampler, at state $s$ with parameters $\theta$ is defined as \(G_{\theta,s}(u):=G_\theta(s,u)\). The induced conditional policy is the push-forward of the Gaussian base distribution \(\pi_\theta(\cdot\mid s)=(G_{\theta,s})_{\#}p_0=(G_{\theta,s})_{\#}\mathcal N(0,I_d).\)
In other words, for any measurable set \(B\subseteq\mathcal A\),
\(\pi_\theta(\cdot\mid s)\) is defined by
\[
\pi_\theta(B\mid s)
=
\Pr_{U\sim\mathcal N(0,I_d)}
\left(G_\theta(s,U)\in B\right)
=
\int_{\{u:\,G_\theta(s,u)\in B\}}
\frac{1}{(2\pi)^{d/2}}e^{-\|u\|_2^2/2}\,du .
\]
In practice, \(G_\theta\) is parameterized by a neural network and trained to transform Gaussian samples into actions whose conditional distribution approximates \(p_{\mathcal D}(\cdot\mid s)\).

We first introduce the notion of a \(\tau\)-represented mode.
\begin{definition}
Let \(\tau\in(0,1]\) and \(s\in\mathcal S\), and assume that
\(p_{\mathcal D}(\cdot\mid s)\) is multimodal, see \ref{def:multimodal}. We say that a mode
\(C_k(s)\) is \(\tau\)-represented by the policy distribution
induced by \(G_{\theta,s}\) if and only if $
\pi_\theta(C_k(s)\mid s)>\tau $.
\end{definition}
This definition means that a mode is considered represented whenever the learned policy places sufficiently large probability, greater than \(\tau\), on actions belonging to that mode. We define the number of \(\tau\)-represented modes by 
\[
N_{\theta}^{(\tau)}(s)
:=
\big|\left\{
k\in\{1,\dots,\mathsf K(s)\}
:
\pi_{\theta}(C_k(s)\mid s)>\tau
\right\}\big|.
\]
The Lipschitz constant of the map \(u\mapsto G_{\theta,s}(u)\) strongly controls the number of modes that the policy distribution generates with sufficiently large probability.

\begin{proposition}\label{prop:multflow}
Let us fix $s\in \mathcal{S}, \tau \in (0,1)$. Let us assume that the transport map $G_{\theta,s}:u\mapsto G_\theta(s,u)$ is $L_{\theta, s}$-Lipschitz. Then
\[
N_{\theta}^{(\tau)}(s)
\le
1+
\left\lfloor
\frac{
2\bigl(\sqrt{2\pi}\,|\Phi^{-1}(\tau)|+1\bigr)
}{
\sqrt{2\pi}\,\tau
}
\frac{L_{\theta,s}}{\Delta(s)}
\right\rfloor ,
\]where \(\Delta(s)\) is the mode separation defined above and $\Phi^{-1}$ is the quantile of the standard normal distribution. 
\end{proposition}

We note that the $L_{\theta,s}$-Lipschitz assumption on $G_{\theta, s}$ is broadly satisfied in practical applications, as it only requires the activation function of the neural network to be Lipschitz. Most commonly used activation functions satisfy this property. If the modes are well separated, a generator with a small Lipschitz constant cannot assign probability mass larger than \(\tau\) to many different modes at the same time. In other words, covering many separated modes requires the generator to stretch the latent space sufficiently. The number of \(\tau\)-represented modes grows with the ratio \(L_{\theta,s}/\Delta(s)\). This result can be interpreted as a mode-representation limitation for Lipschitz generators. A smoother or less expansive generator is more constrained in the number of modes it can represent, while a generator with larger Lipschitz constant has more capacity to spread probability mass across distant regions of the action space. In continuous-time diffusion and flow matching policies, separating modes requires extreme spatial stretching near the terminal sampling steps \citep{stephanovitch2026lipschitz}. Consequently, applying any stabilization or regularization technique that overly restricts the network's Lipschitz constant will inevitably compromise this stretching mechanism and force the generative model into mode collapse \citep{salmona2022can, stephanovitch2026lipschitz}.
\begin{remark}
    If the base space $U$ is replaced by the latent space $\mathcal Z$ and the policy $\pi_\theta(\cdot\mid s)$ by the decoder $p_\theta(\cdot\mid z,x)$ of the CVAE, Proposition~\ref{prop:multflow} yields the corresponding bound for the decoder, whenever it satisfies the analogous Lipschitz assumption.
\end{remark}

%% file: sections/05_experiments.tex
\section{Experiments}
\label{sec:experiments}
\begin{figure}[t]
\centering
\includegraphics[width=0.8\linewidth]{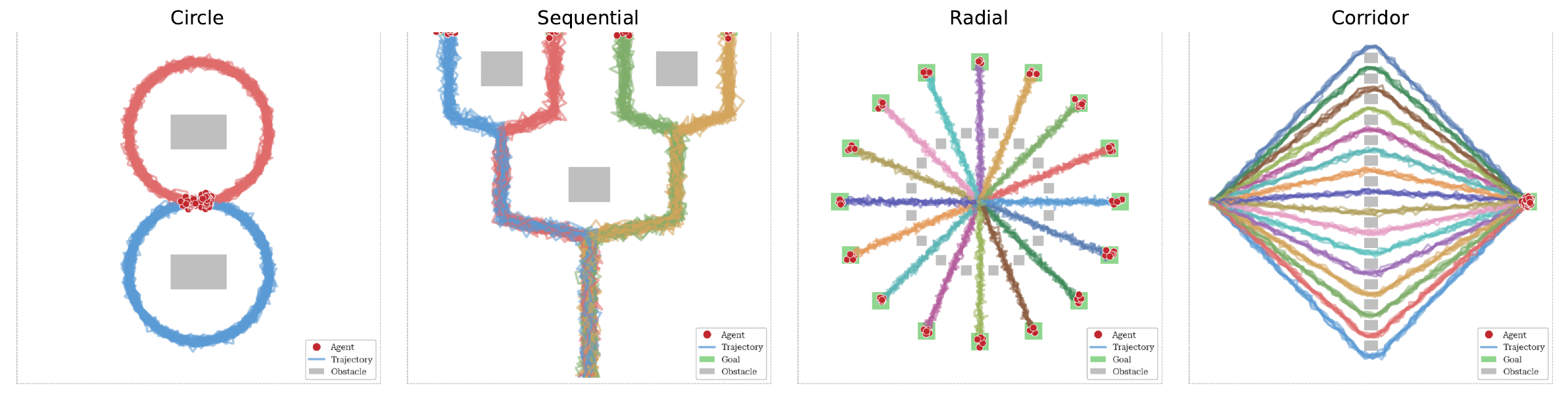}
\caption{Synthetic datasets. Each panel shows expert trajectories for
one task, color-coded by ground-truth mode. Grey rectangles denote
obstacles, green rectangles denote goal regions, and red dots denote final agent positions. Policies
receive a single RGB observation at $t=0$ and predict the
full open-loop chunk $a_{0:\mathsf H-1}\in\mathbb{R}^{60\times 2}$ without
replanning.}
\label{fig:synthetic-overview}
\end{figure}

\paragraph{Synthetic navigation tasks.}
We design four 2D multimodal navigation tasks with ground-truth modes (Figure~\ref{fig:synthetic-overview}), which allow us to test five research questions from the theory: (RQ1) Is mode coverage necessary for successful rollouts when the true data distribution is multimodal? (RQ2) Do the lower bounds from Proposition \ref{prop:mode-info} and Corollary \ref{cor:logk} hold in practice, and how tight are they? (RQ3) How can we provide practical recipes to pick a value for the KL regularizer $\beta$?  (RQ4, Appendix \ref{app:results-agg-matching}) Can aggregate posterior matching preserve latent mode information, and how does the geometry of the deployment prior interact with the training posterior? (RQ5) What are the practical consequences of the relationship between $L_{\theta,s}$ and the number of $\tau$-represented modes $N_{\theta}^{(\tau)}$ of the action-space generators identified by Proposition \ref{prop:multflow}? 
\paragraph{Simulated and real-world robot demonstrations.}
While the synthetic tasks provide ground-truth modes for testing the theory, robot demonstration datasets typically lack such labels. We therefore introduce
a heuristic to estimate the conditional action multimodality of a demonstration dataset. This enables us to study the interaction between the multimodality of the policies and the eventual multimodality of the training demonstrations. In simulation, we use Push-T~\citep{chi2025diffusion},
UR3 BlockPush~\citep{florence2021implicit}, LIBERO\citep{liu2023libero}, Meta-World \citep{yu2020meta}, and
Kitchen~\citep{gupta2019relay}, benchmarks previously used to evaluate
multimodal and vision-language policies~\citep{chi2025diffusion, lee2024vqbet,
shafiullah2022behavior, sheebaelhamd2025quantization}.
For the real-world evaluation, we design a proof-of-concept bimodal tissue-grasping task using a surgical robotic
platform~\citep{rodriguez2026open}.

 An extensive description of the policy parametrizations, architectures, training recipes, and hyperparameters is given in Appendices~\ref{app:exps} and~\ref{app:hyperparameters}.

\subsection{Answers to the research questions from the theory}
\paragraph{RQ1: when data is multimodal, mode coverage is necessary but not sufficient.}
Full per-task synthetic results are reported in Appendix~\ref{app:results}. 
In the presence of multimodal data, mode coverage proves to be necessary for successful rollouts: the deterministic baseline indeed collapses on the invalid average trajectory on all four tasks (Fig. \ref{fig:baseline-collapse}), achieving zero success.  However, mode coverage alone is not a sufficient condition for valid deployment: the generative policies recover high valid mode coverage, but on the two hardest \(\mathsf K=16\) tasks, success ranges from \(0.37\) to \(0.52\) despite near-complete coverage. 


\begin{figure}[t]
    \centering
    \includegraphics[width=0.8\linewidth]{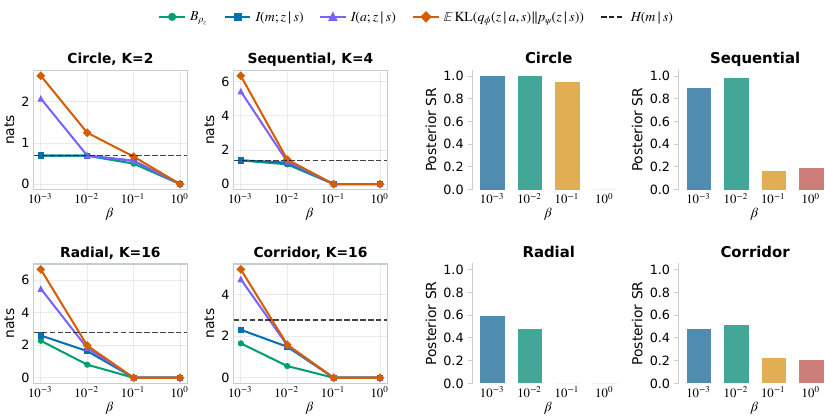}
\caption{
\textbf{Pointwise KL can suppress latent mode information.}
Information decomposition for KL-CVAE under increasing pointwise KL
regularization. For each task, we estimate the latent-space Fano
lower bound \(B_{\rho_z}\), the mode information
\(I(M;Z\mid S)\), the action information \(I(A;Z\mid S)\), and the
pointwise KL regularizer
\(\mathcal K_{\rm pt}
=
\E_{(S,A)\sim p_{\mathcal D}}
\KL(q_\phi(\cdot\mid A,S)\,\|\,p_\psi(\cdot\mid S))\).
The dashed line denotes the empirical mode entropy
\(H(m\mid s)= \log \mathsf K\), since the $K$ modes are balanced. On the right, the success rates (SR) achieved by sampling from the posterior.}
\label{fig:kl-info-decomposition}
\end{figure}

\paragraph{RQ2: tightness of the lower bounds and exact mode recovery.}
\label{par:kl-collapse}
In Figure \ref{fig:kl-info-decomposition}, the lower bounds from Proposition \ref{prop:mode-info} hold. The gap between \(B_{\rho_z}\) and
\(I(M;Z\mid S)\) shows the looseness of the Fano bound, which depends on both the mode-recovery error rate and how errors are distributed. The gap between \(\mathcal K_{\rm pt}\) and \(I(A;Z\mid S)\) is the mismatch term $\mathbb E_S(D_{KL}(q_{\phi, \mathcal D}(\cdot|S), p_{\psi, \mathcal D}(\cdot |S))$ ~\citep{hoffman2016elbosurgery}, which measures how well the learned aggregated posterior matches the prior, and gives an indication of how well the model will perform at inference time, when sampling $Z$ from the prior.
Consistent with the mode recovery statement in Corollary \ref{cor:logk}, at small \(\beta\), the mode information \(I(M;Z\mid S)\) is close to \(H(M\mid S)\), indicating that the latent variable nearly identifies the demonstrated mode. As \(\beta\) increases, both \(I(M;Z\mid S)\) and \(I(A;Z\mid S)\) decrease sharply beyond $H(M|S)$; consequently, the downstream success rate of the posterior-sampled policy collapses toward zero.  

\paragraph{RQ3: selection rule for the value of $\beta$.}
Let us fix a desired error probability $\hat \rho$ and compute an estimate of the Fano lower bound $b_0(\bar\rho)$ as in \eqref{eq:estimate_of_Fano}. Corollary~2 provides a threshold $\beta_{\max}:=\frac{\mathcal L^0}{b_0(\bar\rho)}$ above which no solution can have the desired mode-recovery accuracy, but for a choice of $\beta$ below that bound it does not guarantee that training will find a multimodal solution. In Appendix~\ref{app:kl-latent-mode-collapse}, we provide a heuristic for choosing $\beta$ for a desired mode-recovery error $\bar\rho$.
To fulfill the hypothesis of Corollary \ref{cor:beta-critical}, it is required to check whether the trained policy (with $\beta<\beta_{\max}$) achieves the desired mode-recovery error and a total loss below the collapsed reference $\mathcal L^0$. In Table \ref{sweepbeta} we evaluate several values of $\beta$ for two of the synthetic tasks. In every run,
the empirical inequality\( B_{\hat\rho}\leq D_{\mathrm{KL}}\leq \mathcal L^0/\beta \) holds.  For both tasks, all tested $\beta$ values below the heuristic
recommendation successfully achieve $\hat\rho\le\bar\rho$. Additionally, consistently with Corollary \ref{cor:beta-critical}, every run with $D_{\mathrm{KL}}<b_0(\bar\rho)$ has
$\hat\rho>\bar\rho$, and no run above $\beta_{\max}$ achieves
the desired mode recovery. Several runs below $\beta_{\max}$ also fail to achieve it, confirming that satisfying the
theoretical threshold alone does not ensure successful mode recovery.

\begin{figure}[t]
\centering
\includegraphics[width=0.75\linewidth]{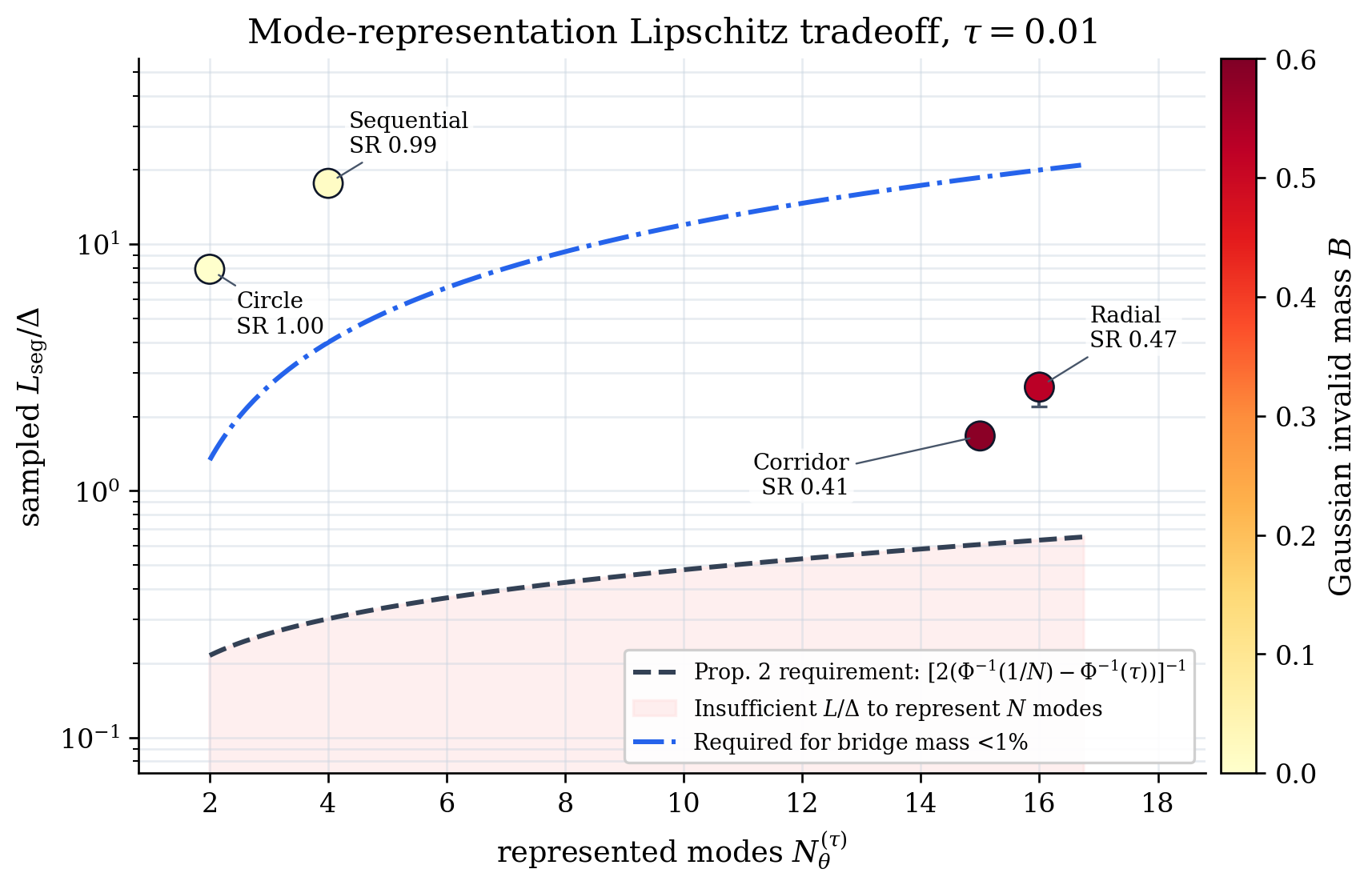}
\caption{ \textbf{Action-space sampler tradeoff.} Diffusion and flow-matching-based samplers trade off a high Lipschitz constant for bridge regions. Act-Flow results on the synthetic benchmark are represented.}
\label{fig:action-bridge-bound}
\end{figure}

\paragraph{RQ5: relationship between $L_{\theta, s}$ and $N_\theta^{(\tau)}$ in action-space samplers.}
From Proposition~\ref{prop:multflow} and Corollary \ref{cor:stretch-bridge-tradeoff}, it follows that a Lipschitz base-to-action
map cannot place non-negligible mass on many well-separated modes unless its Lipschitz constant is sufficiently large, or its bridge (the amount of probability mass in the base space that decodes to invalid trajectories) is large. We aim to test this trade-off empirically. We first estimate the Lipschitz constant using finite differences of the cross-mode transitions (details in Appendix \ref{app:action-bridge-diagnostic}), and then measure the empirical $L_{seg}/\Delta$; see \eqref{eq:sseg} for the definition, and $\Delta$ is the mode distance. Figure~\ref{fig:action-bridge-bound} compares the measured
$L_{\mathrm{seg}}/\Delta$ with the necessary theoretical lower bounds for mode representation at $\tau=0.01$ (Proposition~\ref{prop:multflow}) and for maintaining that representation while limiting bridge mass to $1\%$ (Corollary~\ref{cor:stretch-bridge-tradeoff}).
The benchmarks that can represent all ground-truth modes at $\tau=0.01$ lie above the bound; Corridor represents $15$ of its $16$ modes despite exceeding it, as the bound is a necessary condition.
The Circle and Sequential runs have high normalized transition sensitivities of approximately $7.9$ and $17.7$, negligible interpolation bridge fractions, and success rates of $1.00$ and $0.99$. On the other hand, Radial and Corridor exhibit lower measured sensitivities ($2.63$ and $1.67$), substantial interpolation bridge fractions ($0.46$ and $0.55$), and lower success rates ($0.47$ and $0.41$). These observations associate broad invalid bridge regions with lower Lipschitz constant and lower successful deployment, despite complete or near-complete mode coverage.

\subsection{Multimodality in simulation and real robot demonstrations}
\begin{figure}[t]
\centering
\includegraphics[width=\linewidth]{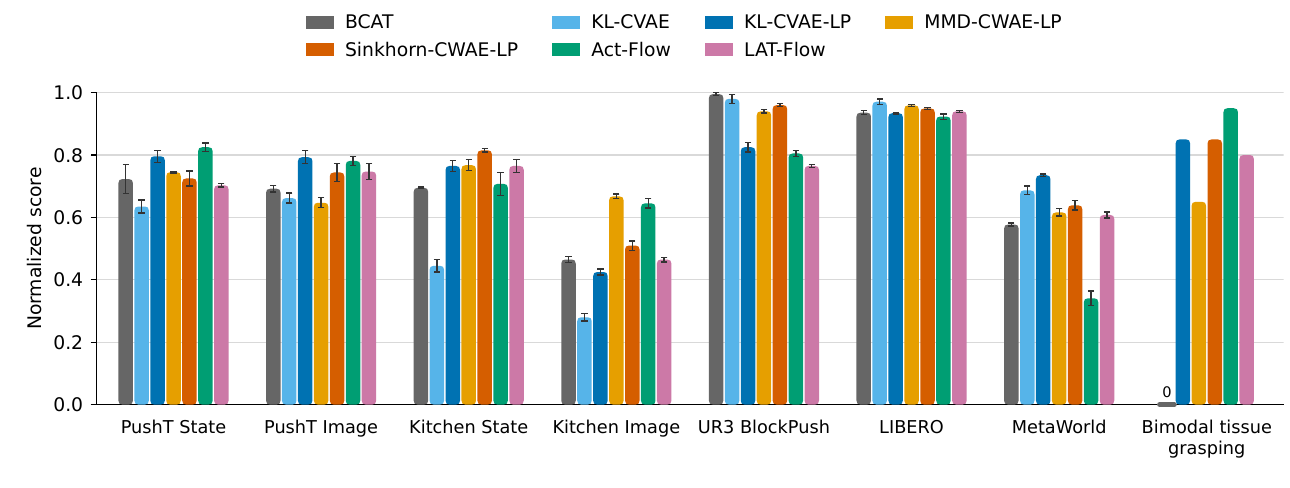}
\caption{Policy success rate on robot benchmarks, normalized to $[0,1]$. }
\label{fig:policy-performance}
\end{figure}

\paragraph{Estimating the number of modes for unlabeled data.} \label{sec:sim-results} 
Estimating $K(S)$ and $H(M|S)$ without knowledge about task semantics is a hard problem in general. We propose an estimator based on Gaussian-mixture modal clustering \citep{chacon2019mixture} of future action chunks associated
with nearby observation states (implementation details in Appendix \ref{app:results-data-ambiguity}), which yields estimates for the conditional local mode count and entropy.  We validate the estimator on the synthetic benchmarks (Table \ref{tab:synthetic-modes-compact}), where it reconstructs the true number of modes. We note here that the multimodality property of the data is horizon-dependent; therefore, the horizons chosen for evaluation are closely related to those used for policy training. Surprisingly, despite their use in prior work to evaluate policies designed to capture multimodality \cite{chi2025diffusion, shafiullah2022bet, lee2024vqbet, sheebaelhamd2025quantization}, most simulation benchmarks have almost-unimodal support across the dataset states. Push-T, UR3, LIBERO, and Meta-World all give estimates extremely close to 1, while Kitchen is the most multimodal at $1.231$ when the action horizon is 30. In our proof-of-concept robot dataset of bimodal tissue grasping, the multimodality is higher, and when the action horizon is long enough, it approximates the ground-truth value of $2$ semantic modes. 
\begin{table}[t]
    \centering
    \begin{minipage}[t]{0.38\linewidth}
        \centering
        \small
        \captionof{table}{
            Synthetic conditional-mode estimates, action horizon $=60$,
            radius $\varepsilon=0.5$, and modal-mass cutoff $2.5\%$.
            Entropy $\hat H$ is in bits.
        }
        \label{tab:synthetic-modes-compact}
        \begin{tabular}{@{}lrr@{}}
        \toprule
        Benchmark & $\widehat K$ & $\widehat H$ \\
        \midrule
        Circle     & $2.000$  & $0.998$ \\
        Sequential & $4.000$  & $1.986$ \\
        Radial     & $15.970$ & $3.972$ \\
        Corridor   & $15.920$ & $3.960$ \\
        \bottomrule
        \end{tabular}
    \end{minipage}
    \hfill
    \begin{minipage}[t]{0.58\linewidth}
        \centering
        \small
        \captionof{table}{
            Dataset-wide conditional-mode estimates at radius
            $\varepsilon=0.5$ and modal-mass cutoff $2.5\%$.
            Each value is reported per action horizon $=10/30$.
        }
        \label{tab:dataset-modes-compact}

        \begin{tabular}{@{}lrr@{}}
        \toprule
        Dataset & $\widehat K$ & $\widehat H$ \\
        \midrule
        UR3 BlockPush & $1.136/1.164$ & $0.082/0.103$ \\
        Kitchen       & $1.143/1.231$ & $0.083/0.147$ \\
        PushT         & $1.068/1.005$ & $0.038/0.004$ \\
        LIBERO        & $1.014/1.027$ & $0.007/0.017$ \\
        MetaWorld     & $1.037/1.067$ & $0.020/0.043$ \\
        Bimodal Tissue Grasping
                      & $1.468/2.267$
                      & ${0.432}/{0.760}$ \\
        \bottomrule
        \end{tabular}
    \end{minipage}

\end{table}  
\paragraph{Policy performance.} \label{sec:sim-results} 
We evaluate the baseline policies on the simulation and robot benchmarks (Fig. \ref{fig:policy-performance} and Appendix \ref{app:benchmark-results}). We emphasize here that our goal is not to obtain state-of-the-art performance, but to study how (estimated) multimodality of each robot benchmark's data distribution interacts with different policy parametrizations, and whether policy parametrizations that can represent this property lead to better downstream performance.
The largest performance gap between regression and multimodal-capable policies
occurs on bimodal tissue grasping, the dataset with the highest estimated multimodality: BCAT averages the two modes and achieves $0\%$ success. A parametrization unable to represent conditional multimodality cannot recover the full conditional expert distribution when the latter exhibits distinct modes. Mode averaging may still yield successful rollouts if the interpolation happens to be valid, but this reflects a fortuitous compatibility with the task constraints rather than faithful, safe imitation. Mode coverage, however, is not a necessary condition for success: Sinkhorn-CWAE collapses to a single mode during training and still retains $85\%$ success. On the other hand, on the estimated nearly unimodal simulation data, deterministic regression remains competitive, while multimodal-capable policies can improve or worsen performance, depending on their training dynamics, and generally increase inference latency. 
This is consistent with concurrent work~\cite{pan2026much}, which attributes benefits of flow-based
policies on common BC benchmarks primarily to supervised iterative
computation and training-time noise injection rather than the representation of
multimodal action distributions supposedly present in the data.

%% file: sections/06_conclusion.tex
\section{Conclusion}
\label{sec:conclusion}
In this work, we formalize multimodality in imitation learning and show that, for both CVAE-based and action-space generative policies, mode preservation is governed by explicit quantities. For CVAEs, the key factor is posterior-prior regularization: stronger regularization improves alignment but can remove the action-conditioned information needed to recover distinct modes. For action-space generative policies with deterministic samplers, the key factor is the Lipschitz constant of the base-to-action map: excessive smoothness prevents assigning significant mass to many well-separated modes. To further emphasize the practical relevance of our analysis, we note that the default choice of $\beta$ is 10 in ACT ~\citep{zhao2023learning}. Our analysis reveals that this choice is likely to produce unimodal policies. We empirically validate this in Section \ref{par:kl-collapse}. 
On the simulation benchmarks, our diagnostics detect little
conditional multimodality, and deterministic regression remains competitive. These findings suggest that representing multiple
action modes may not be the main bottleneck for policy performance
in these settings.
\paragraph{Limitations and future work.}
Current limitations are that our results are training-independent: Corollary~\ref{cor:beta-critical} does not ensure that training finds an optimizer $(\hat{\theta},\hat{\phi},\hat{\psi})$ with the desired multimodal properties, while Proposition~\ref{prop:multflow} only identifies a structural limitation of low-Lipschitz samplers. Since common regularizers penalize Lipschitz behavior, they may bias policies toward mode collapse.
Future work will examine how to estimate the number of modes for datasets where only visual features are available, develop training procedures that balance sampler smoothness and mode preservation, and understand whether the limited conditional multimodality observed in
robotic simulation benchmarks also characterizes large-scale real-robot
datasets.

%% file: sections/07_ack.tex
\section{Acknowledgements}

Lorenzo Mazza, Massimiliano Datres, Gitta Kutyniok and Stefanie Speidel acknowledge support by the project ”Next Generation AI Computing (gAIn),” funded by the Bavarian Ministry of Science and the Arts and the Saxon Ministry for Science, Culture, and Tourism as well as by the Hightech Agenda Bavaria. 

Gitta Kutyniok is also grateful for partial support from the Konrad Zuse School of Excellence in Reliable AI (DAAD). 

Additionally, Gitta Kutyniok and Massimiliano Datres acknowledge support by the Munich Center for Machine Learning (MCML). 

Gitta Kutyniok furthermore acknowledges support by the German Research Foundation under Grants DFG-SPP-2298, KU 1446/31-1 and KU 1446/32-1, and by the Bavarian Ministry for Digital Affairs.

Ariel Rodriguez, Sebastian Bodenstedt and Stefanie Speidel acknowledge the support by the German Research Foundation (DFG, Deutsche Forschungsgemeinschaft) as part of Germany’s Excellence Strategy – EXC 2050/1 – Project ID 390696704 – Cluster of Excellence “Centre for Tactile Internet with Human-in-the-Loop” (CeTI) of Technische Universität Dresden.

Ariel Rodriguez and Stefanie Speidel would additionally like to thank the Federal Ministry of Research, Technology, and Space (BMFTR) for its support as part of the research program Communication Systems “Souverän. Digital. Vernetzt.”.Joint project 6G-life, project identification number: 16KIS2413K. 

This work has been partially supported by the German Federal Ministry of Research, Technology and Space (BMFTR) under the Robotics Institute Germany (RIG).

%% file: sections/A_appendix.tex
\section{Notation, Definitions, and Known Results}\label{app:notation}
We denote generic spaces with calligraphic letters, random variables with
uppercase letters, their realizations with lowercase letters, and fixed
problem constants with sans-serif capital letters. Generic measurable
sets are denoted by uppercase letters such as \(B,C\). Given a space $\mathcal{X}$, \(\mathcal X^{\mathsf T}\) denotes the space of length-\(\mathsf T\) sequences with entries in a generic space \(\mathcal X\), namely \(x=(x_1,\dots,x_{\mathsf T})\) with $x_i\in\mathcal{X}$. Event
probabilities are written as \(\Pr(\cdot)\) while expectations are denoted with $\mathbb{E}$. We use subscripts to indicate the random variable with respect to which $\Pr$ and $\mathbb{E}$ are taken whenever this is not clear from the context. Let \((\mathcal A,d_{\mathcal A})\) be a metric space. For \(B,C\subseteq\mathcal A\), their distance is
\[
\operatorname{dist}_{\mathcal A}(B,C)
=
\inf_{b\in B,\;c\in C} d_{\mathcal A}(b,c).
\]
When \(\mathcal A \subseteq \mathbb R^{d_a}\), we set $d_{\mathcal A}(a,a') = \|a-a'\|_2$. For notational simplicity, we denote the Euclidean norm simply by \(\|\cdot\|\). All logarithms are natural, so information quantities are measured in
nats. 
The support of a random variable \(X\), denoted by \(\operatorname{supp}(X)\), is the support of its law \(\mu_X\). In particular, in the discrete case, $\operatorname{supp}(X)=\{x : p_X(x)>0\}$.

We write \(X \sim \mu\) to mean that the random variable \(X\) has distribution \(\mu\); that is, for every measurable set \(B\),
$\Pr(X \in B) = \mu(B)$. When \(\mu\) is absolutely continuous with respect to the Lebesgue measure, we denote its density by \(p(x)\). Equivalently, in this case we may write \(X \sim p\) meaning that $\mu(B)=\int_B p(x)dx$. If the support of $\mu$ is discrete, then $\mu$ does not admit a density with respect to the Lebesgue measure. In this case, writing $\operatorname{supp}(\mu)=\{x_i\}_{i\in I}$ where \(I\) is finite or countable, we identify \(\mu\) with its probability mass function
\[
p(x_i) = \mu(\{x_i\}), \qquad i\in I.
\]
Equivalently, in the sense of measures, one may write $\mu = \sum_{i\in I} p_\mu(x_i)\,\delta_{x_i}$, where \(\delta_{x_i}\) denotes the Dirac measure at \(x_i\). We write \(\mu \ll \nu\) to mean that \(\mu\) is absolutely continuous with respect to \(\nu\), namely, for every measurable set \(A\), \(\nu(A)=0\) implies \(\mu(A)=0\).
We denote by \(\mu_{X,Y}\) the joint law of a pair of random variables \((X,Y)\), by \(\mu_{X\mid Y=y}\) the conditional law of \(X\) given \(Y=y\), and by \(p_{X,Y}\) and \(p_{X\mid Y}(\cdot\mid y)\) their corresponding densities, with
\[
p_{X\mid Y}(x\mid y)=\frac{p_{X,Y}(x,y)}{p_Y(y)}
\qquad \text{for }p_Y(y)>0.
\]

We will measure the discrepancy between probability measures using the Kullback-Leibler divergence, defined as follows.
\begin{definition}[Kullback-Leibler divergence]
Let \(X\sim \mu\) and \(Y \sim \nu\) be random variables such that \(\mu \ll \nu\). The Kullback--Leibler divergence from \(\nu\) to \(\mu\) is defined as
\[
\KL(\mu\|\nu)
=
\int \log\!\left(\frac{d\mu}{d\nu}\right)\,d\mu .
\]
\end{definition}


We define next the notions of Shannon entropy, differential entropy,
conditional entropy, and conditional mutual information. For a discrete
random variable, the Shannon entropy \(H(X)\) measures the uncertainty of
the probability mass assigned to its possible values. For a continuous
random variable, the differential entropy \(h(X)\) measures the spread of
its density. Conditional entropy measures the remaining uncertainty after
observing another variable. Conditional mutual information
\(I(X;Y\mid Z)\) measures how much information \(X\) and \(Y\) share once
\(Z\) is known.

\begin{definition}[Entropy and conditional entropy]
Let \(X\sim \mu\), $Y\sim \nu $ be random variables. The entropy of \(X\) is defined as
\[
H(X)
:=
-\int\log p_X(x)\,\mu(dx),
\]
with the convention \(0\log 0=0\).
The conditional entropy of \(X\) given \(Y\) is defined as
\[
H(X\mid Y) := -\int \log p_{X\mid Y}(x\mid y)\, \mu_{X,Y}(dx, dy) \, .
\]
\end{definition}

We note that, for any random variable \(X\) supported on a finite set
\(\mathcal X\),
\[
H(X)\le \log |\mathcal X|,
\]
with equality if and only if \(X\) is uniformly distributed on
\(\mathcal X\). For \(p\in[0,1]\), we denote by \(H_b(p)\) the binary
entropy function,
\[
H_b(p)
=
-p\log p-(1-p)\log(1-p).
\]
Equivalently, if \(X\sim\operatorname{Ber}(p)\), then \(H(X)=H_b(p)\).

Mutual information provides a measure of statistical dependence between random variables, quantifying how much knowing one variable reduces uncertainty about another.
\begin{definition}[Conditional mutual information]
\label{def:mutual}
For random variables \(X,Y,Z\), the \emph{conditional mutual information}
between \(X\) and \(Y\) given \(Z\) is
\[
I(X;Y\mid Z)
:=
H(X\mid Z)-H(X\mid Y,Z),
\]
whenever \(X\) is discrete and the conditional entropy terms are well
defined. If \(X\) is continuous, the same identity holds with
differential entropies \(h(\cdot)\) in place of \(H(\cdot)\), whenever
the terms are well defined.
When \(X,Y,Z\) are continuous and admit densities, this is equivalently
\[
I(X;Y\mid Z)
=
\int p(x,y,z)
\log
\frac{p(x,y\mid z)}
{p(x\mid z)p(y\mid z)}
\,dx\,dy\,dz .
\]
\end{definition}

Two standard properties of mutual information that will be used
throughout are the following:
\begin{enumerate}
    \item[(P1)] For random variables \(X,Y\), mutual information is
    symmetric:
    \[
    I(X;Y)=I(Y;X).
    \]

    \item[(P2)] For random variables \(X,Y,Z\), conditional mutual
    information satisfies
    \[
    I(X;Y\mid Z)
    =
    H(X\mid Z)-H(X\mid Y,Z),
    \]
    whenever \(X\) is discrete and the entropy terms are well defined.
    For continuous \(X\), the analogous identity holds with differential
    entropies.
\end{enumerate}

Of particular interest in this work are Markov chains of random variables.
\begin{definition}[Markov chain]
\label{def:MC}
Let \(p\) be a joint distribution on
\(\mathcal X\times\mathcal Y\times\mathcal Z\). We write
\(X\to Y\to Z\) if \(X\) and \(Z\) are conditionally independent given
\(Y\), denoted by \(X\perp Z\mid Y\). Equivalently, the conditional
distribution of \(Z\) given \((X,Y)\) depends only on \(Y\), that is
\[
p(z\mid x,y)
=
p(z\mid y)
\qquad
p\text{-a.s.}
\]
\end{definition}
A fundamental standard consequence of the Markov-chain structure is the following inequality, called data-processing inequality.
\begin{lemma}[Data-processing inequality]
\label{lem:dpi}
Let \((X,Y,Z)\sim p\) such that \(X\to Y\to Z\) forms a Markov chain.
Then
\[
I(X;Z)\le I(X;Y).
\]
\end{lemma}
Fano’s inequality relates the probability of incorrectly estimating a discrete random variable to its remaining conditional entropy given the estimator.

\begin{lemma}[Fano's inequality]
\label{lem:fano}
Let \(M\) be a random variable taking values in a finite set
\( C\) with \(|C|=\mathsf K\), where
\(\mathsf K>1\). Let \(\hat M\) be an estimator of \(M\), that is, a
random variable taking values in \(C\). Define the probability
of error
\[
e := \Pr(\hat M\neq M).
\]
Then
\[
H(M\mid \hat M)
\le
H_b(e)+e\log(\mathsf K-1),
\]
where
\[
H_b(e)
=
-e\log e-(1-e)\log(1-e)
\]
is the binary entropy function.
\end{lemma}

\section{Proofs}
\label{app:proofs}
\subsection{Latent-Variable Information Bounds and Collapse}
\label{app:latent-info-collapse}

We first prove the latent-variable results from
\S\ref{sec:theory-latent}. The subsection has three parts. We prove the
mode-information lower bound, then specialize it to the uniform
exact-recovery case, and finally derive the objective-level collapse
criterion for pointwise-KL regularization.
Let
\[
q_{\phi,\mathcal D}(s,a,z)
:=
p_{\mathcal D}(s,a)\,q_\phi(z\mid a,s)
=
p_{\mathcal D,S}(s)\,
p_{\mathcal D}(a\mid s)\,
q_\phi(z\mid a,s)
\]
denote the posterior-encoder-induced joint distribution over
\((S,A,Z)\). Conditioning the joint law \(q_{\phi,\mathcal D}\) on \(S=s\) gives
\[
q_{\phi,\mathcal D}(a,z\mid S=s)
=
p_{\mathcal D}(a\mid s)\,q_\phi(z\mid a,s).
\]
Throughout the subsection, all statewise probabilities, entropies, and mutual informations below are taken under this conditional law.

\paragraph{Proof of Proposition \ref{prop:mode-info}}
We prove Proposition~\ref{prop:mode-info}. The proof rests on two Markov chains, \(M\to A\to Z\) and \(M\to Z\to \hat M\), each yielding a data-processing inequality. Combining them with Fano's inequality gives the bound. We now make this precise.
Fix \(s\in\operatorname{supp}(p_{\mathcal D,S})\) with
\(\mathsf K(s)>1\). Under the conditional law at \(S=s\), we have
\[
A\sim p_{\mathcal D}(\cdot\mid s),
\qquad
Z\sim q_\phi(\cdot\mid A,s),
\qquad
M=g(A,s),
\qquad
\hat A=f_\theta(Z,s).
\]
Let \(\hat M\) be the mode label induced by \(\hat A\): if \(\hat A\in C_j(s)\), set \(\hat M=j\); if \(\hat A\notin\bigcup_{j=1}^{\mathsf K(s)} C_j(s)\), assign \(\hat M\) arbitrarily in \(\{1,\ldots,\mathsf K(s)\}\).
We start by noticing that since \(M=g(A,s)\) is a deterministic function of \(A\) at fixed \(s\), the conditional law of \(M\) given \(A=a\) is degenerate:
\[
q_{\phi,\mathcal D}(M=m\mid A=a,S=s)
=
\mathbf 1_{\{m=g(a,s)\}}.
\]
Moreover, \(Z\) is sampled from \(q_\phi(\cdot\mid A,s)\), so conditioning additionally on \(M\) does not change its law:
\[
q_{\phi,\mathcal D}(z\mid A=a,M=m,S=s)
=
q_{\phi,\mathcal D}(z\mid A=a,S=s)
=
q_\phi(z\mid a,s).
\]
Therefore, at fixed \(s\), we have the Markov chain
\[
M\to A\to Z .
\]

Since \(M=g(A,s)\) is a deterministic function of \(A\), we also have
\(Z\to A\to M\) at fixed \(s\). By data processing,
\[
I(Z;M\mid S=s)\le I(Z;A\mid S=s).
\]
By symmetry of mutual information, this gives
\begin{equation}
\label{eq:1}
I(M;Z\mid S=s)\le I(A;Z\mid S=s).
\end{equation}

Similarly, at fixed \(s\), \(\hat M\) is a deterministic function of
\(Z\), hence
\[
M\to Z\to \hat M .
\]
Therefore, by Lemma~\ref{lem:dpi}, it holds
\begin{equation}
\label{eq:2}
I(M;\hat M\mid S=s)
\le
I(M;Z\mid S=s).
\end{equation}

Combining \eqref{eq:1} and \eqref{eq:2}, it holds
\begin{equation}\label{eq:3}
    I(A;Z\mid S=s)\ge I(M;Z\mid S=s) \ge I(M;\hat M\mid S=s).
\end{equation}

Let
\[
e(s):=\Pr(\hat M\neq M\mid S=s)
\]
be the conditional decoded-mode error. By construction of \(\hat M\), see \eqref{eq:modeassdef}, if
\(\hat A\in C_M(s)\), then \(\hat M=M\). Hence
\[
\{\hat M\neq M\}
\subseteq
\{\hat A\notin C_M(s)\}
=
\{f_\theta(Z,s)\notin C_M(s)\}.
\]
Therefore, by the definition of \(\rho(s)\) in
\eqref{eq:assprop},
\[
e(s)
=
\Pr(\hat M\neq M\mid S=s)
\le
\Pr\!\left(
f_\theta(Z,s)\notin C_M(s)
\,\middle|\,
S=s
\right)
=
\rho(s).
\]

Given \(s\), since \(M\) takes values in \(\{1,\ldots,\mathsf K(s)\}\), Lemma~\ref{lem:fano} gives, for \(\mathsf K(s)>1\),
\begin{align}
\label{eq:usingfano}
H(M\mid \hat M,S=s)
&\le
H_b(e(s))+e(s)\log(\mathsf K(s)-1) \\
&\le
H_b(\rho(s))+\rho(s)\log(\mathsf K(s)-1).
\notag
\end{align}
The last inequality follows from \(e(s)\le\rho(s)\) and from the fact that, if \(\rho(s)\le 1-1/\mathsf K(s)\), then
\[
p\mapsto H_b(p)+p\log(\mathsf K(s)-1)
\]
is non-decreasing on \([0,\rho(s)]\).

Combining (P2), \eqref{eq:3}, and \eqref{eq:usingfano}, we obtain
\begin{align*}
I(A;Z\mid S=s)
&\ge I(M;\hat M\mid S=s) \\
&= H(M\mid S=s)-H(M\mid \hat M,S=s) \\
&\ge H(M\mid S=s)
- H_b(\rho(s))
- \rho(s)\log(\mathsf K(s)-1).
\end{align*}
Averaging over \(S\sim p_{\mathcal D,S}\) gives
\[
I(A;Z\mid S)\ge B_\rho,
\]
which proves the first claim of Proposition~\ref{prop:mode-info}.

By the definition of \(B_\rho\), this gives
\[
I(A;Z\mid S)\ge B_\rho .
\]

Recall that \(q_{\phi,\mathcal D}(s,a,z)\) denotes the posterior-encoder-induced joint distribution and $q_{\phi,\mathcal D}(z\mid s)
$ is the state-conditional aggregated posterior. The pointwise latent regularizer decomposes into a mutual-information term between the action and latent code, and a mismatch term between the aggregated posterior and the prior:
\begin{equation}
\E_{(S,A)\sim p_{\mathcal D}} \KL\!\left(q_\phi(\cdot\mid A,S)\,\|\, 
p_\psi(\cdot\mid S) \right)
=
I(A;Z\mid S)
+
\E_{S\sim p_{\mathcal D,S}}
\KL\!\left( q_{\phi,\mathcal D}(\cdot\mid S) \,\|\, 
p_\psi(\cdot\mid S)
\right).
\label{eq:agg_decomp}
\end{equation}
The full derivation of this decomposition can be found, for example, in \citet{hoffman2016elbosurgery}.
The second term in \eqref{eq:agg_decomp} is nonnegative. Hence
\[
\E_{(S,A)\sim p_{\mathcal D}}
\KL\!\left(
q_\phi(\cdot\mid A,S)
\,\|\,
p_\psi(\cdot\mid S)
\right)
\ge
I(A;Z\mid S)
\ge
B_\rho\ ,
\]
which concludes the proof.
\qed

\paragraph{Proof of Corollary~\ref{cor:logk}.}

Using \(\rho(s)=0\) for \(p_{\mathcal D,S}\)-almost every \(s\), we have
\[
H_b(\rho(s))+\rho(s)\log(\mathsf K(s)-1)=0
\]
for every \(s\) with \(\mathsf K(s)>1\), where we recall that \(H_b(p)=-p\log p-(1-p)\log(1-p)\) and the usual convention \(H_b(0)=0\).

Moreover, since
\(M\mid S=s\sim\operatorname{Unif}(\{1,\ldots,\mathsf K(s)\})\), we have
\[
\E_{S\sim p_{\mathcal D,S}}[H(M\mid S)]
=
\E_{S\sim p_{\mathcal D,S}}[\log \mathsf K(S)].
\]

These two observations, combined with Proposition~\ref{prop:mode-info}, imply
\[
I(A;Z\mid S)
\ge
\E_{S\sim p_{\mathcal D,S}}\log \mathsf K(S).
\]
\qed

\paragraph{Proof of Corollary~\ref{cor:beta-critical}}
Let us denote with
\[
\hat\rho(s) :=
\Pr\!\left(
f_{\hat\theta}(Z,s)\notin C_M(s)
\,\middle|\,
S=s
\right).
\]
Assume that
\[
\hat\rho(s)\le 1-\frac{1}{\mathsf K(s)}
\]
for all \(s\in\operatorname{supp}(p_{\mathcal D,S})\) with
\(\mathsf K(s)>1\). By \textit{(H1)} and the optimality of
\((\hat\theta,\hat\phi,\hat\psi)\), we have
\begin{align}
\label{eq:11}
\hat{\mathcal L}^{\beta}
=
\mathcal L^\beta_{\mathrm{var}}(\hat\theta,\hat\phi,\hat\psi)
&\le
\mathcal L^\beta_{\mathrm{var}}(\theta_0,\phi_0,\psi_0) \\
&=
\mathcal L^0_{\mathrm{var}}(\theta_0,\phi_0,\psi_0)
=:
\mathsf C ,
\notag
\end{align}
where the equality follows from
\(\mathit D(q_{\phi_0},p_{\psi_0})=0\), and \(\mathsf C>0\) is
independent of \(\beta\).

By (H2) and Proposition \ref{prop:mode-info}, it holds that 
\begin{align}\label{eq:12}
    I(A;Z\mid S)\ge B_{\hat\rho},
\end{align}

On the other hand, by the decomposition in \eqref{eq:agg_decomp}, applied
to \((\hat\phi,\hat\psi)\), we have
\begin{align} 
\label{eq:11}
I(A;Z\mid S)
&=
\E_{(S,A)\sim p_{\mathcal D}}
\KL\!\left(
q_{\hat\phi}(\cdot\mid A,S)
\,\|\,
p_{\hat\psi}(\cdot\mid S)
\right)
-
\E_{S\sim p_{\mathcal D,S}}
\KL\!\left(
q_{\hat\phi,\mathcal D}(\cdot\mid S)
\,\|\,
p_{\hat\psi}(\cdot\mid S)
\right) \notag \\
&\le
\E_{(S,A)\sim p_{\mathcal D}}
\KL\!\left(
q_{\hat\phi}(\cdot\mid A,S)
\,\|\,
p_{\hat\psi}(\cdot\mid S)
\right)
=
\mathit D(q_{\hat\phi},p_{\hat\psi}).
\end{align}
Since the reconstruction loss is nonnegative, the objective satisfies
\[
\hat{\mathcal L}^{\beta}
=
\mathcal L_{\rm rec}(\hat\theta,\hat\phi)
+
\beta\,\mathit D(q_{\hat\phi},p_{\hat\psi})
\ge
\beta\,\mathit D(q_{\hat\phi},p_{\hat\psi}).
\]
Together with \(\hat{\mathcal L}^{\beta}\le \mathsf C\), this gives
\[
\beta\,\mathit D(q_{\hat\phi},p_{\hat\psi})
\le
\mathsf C.
\]

Therefore,
\[
I(A;Z\mid S)
\le
\mathit D(q_{\hat\phi},p_{\hat\psi})
\le
\frac{\mathsf C}{\beta}.
\]
Combining this with \eqref{eq:12} gives
\[
B_{\hat\rho}
\le
I(A;Z\mid S)
\le
\frac{\mathsf C}{\beta}.
\]
Consequently, if \(B_{\hat\rho}>\mathsf b_0\), then it follows trivially that
\[
\beta < \frac{\mathsf C}{\mathsf b_0}.
\]

\qed

\subsection{Aggregate Matching}
\label{app:proofs-aggregate}
We next show why aggregate matching can preserve mode information even
when pointwise KL would penalize it. The key observation is that the
aggregated posterior is a mixture over mode-conditioned latent
distributions.

We write \(\mathcal Z:=\mathbb R^{\mathsf d_z}\) for the latent space.

\begin{proposition}[Aggregate matching can preserve mode information]
\label{prop:aggregate-matching}
Under the setting introduced above, fix
\(s\in\operatorname{supp}(p_{\mathcal D,S})\). For each mode index
\(k\in\{1,\ldots,\mathsf K(s)\}\) with
\(p_{\mathcal D}(M=k\mid S=s)>0\), define
\[
q_k(\cdot\mid s)
:=
q_{\phi,\mathcal D}(\cdot\mid M=k,S=s).
\]
That is, \(q_k(\cdot\mid s)\) is the latent distribution induced by the
posterior encoder when the demonstrated action belongs to mode \(k\).
Then the state-conditioned aggregated posterior is
\[
q_{\phi,\mathcal D}(\cdot\mid S=s)
=
\sum_{k=1}^{\mathsf K(s)}
p_{\mathcal D}(M=k\mid S=s)\,q_k(\cdot\mid s).
\]
If the prior satisfies
\[
p_\psi(\cdot\mid s)
=
\sum_{k=1}^{\mathsf K(s)}
p_{\mathcal D}(M=k\mid S=s)\,q_k(\cdot\mid s),
\]
then the aggregated posterior--prior mismatch is zero. Moreover, if the
supports of the components \(q_k(\cdot\mid s)\) are mutually disjoint,
then there exists a deterministic decoder
\(h_s:\mathcal Z\to\{1,\ldots,\mathsf K(s)\}\) such that
\[
\Pr(h_s(Z)=M\mid S=s)=1,
\]
and therefore
\[
I(M;Z\mid S=s)
=
H(M\mid S=s).
\]
\end{proposition}

\paragraph{Proof.}
We first compute the state-conditional aggregated posterior and then
show that disjoint latent supports make the mode recoverable from \(Z\).

By the law of total probability over the discrete mode label \(M\),
\[
q_{\phi,\mathcal D}(\cdot\mid S=s)
=
\sum_{k=1}^{\mathsf K(s)}
p_{\mathcal D}(M=k\mid S=s)\,
q_{\phi,\mathcal D}(\cdot\mid M=k,S=s).
\]
By definition,
\[
q_k(\cdot\mid s)
=
q_{\phi,\mathcal D}(\cdot\mid M=k,S=s),
\]
and hence
\[
q_{\phi,\mathcal D}(\cdot\mid S=s)
=
\sum_{k=1}^{\mathsf K(s)}
p_{\mathcal D}(M=k\mid S=s)\,q_k(\cdot\mid s).
\]
Thus, if
\[
p_\psi(\cdot\mid s)
=
\sum_{k=1}^{\mathsf K(s)}
p_{\mathcal D}(M=k\mid S=s)\,q_k(\cdot\mid s),
\]
then
\[
q_{\phi,\mathcal D}(\cdot\mid S=s)=p_\psi(\cdot\mid s),
\]
so the aggregated posterior--prior mismatch is zero.

Recall now that, by definition of \(q_k(\cdot\mid s)\), for each
\(k\in\{1,\ldots,\mathsf K(s)\}\) with
\(p_{\mathcal D}(M=k\mid S=s)>0\), we have
\[
Z\mid(M=k,S=s)\sim q_k(\cdot\mid s).
\]
Let
\[
E_k(s):=\operatorname{supp} q_k(\cdot\mid s).
\]
If the sets \(E_1(s),\ldots,E_{\mathsf K(s)}(s)\) are mutually disjoint, define \(h_s:\mathcal Z\to\{1,\ldots,\mathsf K(s)\}\) by \(h_s(z)=k\) whenever \(z\in E_k(s)\), with arbitrary values outside \(\bigcup_k E_k(s)\). Since \(Z\mid(M=k,S=s)\) is supported on \(E_k(s)\), it follows that
\[
\Pr(h_s(Z)=M\mid S=s)=1.
\]
Therefore, at fixed \(s\), the mode label \(M\) is determined by \(Z\), and hence
\[
H(M\mid Z,S=s)=0.
\]
By Definition~\ref{def:mutual}, it follows trivially that
\[
I(M;Z\mid S=s)
=
H(M\mid S=s).
\]
This concludes the proof.
\qed

\begin{remark}[Deployment-time prior coverage]
Proposition~\ref{prop:aggregate-matching} shows that the prior can match the aggregated posterior while the latent variable still identifies the demonstrated mode, avoiding the pointwise-KL pressure on
\(I(A;Z\mid S)\). The limitation is that aggregate matching constrains only the state-conditioned marginal
\[
q_{\phi,\mathcal D}(\cdot\mid S=s)
=
\int q_\phi(\cdot\mid a,s)\,p_{\mathcal D}(a\mid s)\,da,
\]
and not each action-conditioned posterior \(q_\phi(\cdot\mid a,s)\). Thus two encoder families can have identical, or very similar, aggregated marginals while having very different conditionals \(q_\phi(\cdot\mid
a,s)\). Consequently, matching \(p_\psi(\cdot\mid s)\) to the aggregated
posterior does not by itself guarantee that prior samples decode to valid
actions.

We can see this by defining, for a fixed observation \(s\) and mode index
\(k\in\{1,\ldots,\mathsf K(s)\}\), the decoder-induced latent region
\[
R_k(s)
:=
\{z\in\mathcal Z:\ f_\theta(z,s)\in C_k(s)\}.
\]
Deployment covers mode \(k\) only if \(p_\psi(R_k(s)\mid s)\) is sufficiently large. Thus aggregate matching must align the prior with the posterior mass on the decoder-induced regions \(\{R_k(s)\}_{k=1}^{\mathsf K(s)}\), and not only match aggregate latent marginals in regions irrelevant to the decoder.
\end{remark}

\subsection{Action-Space Generative Policies}
\paragraph{Proof of Proposition \ref{prop:multflow}}
The proof relies on the Gaussian isoperimetric inequality from \citet{sudakov1978extremal} to avoid dependency on the dimension $d$ of the base space of $U$. We will denote with $\gamma_d$ the standard Gaussian measure on $\mathbb R^d$,
so that, for any measurable set $A\subseteq\mathbb R^d$,
\[
\gamma_d(A)
:=
\Pr_{U\sim\mathcal N(0,I_d)}(U\in A).
\]

Let us fix \(s\in\mathcal S\), \(\tau\in(0,1)\), and a parameter vector \(\theta\). 
For each $k\in\{1,\ldots,K\}$, define the set of points in the base space that the generator
$G_{\theta,s}$ maps into action mode $C_k(s)$:
    $$
    A_k
      := G_{\theta,s}^{-1}\bigl(C_k(s)\bigr)
      = \bigl\{u\in\R^d : G_{\theta,s}(u)\in C_k(s)\bigr\}.
    $$ Let $k\neq k'$. If $A_k=\varnothing$ or $A_{k'}=\varnothing$, then by definition $dist(A_k,A_{k'})=\infty.$
If $A_k,A_{k'}\neq\varnothing$, let $u\in A_k$ and $u'\in A_{k'}$. By definition, $G_{\theta,s}(u)\in C_k(s)$ and $G_{\theta,s}(u')\in C_{k'}(s)$.
Hence
\begin{equation*}
  \dist\bigl(G_{\theta,s}(u),G_{\theta,s}(u')\bigr)
  \geq \dist\bigl(C_k(s),C_{k'}(s)\bigr)
  \geq \Delta(s).
\end{equation*}
But, by the $L_{\theta, s}$-Lipschitzianity of $G_{\theta, s}$, it follows that
$$
  \dist\bigl(G_{\theta,s}(u),G_{\theta,s}(u')\bigr)
  \leq L_{\theta,s}\lVert u-u'\rVert,
$$
Therefore, combining the bounds, it holds
\begin{equation*}
  \lVert u-u'\rVert
  \geq \frac{\Delta(s)}{L_{\theta,s}}.
\end{equation*}
Taking the infimum over $u,u'$ proves that
\begin{equation}
\label{eq:multflow-preimage-separation}
\dist(A_k,A_{k'})
=
\inf_{\substack{u\in A_k\\u'\in A_{k'}}}
\lVert u-u'\rVert
\ge \frac{\Delta(s)}{L_{\theta,s}},
\qquad k\neq k'.
\end{equation}

To avoid a dimension-dependent bound, we introduce the following extended set. Let $0<r'<\frac{\Delta(s)}{2L_{\theta,s}}$ and defined 
$$
A_k^{r'} := \{u\in \R^d: \ \operatorname{dist}(u, A_k) \le r'\}
$$
where $\operatorname{dist}(u, A) = \min_{v\in A_k} \, \|u-v\|_2$. Then, it is easy to see that \begin{equation}
\label{eq:multflow-disjoint-enlargements}
A_k^{r'}\cap A_{k'}^{r'}=\varnothing,
\qquad k\neq k',
\qquad 0<r'<\frac{\Delta(s)}{2L_{\theta,s}},
\end{equation} since $\dist(A_k,A_{k'})\ge \frac{\Delta(s)}{L_{\theta,s}} $. 

By the standard Gaussian isoperimetric inequality, it holds
\begin{equation}
\label{eq:multflow-gaussian-enlargement}
  \gamma_d\bigl(A_k^{r'}\bigr)
  \geq
  \Phi\bigl(\Phi^{-1}(\pi_\theta(C_k(s)|S=s))+r'\bigr),
  \qquad k=1,\ldots,K.
\end{equation}
Therefore, since $A_k^{r'}\cap A_{k'}^{r'}=\varnothing$ for $k\neq k'$,
\begin{equation*}
  \sum_{k=1}^{K}\Phi\bigl(\Phi^{-1}(\pi_\theta(C_k(s)|S=s))+r'\bigr)
  \leq \sum_{k=1}^{K}\gamma_d\bigl(A_k^{r'}\bigr)
  = \gamma_d\!\left(\bigcup_{k=1}^{K}A_k^{r'}\right)
  \leq 1.
\end{equation*}

By continuity of $\Phi$,
\begin{equation*}
  \sum_{k=1}^{K}\Phi\!\left(\Phi^{-1}(\pi_\theta(C_k(s)|S=s))+\frac{\Delta(s)}{2L_{\theta,s}}\right)
  = \lim_{r'\to \left(\frac{\Delta(s)}{2L_{\theta,s}}\right)^-}F(r')
  \leq 1.
\end{equation*}

Let us recall the definition of number of $\tau$-represented modes by the sampler as:
\begin{equation*}
  N_\theta^{(\tau)}(s)=|\left\{ k \in \{1,\ldots,K(s)\} :
\pi_{\theta}\bigl(C_k(s)\mid s\bigr) > \tau \right\}|.
\end{equation*}
By monotonicity of $\Phi^{-1}$, for all $k\in \left\{ k \in \{1,\ldots,K(s)\} :
\pi_{\theta}\bigl(C_k(s)\mid s\bigr) > \tau\right\}$,  it holds that 
$$
\Phi^{-1}(\pi_{\theta}\bigl(C_k(s)\mid s\bigr))\geq\Phi^{-1}(\tau)
$$
and hence
\begin{equation*}
  \Phi\bigl(\Phi^{-1}(\pi_{\theta}\bigl(C_k(s)\mid s\bigr))+\frac{\Delta(s)}{2L_{\theta,s}}\bigr)
  \geq
  \Phi\bigl(\Phi^{-1}(\tau)+\frac{\Delta(s)}{2L_{\theta,s}}\bigr).
\end{equation*}
Summing over $k$ gives
\begin{align*}
  1
  &\geq \sum_{k=1}^{K(s)}\Phi\bigl(\Phi^{-1}(\pi_{\theta}\bigl(C_k(s)\mid s\bigr))+\frac{\Delta(s)}{2L_{\theta,s}}\bigr) \\
  &\geq \sum_{k\in \left\{ k \in \{1,\ldots,K(s)\} :
\pi_{\theta}\bigl(C_k(s)\mid s\bigr) > \tau\right\}}\Phi\bigl(\Phi^{-1}(\tau)+\frac{\Delta(s)}{2L_{\theta,s}}\bigr) \\
  &\geq N_\theta^{(\tau)}(s)\,\Phi\bigl(\Phi^{-1}(\tau)+\frac{\Delta(s)}{2L_{\theta,s}}\bigr)  \\
  & \geq N_\theta^{(\tau)}(s) \frac{\sqrt{2\pi}\tau \, \Delta(s)/2L_{\theta, s} }{\sqrt{2\pi}\tau \Delta(s)/2L_{\theta, s}  + \sqrt{2\pi}|\Phi^{-1}(\tau)| +1 } \, ,
\end{align*}
where in the last step we have used the fact that, for all $\tau>0$ and $r>0$, it holds 
\begin{align*}
    \Phi(\Phi^{-1}(\tau) + r) \ge \frac{\sqrt{2\pi}\tau r}{\sqrt{2\pi}\tau r + \sqrt{2\pi}|\Phi^{-1}(\tau)| + 1} \, .
\end{align*}
Inverting the previous equation, we conclude that 
$$
N_\theta^{(\tau)}(s)\le 1 + \Big \lfloor \frac{2(\sqrt{2\pi} |\Phi^{-1}(\tau)|+1)}{\sqrt{2\pi}\tau} \frac{L_{\theta, s}}{\Delta_s}\Big \rfloor \, ,
$$ which concludes the proof.
\hfill$\square$
\paragraph{Corollary: stretch-bridge trade-off}
The argument underlying Proposition~\ref{prop:multflow} also yields
a trade-off between the generator's Lipschitz constant and the bridge mass, i.e., the
probability assigned outside the action modes.

\begin{corollary}
\label{cor:stretch-bridge-tradeoff}
Under the assumptions of Proposition~\ref{prop:multflow}, define
the bridge mass at state $s$ as
\[
B_\theta(s)
:=
\pi_\theta\!\left(
\mathcal A\setminus\bigcup_{k=1}^{\mathsf K(s)}C_k(s)
\,\middle|\,s
\right).
\]
Then,
\[
L_{\theta,s}\,B_\theta(s)
\ge
\bigl(N_\theta^{(\tau)}(s)-1\bigr)
\frac{\Delta(s)}{2}\,
\varphi\!\left(\Phi^{-1}(\tau)\right),
\]
where $\varphi(x)=(2\pi)^{-1/2}e^{-x^2/2}$ is the standard
normal density.
\end{corollary}

\paragraph{Proof of Corollary \ref{cor:stretch-bridge-tradeoff}}
Let us fix $s$. If $N_\theta^{(\tau)}(s)\le1$, the result is immediate.
Otherwise, choose any $\tau$-represented mode with index $k_0$.

For $r'$ as in \eqref{eq:multflow-disjoint-enlargements},
\eqref{eq:multflow-preimage-separation} and
\eqref{eq:multflow-disjoint-enlargements} imply that the added
regions $A_k^{r'}\setminus A_k$ are pairwise disjoint and lie
outside all mode preimages. Thus,
\begin{equation}
\label{eq:bridge-mass-sum}
B_\theta(s)
\ge
\sum_{\substack{
k\neq k_0\\
\pi_\theta(C_k(s)\mid s)>\tau
}}
\left[
\gamma_d(A_k^{r'})-\pi_\theta(C_k(s)\mid s)
\right].
\end{equation}

For each index in this sum, by
\eqref{eq:multflow-gaussian-enlargement} and
\eqref{eq:multflow-disjoint-enlargements},
\[
\Phi\!\left(
\Phi^{-1}(\pi_\theta(C_k(s)\mid s))+r'
\right)
\le \gamma_d(A_k^{r'})
\le 1-\pi_\theta(C_{k_0}(s)\mid s)
<1-\tau.
\]
Since $\pi_\theta(C_k(s)\mid s)>\tau$, it follows that
\[
\Phi^{-1}(\pi_\theta(C_k(s)\mid s))+t
\in
\left[\Phi^{-1}(\tau),-\Phi^{-1}(\tau)\right],
\qquad 0\le t\le r'.
\]
The standard normal density on this interval is at least
$\varphi(\Phi^{-1}(\tau))$. Applying
\eqref{eq:multflow-gaussian-enlargement} therefore gives
\[
\begin{aligned}
\gamma_d(A_k^{r'})-\pi_\theta(C_k(s)\mid s)
&\ge
\int_0^{r'}
\varphi\!\left(
\Phi^{-1}(\pi_\theta(C_k(s)\mid s))+t
\right)\,dt\\
&\ge r'\,\varphi(\Phi^{-1}(\tau)).
\end{aligned}
\]
Substituting into \eqref{eq:bridge-mass-sum}, which contains
$N_\theta^{(\tau)}(s)-1$ terms, yields
\[
B_\theta(s)
\ge
\bigl(N_\theta^{(\tau)}(s)-1\bigr)
r'\,\varphi(\Phi^{-1}(\tau)).
\]
Letting $r'\uparrow\Delta(s)/(2L_{\theta,s})$ and multiplying
by $L_{\theta,s}$ proves the corollary.
\hfill$\square$

\clearpage

%% file: sections/A2_appendix_related_work.tex
\section{Related work}
\label{sec:related}
Multimodal BC has been modeled through several policy families. ACT uses a KL-regularized CVAE with an explicit latent variable to represent trajectory style~\citep{zhao2023learning}. VQ-BeT represents multimodality through learned discrete latent tokens and a two-stage training recipe~\citep{lee2024vqbet,shafiullah2022behavior}. Q-FAT uses continuous Gaussian-mixture heads in an autoregressive transformer~\citep{sheebaelhamd2025quantization}, which is structurally slower than parallel action-chunk decoding. Implicit BC parameterizes the policy as a conditional energy function, requiring sampling or gradient-based optimization at inference rather than a single decoder pass~\citep{florence2021implicit}. More recent methods shift multimodality directly into action-space denoising or flow matching~\citep{chi2025diffusion,reuss2023beso,reuss2024mode,black2024pi0, intelligence2025pi05,shukor2025smolvla}. Across these approaches, comparisons remain largely task-driven and empirical, with limited theoretical guidance on which training quantities affect policies’ ability to represent multimodality.

For one-shot push-forward generators \(A=G_\theta(s,U)\), existing results show that separated multimodal targets force a trade-off between approximation error and generator smoothness~\citep{salmona2022can}. For iterative samplers such as DDIM~\citep{chi2025diffusion} or deterministic flow solvers~\citep{black2024pi0,intelligence2025pi05,shukor2025smolvla}, we study this trade-off for the composed deterministic sampling map induced by the denoising or flow updates. 
A parallel gap appears in latent-based policies: pointwise KL regularization is standard, but its weight is usually treated as an empirical hyperparameter.

%% file: sections/B_appendix.tex
\section{Experimental Details}
\label{app:exps}
This section specifies the policy parameterizations and training
objectives used in the experiments, using the notation introduced in
Appendix~\ref{app:notation}. We write objectives with respect to the population expert state-action law \(p_{\mathcal D}\); in implementation, these objectives are optimized using minibatches \(\{(s_i,a_i)\}_{i=1}^{\mathsf B}\) from the realized dataset \(\tilde{\mathcal D}\). All methods use the same observation encoder and action-chunking interface, and differ only in how they parameterize the conditional distribution over action chunks. We write
\(
\mathcal L_{\rm rec}(\hat a,a)
\)
for the action-chunk reconstruction loss, instantiated as an \(L_1\) loss in all deterministic-decoder latent policies unless stated otherwise. For latent-variable policies, training uses posterior latents
\(Z\sim q_\phi(\cdot\mid A,S)\), while deployment uses prior latents \(Z\sim p_\psi(\cdot\mid S)\).
\subsection{Methods}
\label{app:methods}
We study the following policy parametrizations: deterministic regression (BCAT), 
    KL-regularized conditional variational autoencoder (KL-CVAE), aggregate-matched conditional Wasserstein autoencoder (CWAE), vector-quantized conditional variational autoencoder (Residual VQ-VAE), conditional variational autoencoder with a learned latent flow prior (LAT-Flow), and action-space diffusion/flow-matching-based generative models (Act-Diff, Act-Flow).
\paragraph{BCAT}
\label{app:method-bcat}

The deterministic regression baseline represents the conditional policy by a Dirac
distribution,
\[
\pi_\theta(\cdot\mid s)
=
\delta_{f_\theta(s)} .
\]
It is trained with
\[
\mathcal L_{\rm BC}
=
\E_{(S,A)\sim p_{\mathcal D}}
\left[
\mathcal L_{\rm rec}(f_\theta(S),A)
\right].
\]

\paragraph{KL-CVAE}
\label{app:method-kl-cvae}

KL-CVAE uses a continuous Gaussian posterior, a Gaussian prior, and a
latent-conditioned action decoder. The posterior encoder receives the
demonstrated action chunk together with the observation representation and
outputs a diagonal Gaussian
\[
q_\phi(\cdot\mid a,s)
=
\mathcal N\!\left(
\mu_\phi(a,s),
\operatorname{diag}\exp \ell_\phi(a,s)
\right),
\]
where \(\ell_\phi(a,s)\) denotes the posterior log-variance. Posterior
samples are obtained by reparameterization,
\[
z_q
=
\mu_\phi(a,s)
+
\exp\!\left(\frac{1}{2}\ell_\phi(a,s)\right)\odot \epsilon,
\qquad
\epsilon\sim\mathcal N(0,I_{\mathsf d_z}),
\]
and decoded as
\[
\hat a=f_\theta(z_q,s).
\]

We use two prior parameterizations. The fixed-prior variant uses
\[
p_\psi(\cdot\mid s)=\mathcal N(0,I_{\mathsf d_z}).
\]
The learned-Gaussian variant uses an observation-conditioned diagonal
Gaussian prior,
\[
p_\psi(\cdot\mid s)
=
\mathcal N\!\left(
\mu_\psi(s),
\operatorname{diag}\exp \ell_\psi(s)
\right),
\]
where \(\mu_\psi(s)\) and \(\ell_\psi(s)\) are predicted from the
observation representation. In practice, we observe that learning the
variance and fixing the variance to \(I_{\mathsf d_z}\) yields similar
performance for the simulation benchmarks in Section~\ref{sec:sim-results}. Thus, the reported learned-prior KL-CVAE uses a learned-mean Gaussian prior,
\[
p_\psi(\cdot\mid s)=\mathcal N(\mu_\psi(s),I_{\mathsf d_z}),
\]
where \(\mu_\psi(s)\) is predicted from the observation representation.

The training objective is
\[
\mathcal L_{\rm KL}
=
\E_{(S,A)\sim p_{\mathcal D}}
\E_{Z_q\sim q_\phi(\cdot\mid A,S)}
\left[
\mathcal L_{\rm rec}(f_\theta(Z_q,S),A)
\right]
+
\beta\,
\E_{(S,A)\sim p_{\mathcal D}}
\KL\!\left(
q_\phi(\cdot\mid A,S)
\,\middle\|\,
p_\psi(\cdot\mid S)
\right).
\]
At inference, actions are generated by sampling from the prior,
\[
Z_p\sim p_\psi(\cdot\mid s),
\qquad
\hat a=f_\theta(Z_p,s).
\]

\paragraph{Aggregate-matched latent policies}
\label{app:method-cwae}

MMD-CWAE and Sinkhorn-CWAE use the same latent-conditioned decoder as KL-CVAE, but replace the pointwise KL term by an aggregate posterior matching objective, following the WAE formulation~\citep{tolstikhin2018wae}. For these variants, the posterior encoder is deterministic. Given a minibatch \(\{(s_i,a_i)\}_{i=1}^{\mathsf B}\), the posterior latents are
\[
z_i^q
=
e_\phi(a_i,s_i),
\qquad i=1,\ldots,\mathsf B .
\]
Prior latents are sampled as
\[
z_i^p\sim p_\psi(\cdot\mid s_i),
\qquad i=1,\ldots,\mathsf B .
\]
The corresponding empirical posterior and prior latent measures are
\[
\widehat Q_Z
=
\frac{1}{\mathsf B}\sum_{i=1}^{\mathsf B}\delta_{z_i^q},
\qquad
\widehat P_Z
=
\frac{1}{\mathsf B}\sum_{i=1}^{\mathsf B}\delta_{z_i^p}.
\]

For MMD-CWAE, the aggregate penalty is the squared maximum mean
discrepancy
\[
\MMD_k^2(\widehat Q_Z,\widehat P_Z)
=
\frac{1}{\mathsf B^2}
\sum_{i,j=1}^{\mathsf B}
\Big[
k(z_i^q,z_j^q)
+
k(z_i^p,z_j^p)
-
2k(z_i^q,z_j^p)
\Big],
\]
where \(k\) is the latent kernel. For Sinkhorn-CWAE, the aggregate
penalty is the debiased Sinkhorn divergence~\citep{feydy2019sinkhorn}
\[
\Sink_{\varepsilon,c}(\widehat Q_Z,\widehat P_Z)
=
\mathrm{OT}_{\varepsilon,c}(\widehat Q_Z,\widehat P_Z)
-
\frac{1}{2}\mathrm{OT}_{\varepsilon,c}(\widehat Q_Z,\widehat Q_Z)
-
\frac{1}{2}\mathrm{OT}_{\varepsilon,c}(\widehat P_Z,\widehat P_Z),
\]
where \(\mathrm{OT}_{\varepsilon,c}\) denotes entropic optimal transport
with cost \(c\) and entropic regularization strength \(\varepsilon\).

The aggregate-matched objective is
\[
\mathcal L_{\rm CWAE}
=
\frac{1}{\mathsf B}
\sum_{i=1}^{\mathsf B}
\mathcal L_{\rm rec}
\!\left(f_\theta(z_i^q,s_i),a_i\right)
+
\beta\,
D(\widehat Q_Z,\widehat P_Z),
\]
where
\[
D(\widehat Q_Z,\widehat P_Z)
=
\MMD_k^2(\widehat Q_Z,\widehat P_Z)
\]
for MMD-CWAE, and
\[
D(\widehat Q_Z,\widehat P_Z)
=
\Sink_{\varepsilon,c}(\widehat Q_Z,\widehat P_Z)
\]
for Sinkhorn-CWAE.

The fixed-prior variants use
\[
p_\psi(\cdot\mid s)=\mathcal N(0,I_{\mathsf d_z}),
\qquad
z_i^p\sim \mathcal N(0,I_{\mathsf d_z}).
\]
The learned-prior variants use an observation-conditioned prior
\(p_\psi(\cdot\mid s)\), so that
\[
z_i^p\sim p_\psi(\cdot\mid s_i).
\]
At inference, actions are generated by sampling from the prior and
decoding,
\[
z_p\sim p_\psi(\cdot\mid s),
\qquad
\hat a=f_\theta(z_p,s).
\]
\paragraph{Conditional aggregate matching}
\label{app:method-conditional-cwae}

In the synthetic benchmark, each task is evaluated under a fixed layout,
so matching the marginal latent distribution is often sufficient. In the
simulation benchmarks, the observation \(s\) varies substantially across
episodes and determines which action chunks are valid. Matching only the
marginal latent distribution can therefore mix latent samples coming from
different conditioning states. We address this by matching joint
condition--latent samples.

Let
\[
h_i=h_\eta(s_i)
\]
be the action-free condition embedding emitted by the prior network. In
the conditional matching loss, this vector is \(L_2\)-normalized and
scaled by a state-weight parameter \(\lambda_s\ge 0\):
\[
\tilde h_i
=
\frac{h_i}{\|h_i\|_2},
\qquad
\bar h_i
=
\sqrt{\lambda_s}\,\tilde h_i .
\]
The empirical joint posterior and prior measures are
\[
\widehat Q_{\bar h,Z}
=
\frac{1}{\mathsf B}\sum_{i=1}^{\mathsf B}\delta_{(\bar h_i,z_i^q)},
\qquad
\widehat P_{\bar h,Z}
=
\frac{1}{\mathsf B}\sum_{i=1}^{\mathsf B}\delta_{(\bar h_i,z_i^p)}.
\]

For conditional MMD-CWAE, we use a product kernel on condition--latent
pairs,
\[
k\bigl((\bar h,z),(\bar h',z')\bigr)
=
k_s(\bar h,\bar h')\,k_z(z,z').
\]
The conditional MMD penalty is
\[
\MMD_{\rm cond}^2(\widehat Q_{\bar h,Z},\widehat P_{\bar h,Z})
=
\frac{1}{\mathsf B^2}
\sum_{i,j=1}^{\mathsf B}
k_s(\bar h_i,\bar h_j)
\Big[
k_z(z_i^q,z_j^q)
+
k_z(z_i^p,z_j^p)
-
2k_z(z_i^q,z_j^p)
\Big].
\]
The objective is
\[
\mathcal L_{\rm cMMD}
=
\frac{1}{\mathsf B}
\sum_{i=1}^{\mathsf B}
\mathcal L_{\rm rec}
\!\left(f_\theta(z_i^q,s_i),a_i\right)
+
\beta\,
\MMD_{\rm cond}^2(\widehat Q_{\bar h,Z},\widehat P_{\bar h,Z}) .
\]

For relaxed conditional Sinkhorn-CWAE, we perform transport in the joint
condition--latent space. Define
\[
x_i^q
=
\begin{bmatrix}
\bar h_i\\
z_i^q
\end{bmatrix},
\qquad
x_i^p
=
\begin{bmatrix}
\bar h_i\\
z_i^p
\end{bmatrix}.
\]
Since \(\bar h_i=\sqrt{\lambda_s}\tilde h_i\), the squared Euclidean
cost between posterior sample \(i\) and prior sample \(j\) is
\[
c_{ij}
=
\lambda_s
\|\tilde h_i-\tilde h_j\|_2^2
+
\|z_i^q-z_j^p\|_2^2 .
\]
The conditional Sinkhorn penalty is the debiased Sinkhorn divergence
\[
\Sink_{\varepsilon,c}(\widehat Q_{\bar h,Z},\widehat P_{\bar h,Z})
=
\mathrm{OT}_{\varepsilon,c}(\widehat Q_{\bar h,Z},\widehat P_{\bar h,Z})
-
\frac{1}{2}
\mathrm{OT}_{\varepsilon,c}(\widehat Q_{\bar h,Z},\widehat Q_{\bar h,Z})
-
\frac{1}{2}
\mathrm{OT}_{\varepsilon,c}(\widehat P_{\bar h,Z},\widehat P_{\bar h,Z}),
\]
where \(\mathrm{OT}_{\varepsilon,c}\) denotes entropic optimal transport
with cost \(c\) and entropic regularization strength \(\varepsilon\).
The objective is
\[
\mathcal L_{\rm cSink}
=
\frac{1}{\mathsf B}
\sum_{i=1}^{\mathsf B}
\mathcal L_{\rm rec}
\!\left(f_\theta(z_i^q,s_i),a_i\right)
+
\beta\,
\Sink_{\varepsilon,c}(\widehat Q_{\bar h,Z},\widehat P_{\bar h,Z}) .
\]

\paragraph{Posterior geometry regularization}
\label{app:method-posterior-geometry}

Aggregate matching constrains the batch-level distribution of posterior
latents, but it does not by itself control the local geometry of the
posterior codes used by the decoder. In practice, the deterministic
posterior can become poorly conditioned, for example by concentrating in
thin directions, drifting away from the prior scale, or developing strong
correlations across latent dimensions. We therefore add a simple
minibatch moment regularizer on the posterior latents, together with
optional decoder-side jitter. The regularizer keeps the posterior codes
approximately centered, controls their per-dimension scale, caps excessive
spread, and penalizes off-diagonal covariance.

For aggregate-matched variants, the posterior encoder is deterministic,
so each demonstration pair produces a latent
\[
z_i^q=e_\phi(a_i,s_i).
\]
Let
\[
\bar z^q
=
\frac{1}{\mathsf B}\sum_{i=1}^{\mathsf B}z_i^q,
\qquad
\tilde z_i^q
=
z_i^q-\bar z^q .
\]
For each latent dimension \(d\), define the minibatch standard deviation
\[
\hat\sigma_d
=
\left(
\frac{1}{\mathsf B}
\sum_{i=1}^{\mathsf B}
(\tilde z_{i,d}^q)^2
+\varepsilon_{\rm num}
\right)^{1/2},
\]
and define the empirical covariance
\[
\widehat C
=
\frac{1}{\mathsf B-1}
\sum_{i=1}^{\mathsf B}
\tilde z_i^q(\tilde z_i^q)^\top .
\]
The posterior geometry regularizer is
\[
\mathcal L_{\rm geom}
=
\beta_{\rm mean}\|\bar z^q\|_2^2
+
\beta_{\rm std}
\frac{1}{\mathsf d_z}
\sum_{d=1}^{\mathsf d_z}
(\hat\sigma_d-\sigma_\star)^2
+
\beta_{\rm max}
\frac{1}{\mathsf d_z}
\sum_{d=1}^{\mathsf d_z}
[\hat\sigma_d-\sigma_{\max}]_+^2
+
\beta_{\rm cov}
\frac{1}{\mathsf d_z}
\left\|
\operatorname{offdiag}(\widehat C)
\right\|_F^2 .
\]
Here \(\sigma_\star\) is the target per-dimension standard deviation,
\(\sigma_{\max}\) is the maximum allowed standard deviation, and
\(\varepsilon_{\rm num}>0\) is a numerical stabilizer. When this
regularizer is enabled, it is added to the corresponding
aggregate-matched objective:
\[
\mathcal L
=
\mathcal L_{\rm CWAE}
+
\mathcal L_{\rm geom}.
\]

We also optionally apply decoder-side latent jitter during training. The
latent used for aggregate matching remains \(z_i^q\), while the latent
consumed by the decoder is
\[
z_i^{\rm dec}
=
z_i^q+\sigma_{\rm dec}\epsilon_i,
\qquad
\epsilon_i\sim\mathcal N(0,I_{\mathsf d_z}).
\]
Thus, the aggregate posterior measure is still constructed from
\(\{z_i^q\}_{i=1}^{\mathsf B}\), while reconstruction is computed with
\(f_\theta(z_i^{\rm dec},s_i)\).
\paragraph{Latent flow prior (LAT-Flow)}
\label{app:method-lat-flow}

LAT-Flow uses a deterministic posterior encoder and a latent-conditioned
decoder,
\[
z_i^q=e_\phi(a_i,s_i),
\qquad
\hat a_i=f_\theta(z_i^{\rm dec},s_i).
\]
The prior \(p_\psi(\cdot\mid s)\) is parameterized by a conditional
flow-matching model in latent space. During training, sample
\[
u_i\sim\mathcal N(0,I_{\mathsf d_z}),
\qquad
t_i\sim p(t),
\]
and construct the interpolation
\[
z_{t_i}
=
(1-t_i)u_i+t_i z_i^q .
\]
For the linear path, the target velocity is
\[
v_i^\star
=
z_i^q-u_i .
\]
The flow prior predicts
\[
v_\psi(z_{t_i},t_i,s_i),
\]
and is trained with
\[
\mathcal L_{\rm flow}
=
\frac{1}{\mathsf B}
\sum_{i=1}^{\mathsf B}
\left\|
v_\psi(z_{t_i},t_i,s_i)-v_i^\star
\right\|_2^2 .
\]
The full training objective is
\[
\mathcal L_{\rm LAT}
=
\frac{1}{\mathsf B}
\sum_{i=1}^{\mathsf B}
\mathcal L_{\rm rec}
\!\left(f_\theta(z_i^{\rm dec},s_i),a_i\right)
+
\beta\mathcal L_{\rm flow}.
\]
When enabled, the posterior geometry regularizer
\(\mathcal L_{\rm geom}\) is added to this objective.

At inference, sample
\[
z_0\sim\mathcal N(0,I_{\mathsf d_z})
\]
and integrate
\[
\frac{d z_t}{dt}
=
v_\psi(z_t,t,s)
\]
from \(t=0\) to \(t=1\). The resulting latent prior sample \(z_1\) is
decoded as
\[
\hat a=f_\theta(z_1,s).
\]

\paragraph{Residual VQ-VAE}
\label{app:method-vq-vae}

VQ-VAE uses the same latent-conditioned action decoder as the continuous
latent policies, but replaces the continuous posterior sample by a
residual vector-quantized latent. Given a minibatch
\(\{(s_i,a_i)\}_{i=1}^{\mathsf B}\), the posterior encoder first
produces continuous latents
\[
z_{e,i}=e_\phi(a_i,s_i)\in\mathbb R^{\mathsf d_z},
\qquad i=1,\ldots,\mathsf B .
\]
A residual vector quantizer with \(\mathsf L_{\rm vq}\) layers and
\(\mathsf C_{\rm vq}\) codes per layer maps each \(z_{e,i}\) to a
quantized latent \(z_{q,i}\). Let
\(\mathcal E_\ell=\{e_{\ell,1},\ldots,e_{\ell,\mathsf C_{\rm vq}}\}\)
denote the codebook of layer \(\ell\). Starting with residual
\(r_{i,1}=z_{e,i}\), each layer selects the nearest code
\[
c_{i,\ell}
=
\arg\min_{j\in\{1,\ldots,\mathsf C_{\rm vq}\}}
\left\|
r_{i,\ell}-e_{\ell,j}
\right\|_2^2,
\qquad
q_{i,\ell}=e_{\ell,c_{i,\ell}},
\]
and updates the residual as
\[
r_{i,\ell+1}=r_{i,\ell}-\operatorname{sg}(q_{i,\ell}),
\]
where \(\operatorname{sg}(\cdot)\) denotes stop-gradient. The final
quantized latent is the sum of residual code contributions,
\[
z_{q,i}
=
\sum_{\ell=1}^{\mathsf L_{\rm vq}} q_{i,\ell},
\]
with straight-through gradients used through the quantization operation.

The action decoder receives \(z_{q,i}\) and predicts
\[
\hat a_i=f_\theta(z_{q,i},s_i).
\]
The minibatch training objective is
\[
\mathcal L_{\rm VQ}
=
\frac{1}{\mathsf B}
\sum_{i=1}^{\mathsf B}
\mathcal L_{\rm rec}\!\left(f_\theta(z_{q,i},s_i),a_i\right)
+
\beta_{\rm vq}
\frac{1}{\mathsf B\,\mathsf L_{\rm vq}}
\sum_{i=1}^{\mathsf B}
\sum_{\ell=1}^{\mathsf L_{\rm vq}}
\left\|
r_{i,\ell}
-
\operatorname{sg}(q_{i,\ell})
\right\|_2^2 .
\]
The codebooks are updated by exponential moving averages of assigned
encoder outputs, with first-batch initialization and dead-code
replacement, following vector-quantized autoencoders
~\citep{van2017neural,lee2024vqbet}.

At inference, we use a fixed uniform prior over codebook indices. For
each residual layer,
\[
c_\ell^p
\sim
\operatorname{Unif}\{1,\ldots,\mathsf C_{\rm vq}\},
\qquad
\ell=1,\ldots,\mathsf L_{\rm vq},
\]
and reconstruct the prior latent by decoding the sampled code indices,
\[
z_p
=
\sum_{\ell=1}^{\mathsf L_{\rm vq}} e_{\ell,c_\ell^p}.
\]
The predicted action chunk is then
\[
\hat a=f_\theta(z_p,s).
\]

\paragraph{Action-space flow and diffusion policies}
\label{app:method-action-space}

Act-Flow and Act-Diff model the action chunk directly in
\(\mathcal A^{\mathsf H}\), without an explicit posterior latent
variable. For a fixed observation \(s\), both methods define a
deterministic sampler
\[
G_{\theta,s}(u):=G_\theta(s,u),
\]
which pushes forward a simple Gaussian base distribution
\(p_0=\mathcal N(0,I)\) on the flattened action-chunk space to the conditional action chunk distribution:
\[
\pi_\theta(\cdot\mid s)
=
(G_{\theta,s})_{\#}p_0 .
\]
Equivalently, sampling from the policy is written as
\[
U\sim p_0,
\qquad
A=G_{\theta,s}(U).
\]
For Act-Flow, \(G_{\theta,s}\) is the ODE solution map induced by the
learned velocity field. For Act-Diff, \(G_{\theta,s}\) is the full
composition of deterministic DDIM denoising steps. Both variants use the
same conditional action transformer backbone: noisy or interpolated
action tokens attend jointly with observation tokens, and the network is
conditioned on a timestep embedding.

For Act-Flow, we train a conditional flow-matching model in action space. Given a minibatch \(\{(s_i,a_i)\}_{i=1}^{\mathsf B}\), sample
\[
u_i\sim\mathcal N(0,I),
\qquad
t_i\sim p(t),
\]
and form the linear interpolation
\[
a_{t_i}
=
(1-t_i)u_i+t_i a_i .
\]
For this path, the target velocity is
\[
v_i^\star
=
a_i-u_i .
\]
The model predicts a velocity field
\[
v_\theta(a_{t_i},t_i,s_i),
\]
and is trained with
\[
\mathcal L_{\rm ActFlow}
=
\frac{1}{\mathsf B}
\sum_{i=1}^{\mathsf B}
\left\|
v_\theta(a_{t_i},t_i,s_i)-v_i^\star
\right\|_2^2 .
\]
At inference, we sample \(a_0\sim\mathcal N(0,I)\) and integrate
\[
\frac{d a_t}{dt}
=
v_\theta(a_t,t,s)
\]
from \(t=0\) to \(t=1\). The resulting \(a_1\) is used as the predicted
action chunk.

For Act-Diff, we train a denoising diffusion model in action space. Let
\(\bar\alpha_t\) denote the cumulative noise schedule. Given a
minibatch \(\{(s_i,a_i)\}_{i=1}^{\mathsf B}\), sample a timestep \(t_i\)
and noise \(\epsilon_i\sim\mathcal N(0,I)\), and construct
\[
a_{t_i}
=
\sqrt{\bar\alpha_{t_i}}\,a_i
+
\sqrt{1-\bar\alpha_{t_i}}\,\epsilon_i .
\]
The model predicts the clean action chunk directly,
\[
\hat a_i
=
g_\theta(a_{t_i},t_i,s_i),
\]
and is trained with the sample-prediction objective
\[
\mathcal L_{\rm ActDiff}
=
\frac{1}{\mathsf B}
\sum_{i=1}^{\mathsf B}
\left\|
g_\theta(a_{t_i},t_i,s_i)-a_i
\right\|_2^2 .
\]
At inference, we initialize \(a_{\mathsf T_{\rm diff}}\sim\mathcal N(0,I)\)
and apply the deterministic DDIM sampler to obtain the final action
chunk. The full composition of denoising steps defines the sampling map
\(G_{\theta,s}:u\mapsto a\) used in Section \ref{par:action-space-theory}.

\subsection{Network Architectures}
\label{app:network-arch}
This section describes the neural implementations, the benchmarks, and the training and evaluation details of the experiments.
All policies predict action chunks non-autoregressively. Within each benchmark, methods share the same observation encoder, action representation, normalization pipeline, and training loop, and differ only in the output parameterization.

\paragraph{Observation encoders.}
Image observations are encoded with a convolutional encoder, initialized
from pretrained weights and fine-tuned during policy training. The
encoder preserves the spatial feature map, uses no global pooling, and
uses frozen batch-normalization statistics. Proprioceptive and state
observations are encoded with a single projection layer.

\paragraph{Action decoder.}
The deterministic and latent-variable policies use non-autoregressive
action transformer decoders. Encoded observations are first converted into
a token sequence. Spatial feature maps receive two-dimensional sinusoidal
positional encodings, while flat features receive one-dimensional
positional encodings. Let
\[
Q\in\mathbb R^{\mathsf H\times \mathsf d}
\]
denote the learnable action-query sequence, with one query embedding per
predicted action timestep. Given the observation-token sequence, the
decoder maps \(Q\) to \(\mathsf H\) action embeddings in parallel, and
task-specific action heads map these embeddings to the predicted action
components.

The deterministic baseline uses the non-latent version of this decoder.
For an observation \(s\), the action transformer produces action
embeddings
\[
U_{1:\mathsf H}=D_\theta(Q,X_s),
\]
where \(X_s\in\mathbb R^{\mathsf N_s\times \mathsf d}\) denotes the
encoded observation tokens. The action heads then produce a single
predicted chunk,
\[
\hat a=\operatorname{Head}_\theta(U_{1:\mathsf H})
      =f_\theta(s).
\]

\paragraph{LACT.}
Latent-variable policies use LACT, a latent-conditioned variant of the
previous action decoder. Given a latent \(z\), LACT conditions each
decoder block through AdaLN-Zero-style modulation, as used in DiT-style
transformers. The conditioning network maps \(z\) to scale, shift, and
residual-gating parameters, which modulate the normalization and residual
branches of the self-attention, cross-attention, and feedforward blocks.
Thus, the latent affects the full action-query decoding process, and the
decoder implements the map
\[
\hat a=f_\theta(z,s).
\]

\paragraph{MMDiT action decoder.}
Action-space flow and diffusion policies use an MMDiT-style dual-stream
transformer~\citep{esser2024scaling}. The model receives observation
tokens and action tokens as separate streams. The action stream contains
either interpolated actions for flow matching or noisy actions for
diffusion. A timestep embedding conditions the transformer layers through
adaptive normalization. The dual-stream decoder allows bidirectional
attention between observation and action tokens, and a final prediction
layer maps the action-token outputs to either velocities for Act-Flow or
denoising targets for Act-Diff.

For Act-Flow, the MMDiT predicts the action-space velocity field used by
the flow solver. For Act-Diff, the MMDiT predicts the diffusion training
target and is sampled with a deterministic DDIM scheduler.

\paragraph{Latent encoders.}
Latent-variable policies use transformer encoders to parameterize the
posterior and, when used, the learned prior. The posterior encoder
receives a sequence of tokens containing the ground-truth action sequence
together with the encoded observation tokens. A learnable class token is
appended to the sequence, and a transformer encoder maps the final
class-token representation to the latent output.

For KL-CVAE, the posterior projection has output dimension
\(2\mathsf d_z\) and is split into a mean and log-variance,
\[
(\mu_\phi(a,s),\ell_\phi(a,s))
=
P_\phi\!\left(\operatorname{CLS}_\phi(a,s)\right),
\]
which define the diagonal Gaussian posterior
\(q_\phi(\cdot\mid a,s)\). For aggregate-matched variants and LAT-Flow,
the projection has output dimension \(\mathsf d_z\) and directly returns
\[
z^q=e_\phi(a,s).
\]

The VQ-VAE posterior uses the same action-plus-observation transformer
encoder, but replaces the continuous latent output by a residual
vector-quantized bottleneck. The class-token representation is first
projected to a continuous vector
\[
z^{\rm cont}=P_\phi(\operatorname{CLS}_\phi(a,s)),
\]
which is then quantized by a residual codebook stack. The quantized
latent \(z^q\) is passed to the action decoder.

Learned priors use a separate transformer encoder that receives only
observation tokens. A class token is appended to the observation sequence,
and the final class-token representation is projected to prior
parameters,
\[
(\mu_\psi(s),\ell_\psi(s))
=
P_\psi\!\left(\operatorname{CLS}_\psi(s)\right),
\]
defining
\[
p_\psi(\cdot\mid s)
=
\mathcal N\!\left(
\mu_\psi(s),
\operatorname{diag}\exp\ell_\psi(s)
\right).
\]
For conditional aggregate matching, the same prior encoder also emits an
action-free condition vector \(h_\eta(s)\). This vector is obtained by
pooling the observation tokens and used to compute the conditional
MMD/Sinkhorn losses.

\subsection{Benchmarks}
\label{app:benchmarks}

All benchmarks use observation horizon \(\mathsf T_{\rm obs}=1\). Thus,
each policy conditions on a single current observation, without
observation history. The action horizon depends on the benchmark.
Synthetic tasks use \(\mathsf H=60\), corresponding to the full task
trajectory. Simulation and real-world robot benchmarks use \(\mathsf H=10\); at evaluation
time, we execute the first two actions from each predicted chunk before
replanning.

\paragraph{Synthetic benchmark.}
The synthetic benchmark consists of four 2D navigation tasks. All
trajectories lie in the unit square \([0,1]^2\). Each task contains
\(1000\) demonstrations. Policies receive a single \(64\times64\) RGB
rendering at \(t=0\), showing the agent initial position, obstacles, and
goal regions, and predict the complete trajectory chunk
\(a_{0:\mathsf H-1}\in\mathbb R^{60\times2}\), with \(\mathsf H=60\),
without replanning. Thus, each trajectory effectively contains one single training sample pair. The four tasks differ in their mode structure. \emph{Circle} has
\(\mathsf K=2\) circular trajectory modes around obstacles.
\emph{Sequential} has two consecutive binary decisions, yielding
\(\mathsf K=4\) modes separated by three obstacles. \emph{Radial} has
\(\mathsf K=16\) modes from a common start to sixteen evenly spaced
radial goals, with obstacles between neighboring directions.
\emph{Corridor} has \(\mathsf K=16\) valid routes from a common start to
a common endpoint through sixteen corridors induced by fifteen obstacles. For synthetic evaluation, a generated trajectory is successful if it
satisfies all three criteria: it does not collide with obstacles; its
endpoint is within distance \(0.1\) of one valid mode endpoint; and its
path length is at least \(0.8\) times the mean expert path length. The
path-length condition rules out trivial collapsed trajectories that reach
an endpoint without following a valid path. We also report valid mode
coverage and valid mode entropy ratio, computed only over successful
trajectories.

\paragraph{Simulation benchmarks.}
We use PushT, Franka Kitchen, and UR3 BlockPush as robotic simulation
benchmarks previously employed in the literature of multimodal policy
\citep{florence2021implicit,chi2025diffusion,shafiullah2022behavior,
lee2024vqbet,sheebaelhamd2025quantization}, together with the language-conditioned benchmarks LIBERO and Meta-World, extensively used in the vision-language-action model benchmarking, see for example \cite{black2024pi0, shukor2025smolvla}.
PushT requires pushing a T-shaped block to a target pose using 2D
end-effector control~\citep{chi2025diffusion}. We use both image and
state variants. The image observation is a single \(96\times96\) RGB
agent-view image. The state observation contains the end-effector
position, block position, block angle, block keypoints, and contact
indicator. PushT performance is measured by final intersection-over-union
(IoU) between the block pose and target pose. Franka Kitchen~\citep{gupta2019relay} is a long-horizon manipulation
environment in which a Franka arm completes a subset of kitchen subtasks
in sequence. We use state and image variants. The state observation
contains \(9\) robot-arm configuration variables and \(21\) object
configuration variables. The image observation is a single \(112\times112\)
RGB rendering. Performance is reported as the number of completed goals
out of four. UR3 BlockPush requires moving two blocks to two target regions. The task
is multimodal because the blocks can be pushed in different orders. We use
the unconditional state variant. The state contains the 2D end-effector
position and the 2D positions of both blocks; the action is a 2D
end-effector target. Performance is reported as the number of achieved
goals out of two.
LIBERO~\citep{liu2023libero} is a language-specified robotic manipulation
benchmark. We train one policy on four suites with ten tasks each: LIBERO-Spatial, LIBERO-Object, LIBERO-Goal, and LIBERO-Long. The first three suites vary
spatial relationships, manipulated objects, and task goals, respectively,
while LIBERO-Long emphasizes longer-horizon manipulation. 
Meta-World~\citep{yu2020meta} comprises 50 robotic manipulation tasks
performed by a simulated Sawyer arm. Its four-dimensional action space
consists of three-dimensional end-effector displacements and a gripper
command. We train on the MT50 multi-task suite and report results for the Easy, Medium, Hard, and
Very Hard difficulty groups, together with aggregate performance across
all evaluated tasks.

\begin{figure}[t]
    \centering
    \begin{minipage}[c]{0.32\linewidth}
        \centering
        \includegraphics[width=\linewidth]{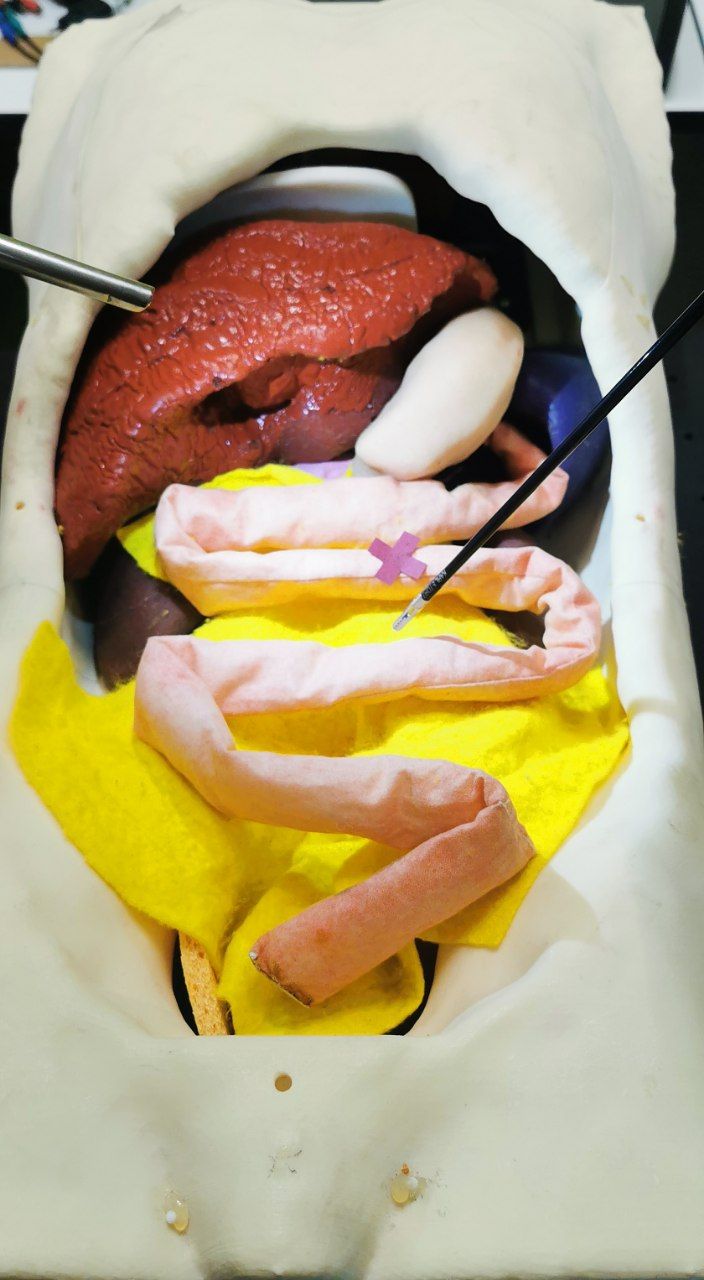}
        
    \end{minipage}
    \hspace{0.02\linewidth}
    \begin{minipage}[c]{0.62\linewidth}
        \centering
        \includegraphics[width=0.8\linewidth]{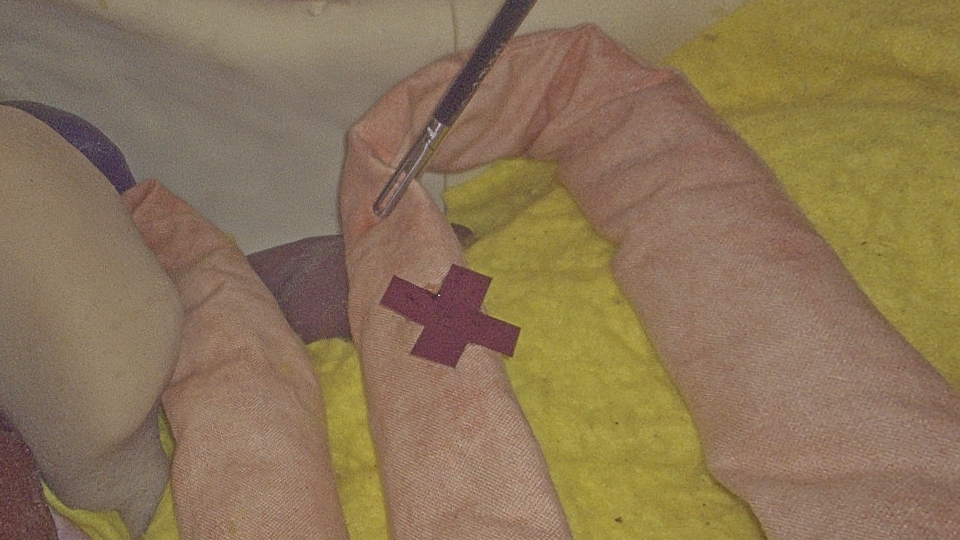}

        \vspace{0.5em}

        \includegraphics[width=0.8\linewidth]{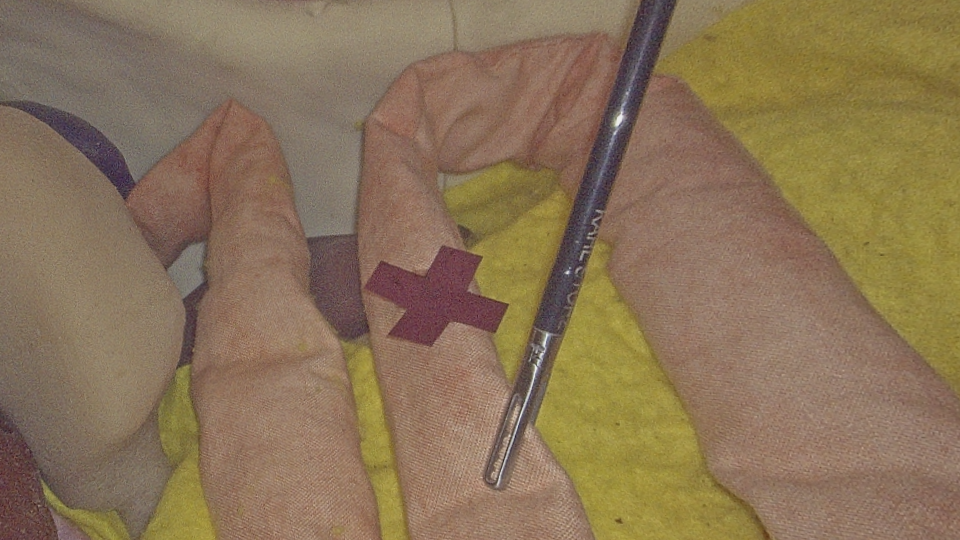}
    \end{minipage}

    \caption{
        Real-world surgical robot task: bimodal tissue grasping.
        The robotic instrument needs to grasp the bowel tissue while avoiding
        a sensitive region of interest marked with a red cross.
    }
    \label{fig:bowelgrasping}
\end{figure}

\paragraph{Real-world benchmark: bimodal tissue grasping}
 To validate the policies on a real-world robotic platform, we design the surgical task of bimodal tissue grasping. During minimally invasive surgery for gastrointestinal cancer, surgical assistants commonly manipulate and retract the small bowel to expose the surgical field for procedures such as anastomosis and tumor resection \citep{konstantinidis2020trends}. In the bimodal bowel tissue grasping task, a segment of bowel tissue must be grasped and displaced while avoiding a sensitive region of interest, representing, for example, a tumor or another area where grasping could cause tissue trauma. Two separate safe grasping regions are available and located on opposite sides of the region of interest.  Both grasps are equally valid, but they correspond to distinct trajectory strategies, for example, reflecting left-side versus right-side approach preferences. The resulting conditional action-chunk distribution is therefore bimodal. We collect 90 demonstrations, balanced across the two modes with 45 demonstrations per mode, using an open-source surgical robotics platform \citep{rodriguez2026open} and the Open-HELP abdominal phantom \citep{openhelp}. A picture of the setup can be seen in Figure \ref{fig:bowelgrasping}. The observation space comprises the current Cartesian 3D position of the surgical instrument tip and the binary gripper state; the action space is the delta 3D movement of the tip and the binary gripper action.

\begin{table}[t]
\centering
\small
\setlength{\tabcolsep}{4pt}
\caption{Benchmark summary. 'L' stands for language instruction.}
\label{tab:benchmark-summary}
\begin{tabular*}{\linewidth}{@{\extracolsep{\fill}}lcccl@{}}
\toprule
Benchmark
& Demonstrations
& Observation
& Horizon \(\mathsf H\)
& Evaluation \\
\midrule
Synthetic
& \(1000\)/task
& RGB \(64\times64\)
& \(60\)
& success, coverage, entropy \\
PushT-State
& \(206\)
& state, 24-d
& \(10\)
& final IoU \\
PushT-Image
& \(206\)
& RGB \(96\times96\)
& \(10\)
& final IoU \\
Kitchen-State
& \(566\)
& state, 30-d
& \(10\)
& goals achieved out of 4 \\
Kitchen-Image
& \(566\)
& RGB \(112\times112\)
& \(10\)
& goals achieved out of 4 \\
UR3 BlockPush
& 600
& state, 6-d
& \(10\)
& goals achieved out of 2 \\
LIBERO
& 1693
& 2xRGB \(256\times256\), L
& \(10\)
& success \\
Meta-World
& 2500
& RGB \(480\times480\), L
& \(10\)
& success \\
Bimodal Tissue Grasping
& 90
& state, 4-d
& \(10\)
& success rate \\

\bottomrule
\end{tabular*}
\end{table}

\subsection{Training Details}
\label{app:training-details}

All runs used 32-bit precision. All policies in a given benchmark share
the same observation representation, normalization, action horizon, data
loader, optimizer, and training loop. The network architectures are described
in Appendix~\ref{app:network-arch}. The full hyperparameter tables are reported in Appendix \ref{app:hp}.
Each experiment was executed as a single-GPU job on an NVIDIA H100 94GB
GPU. The available memory was not fully utilized for smaller models; in
particular, synthetic runs required less than 4GB of VRAM. Synthetic
experiments used 8 CPU workers per job and required approximately
20--40 minutes of wall-clock time. Simulation and robot experiments used 14 CPU
workers per job and required approximately 10--20 hours of wall-clock
time, depending on the benchmark and model family.
\subsection{Evaluation Details}
\label{app:eval-details}

Policies are evaluated by rolling them out in the benchmark environments
on which they were trained. For the synthetic benchmarks, we perform
\(200\) rollouts per policy every \(100\) epochs and report each metric as
the average over the last five evaluation checkpoints, corresponding to
\(5\times200=1000\) rollouts. For the simulation benchmarks, we evaluate each policy on \(200\) rollout trials using a reproducible random seed that is varied per each roll-out, in order to sample reproducibly randomized
initial conditions. We repeat the evaluation over three seeds. We evaluate policies at fixed and latest checkpoint intervals. For the real-robot experiment, we evaluate each policy for 20 trials.
\newpage

%% file: sections/C_appendix.tex

\section{Additional Results}
\label{app:results}
\subsection{Complete Synthetic Benchmarks Results}

Table~\ref{tab:synthetic-full} reports the complete per-task synthetic
results. Valid mode coverage is the fraction of ground-truth modes reached
by at least one successful rollout, so invalid or colliding trajectories
do not increase coverage. The valid mode-entropy ratio (MER) is the
entropy of the empirical distribution \(\widehat p_{\rm valid}\) over
successful mode labels \(M\), normalized by the maximum entropy
\(\log \mathsf K\):
\[
\mathrm{MER}
=
\frac{H(\widehat p_{\rm valid})}{\log \mathsf K}.
\]
When no rollout is successful, both coverage and entropy ratio are set to
zero.

\begin{table}[h]
\centering
\small
\caption{Full synthetic results:
\textbf{success rate (SR) / valid mode coverage (Cov) / valid
mode-entropy ratio (MER)}. Values are 3-seed means, each run summarized
by the running mean over the last five rollout checkpoints, with 200
episodes per checkpoint. Bold indicates the best value per task and
metric.}
\label{tab:synthetic-full}
\begin{tabular}{@{}lcccc@{}}
\toprule
Method & Circle & Sequential & Corridor & Radial \\
\midrule
BCAT (deterministic)
& $0.00/0.00/0.00$
& $0.00/0.00/0.00$
& $0.00/0.00/0.00$
& $0.00/0.00/0.00$ \\

KL-CVAE
& $0.96/\mathbf{1.00}/0.99$
& $0.94/\mathbf{1.00}/\mathbf{0.99}$
& $0.47/0.94/0.92$
& $0.48/\mathbf{1.00}/0.95$ \\

MMD-CWAE
& $0.92/\mathbf{1.00}/\mathbf{1.00}$
& $0.94/\mathbf{1.00}/\mathbf{0.99}$
& $0.40/0.98/0.94$
& $0.34/\mathbf{1.00}/\mathbf{0.98}$ \\

Sinkhorn-CWAE
& $0.88/\mathbf{1.00}/0.98$
& $0.93/\mathbf{1.00}/\mathbf{0.99}$
& $0.43/\mathbf{0.99}/0.95$
& $0.47/\mathbf{1.00}/0.96$ \\

VQ-VAE
& $0.57/0.60/0.59$
& $0.67/0.67/0.67$
& $\mathbf{0.52}/0.94/0.93$
& $\mathbf{0.52}/0.99/0.95$ \\

Act-Flow
& $0.94/\mathbf{1.00}/0.99$
& $0.91/\mathbf{1.00}/0.97$
& $0.44/0.98/\mathbf{0.96}$
& $0.49/0.99/0.96$ \\

Act-Diff
& $\mathbf{1.00}/\mathbf{1.00}/0.99$
& $0.89/\mathbf{1.00}/0.94$
& $0.41/0.95/0.93$
& $0.44/0.97/0.93$ \\

LAT-Flow
& $\mathbf{1.00}/\mathbf{1.00}/\mathbf{1.00}$
& $\mathbf{0.96}/\mathbf{1.00}/\mathbf{0.99}$
& $0.48/\mathbf{0.99}/\mathbf{0.96}$
& $0.46/0.99/0.96$ \\
\bottomrule
\end{tabular}
\end{table}

\subsection{Collapse Diagnostics for Baseline Policies} \label{app:collapse-diagnostics} 
\begin{figure*}[t]
    \centering

    \makebox[\linewidth][c]{%
        \begin{subfigure}[t]{0.18\textwidth}
            \centering
            \includegraphics[width=\linewidth]{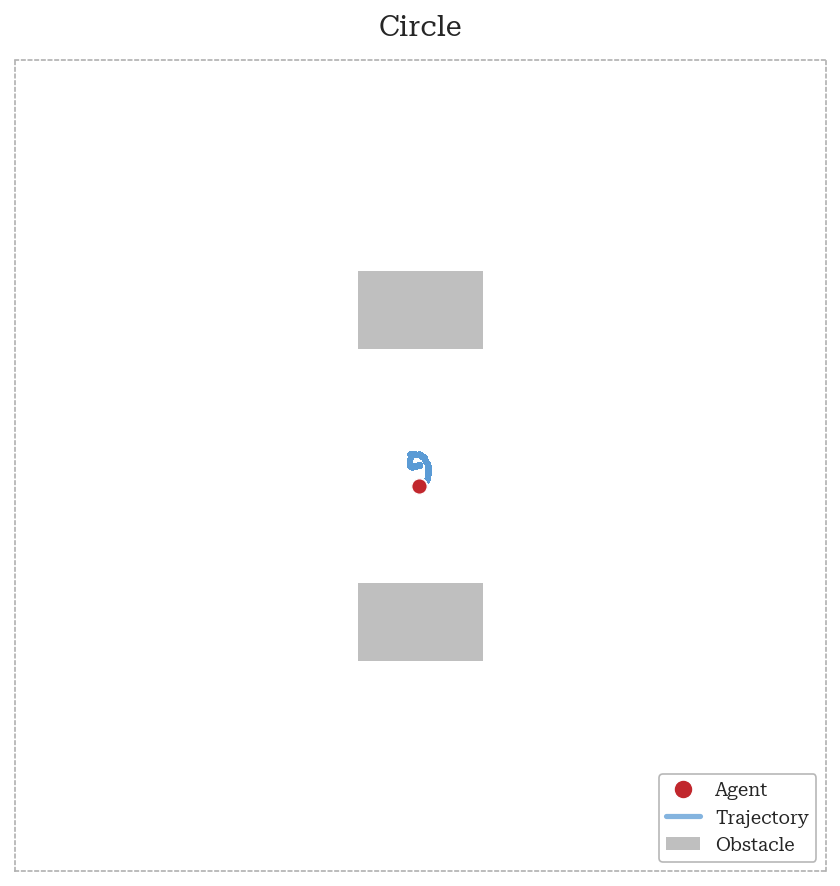}
            \caption{BCAT collapse on Circle.}
        \end{subfigure}
        \hspace{0.015\textwidth}
        \begin{subfigure}[t]{0.18\textwidth}
            \centering
            \includegraphics[width=\linewidth]{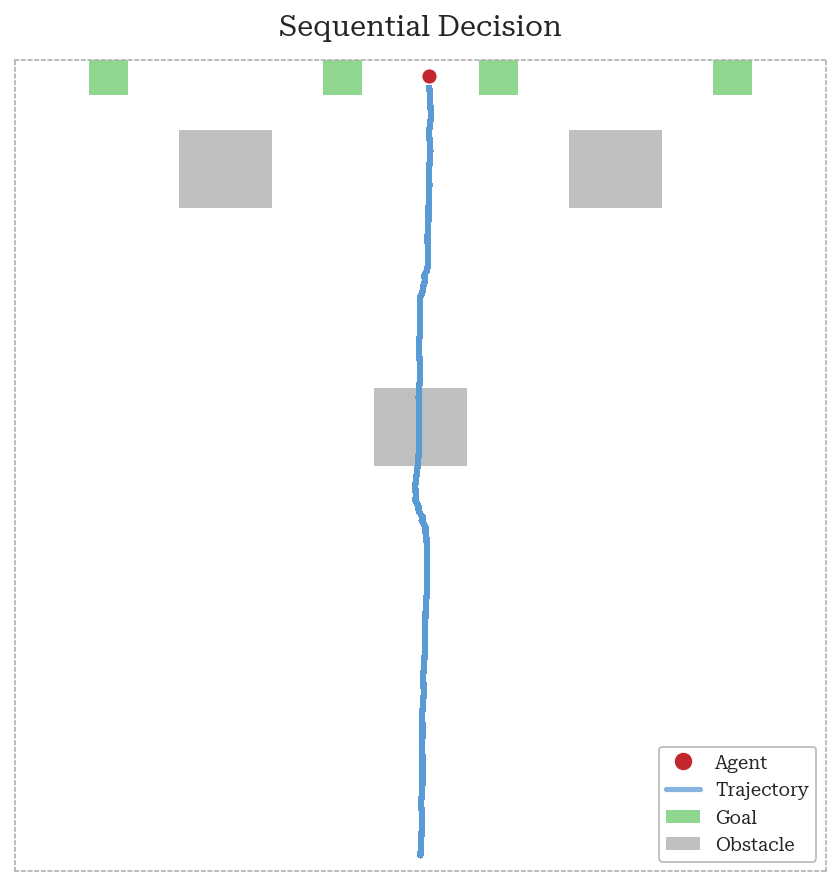}
            \caption{BCAT collapse on Sequential.}
        \end{subfigure}
        \hspace{0.015\textwidth}
        \begin{subfigure}[t]{0.18\textwidth}
            \centering
            \includegraphics[width=\linewidth]{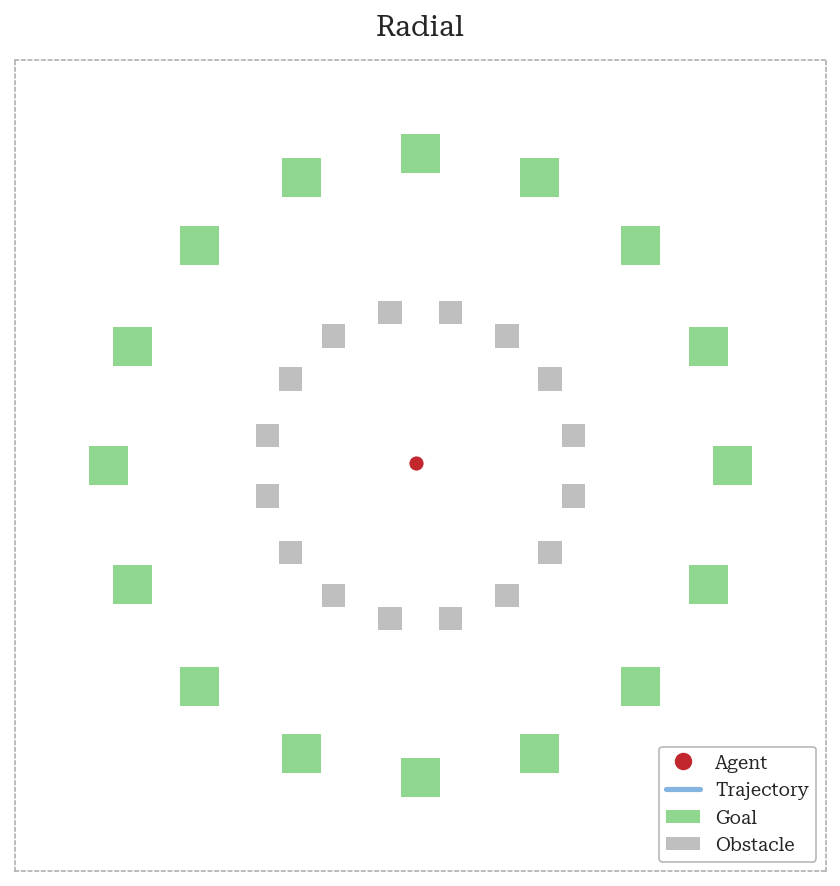}
            \caption{BCAT collapse on Radial.}
        \end{subfigure}
        \hspace{0.015\textwidth}
        \begin{subfigure}[t]{0.18\textwidth}
            \centering
            \includegraphics[width=\linewidth]{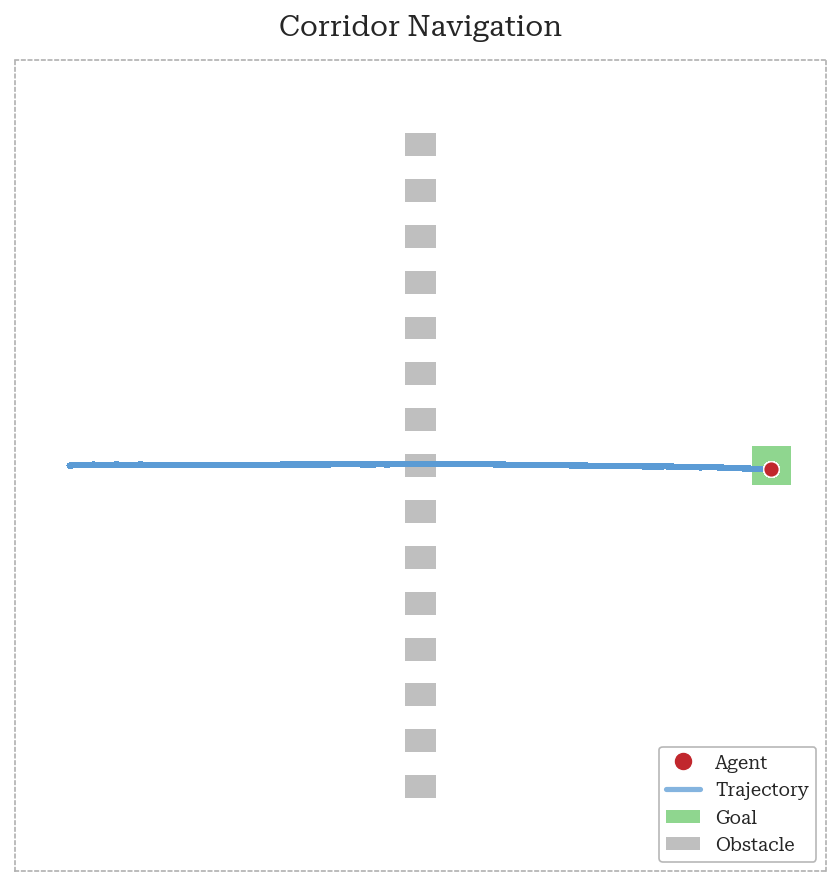}
            \caption{BCAT collapse on Corridor.}
        \end{subfigure}
        \hspace{0.015\textwidth}
        \begin{subfigure}[t]{0.18\textwidth}
            \centering
            \includegraphics[width=\linewidth]{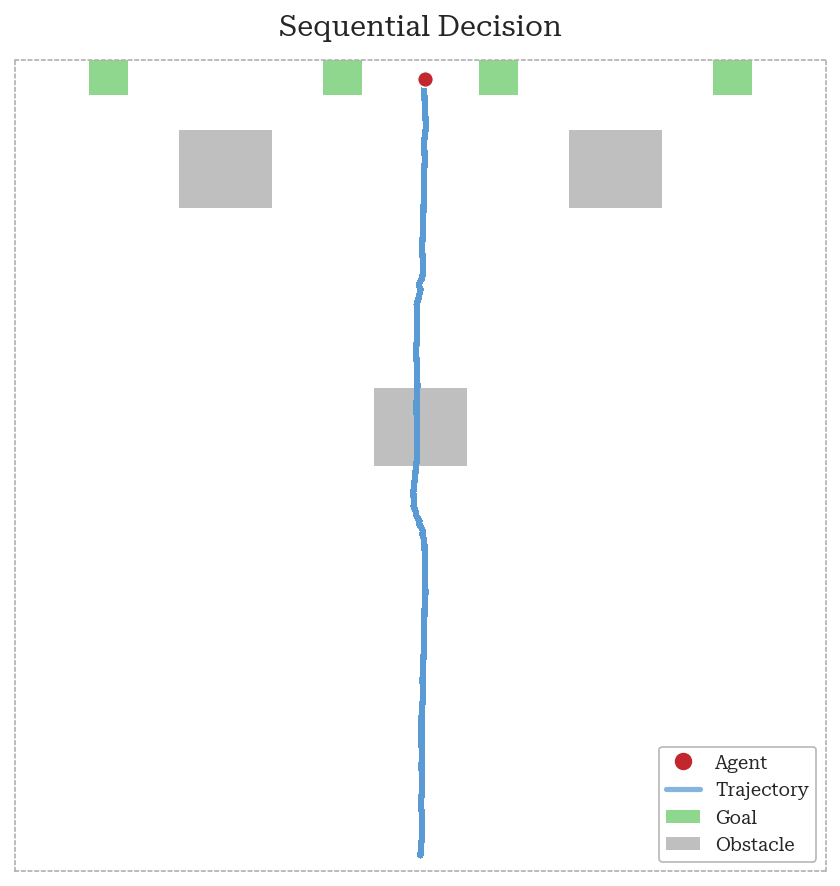}
            \caption{VQ-VAE collapse on Sequential.}
        \end{subfigure}%
    }

    \caption{
        Collapse diagnostics for baseline policies. The first four panels show
        representative predictions from deterministic BCAT on the four
        synthetic benchmarks. In all cases, the policy collapses to an
        averaged trajectory between valid modes and yields no successful
        rollouts. The fifth panel shows a representative VQ-VAE collapsed run
        on the Sequential task, illustrating the training instability observed
        for this baseline.
    }
    \label{fig:baseline-collapse}
\end{figure*}

Deterministic behavioral cloning exhibits complete collapse on all four synthetic benchmarks. As expected from the unimodal regression discussion in
Section~\ref{sec:intro}, the policy predicts an averaged trajectory between valid modes, and consequently fails to produce a single successful rollout on any task. Figure~\ref{fig:baseline-collapse} visualizes representative collapsed predictions for the four synthetic tasks. The VQ-VAE baseline was considerably less stable during training. While it achieved some of the strongest success rates on the more challenging \(\mathsf K=16\) tasks, it also collapsed completely on the easier Sequential task (\(\mathsf K=4\)), indicating high sensitivity to optimization dynamics and codebook usage. A representative failure run is shown in Figure~\ref{fig:baseline-collapse}. In preliminary PushT experiments, the same policy also performed poorly, reaching a score below \(0.05\) final IoU. We therefore did not pursue a broader simulation evaluation for this baseline. A plausible explanation is training instability in the single-stage setup. More structured two-stage training procedures may be beneficial here, as also adopted in prior vector-quantized imitation learning methods such as VQ-BeT~\citep{lee2024vqbet}, but we leave further investigation to future work.

\subsection{KL-regularized latent policies}
\label{app:kl-latent-mode-collapse}
\begin{figure}[t]
    \centering
    \includegraphics[width=\linewidth]{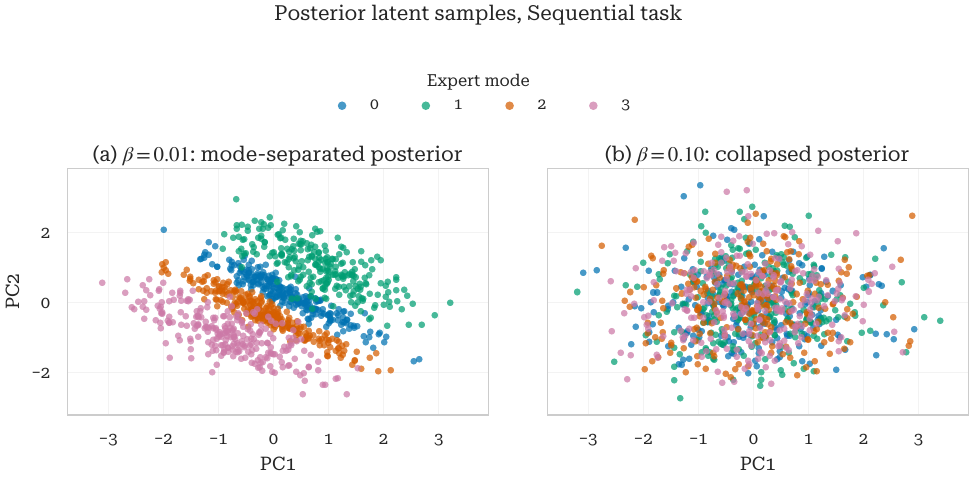}
    \caption{
    \textbf{Posterior latent geometry under pointwise KL pressure.}
    PCA projection of posterior latent samples
    \(Z\sim q_\phi(\cdot\mid A,S)\) on the synthetic benchmark
    \textit{Sequential} task, colored by ground-truth mode label. At
    \(\beta=0.01\), the posterior separates according to the demonstrated
    mode. At \(\beta=0.1\), the posterior collapses into an overlapping
    cloud, consistent with the loss of \(I(M;Z\mid S)\) in
    Figure~\ref{fig:kl-info-decomposition}.}
    \label{fig:posterior-pca-sequential}
\end{figure}

\paragraph{Mode collapse under too strong regularization.}
Figure~\ref{fig:posterior-pca-sequential} provides a qualitative view of
the mechanism quantified in Figure~\ref{fig:kl-info-decomposition}.  We
visualize posterior latent samples \(Z\sim q_\phi(\cdot\mid A,S)\) on the
synthetic benchmark \textit{Sequential} using PCA, with points colored by
the ground-truth mode label. At \(\beta=0.01\), posterior latents form
mode-separated clusters, indicating that \(Z\) retains information about
which demonstrated branch is being reconstructed. At \(\beta=0.1\), the
posterior latents collapse into an overlapping cloud, so the latent no
longer separates the four modes. This visualizes the loss of
\(I(M;Z\mid S)\) under stronger pointwise KL regularization.

\paragraph{Heuristic for choosing the value for $\beta$.}
We present here the extended description of the heuristic and the results on the training sweep over the value of $\beta$. 
Let us choose a target posterior same-mode
decoding error of \(\bar\rho=0.1,\)
corresponding to at least $90\%$ same-mode reconstruction. The required information
level is
\begin{equation}\label{eq:estimate_of_Fano}
\widehat b_0(\bar\rho)
=
\frac{1}{Q}
\sum_{i:\widehat K_i>1}
\left[
\widehat H_i
-
H_b(\bar\rho)
-
\bar\rho\log(\widehat K_i-1)
\right].    
\end{equation}

We next construct an explicit collapsed comparator within the same CVAE decoder
family by setting the posterior equal to the prior and disabling the latent
conditioning of the decoder. Its pointwise KL is therefore zero by construction.
We denote its reconstruction loss by $C$. We obtain $C=0.459$ for Circle and $C=0.347$ for Sequential.
Because both dataset modes are exactly balanced, $b_0=0.368$ for Circle (K=2), $b_0=0.951$ for Sequential (K=4).
For any actual trained checkpoint satisfying
$L_\beta\leq C$,
we can directly verify
\(
B_{\hat\rho}
\leq I(M;Z\mid S)
\leq I(A;Z\mid S)
\leq D_{\mathrm{KL}}
\leq \frac{L_\beta}{\beta}
\leq \frac{C}{\beta}.
\)
It follows from the corollary that
achieving $\hat\rho\leq\bar\rho$ is impossible when \( \beta>\beta_{\max},
\beta_{\max}:=\frac{C}{b_0}.\)
This gives $\beta_{\max}=1.25$ for Circle, and
$\beta_{\max}=0.365$ for Sequential.

As mentioned before, this is a necessary condition: values below it are permitted, but are not guaranteed to work. To obtain a suitable value below this threshold, we use one low-$\beta$ calibration
run that correctly retains the mode information.  
Let its reconstruction loss and pointwise KL be
$R_p$ and $D_p$. Holding this observed solution fixed, its objective remains below
the collapsed comparator while
\(
R_p+\beta D_p<C.
\)
This gives the estimate
\(
\beta^*
=
\frac{C-R_p}{D_p}.
\)
We then use $\beta_{heu}=\beta^*/2$ as a safety margin.

For Circle, the calibration run at $\beta_0=0.1$ has
\(
R_p
=
0.3589-0.1(0.667)
=
0.2922,
\beta^*
=
\frac{0.459-0.2922}{0.667}
=
0.250.
\)
The resulting recommendation is $\beta_{heu}=0.125$. In the sweep, the target $\bar\rho$ is met at
$\beta=0.1$ and missed at $\beta=0.3117$.

For Sequential, the calibration run at $\beta_0=0.01$ has
\(
R_p
=
0.3218-0.01(1.437)
=
0.3074, 
\beta^*
=
\frac{0.347-0.3074}{1.437}
=
0.0275.
\)
The resulting recommendation is $\beta_{heu}=0.0138$. The target $\bar\rho$ is met at
$\beta=0.01$, missed at $\beta=0.02$, and complete latent collapse occurs at
$\beta=0.04$.

Additionally, we note that the pointwise KL value provides an early training-time failure diagnostic:
\[
D_{\mathrm{KL}}<b_0(\bar\rho)
\quad\Longrightarrow\quad
\hat\rho>\bar\rho.
\]
Therefore, if during training, the KL stabilizes below $b_0$, the current run cannot meet the
specified mode-recovery target and $\beta$ should be reduced.

Table \ref{sweepbeta} contains 20 runs across Circle and Sequential. In every run,
the empirical inequality
\(
B_{\hat\rho}\leq D_{\mathrm{KL}}\leq C/\beta
\)
holds. Every run achieving $\hat\rho\leq0.1$ has
$D_{\mathrm{KL}}\geq b_0$, every run with
$D_{\mathrm{KL}}<b_0$ misses the target, and no run above
$\beta_{\max}$ meets it. Several runs below $\beta_{\max}$ also fail, confirming
that the theoretical threshold is necessary rather than sufficient.
\begin{table}[t]
\centering
\small
\caption{
Full $\beta$-sweep on Circle ($K{=}2$, $\beta_{heu}=0.125$) and Sequential
($K{=}4$, $\beta_{heu}=0.0138$). Success is computed over the last five evaluation
epochs ($\pm$ s.d., $n{=}5$), measured with $200$ rollouts per evaluation.
}
\label{sweepbeta}

\begin{subtable}[t]{\linewidth}
\centering
\caption{Circle}
\label{tab:sweepbeta-circle}

\resizebox{\linewidth}{!}{%
\begin{tabular}{rrrrrrrccc}
\toprule
$\beta$
& $\beta/\beta_{\mathrm{max}}$
& $D_{\mathrm{KL}}$
& $I(M;Z\mid S)$
& $\hat\rho$
& $B_{\hat\rho}$
& $L_\beta$
& $L_\beta \le C$
& $\hat\rho \le 0.1$
& Success \\
\midrule
0.001  & 0.00 & 2.627 & 0.693 & 0.000 & 0.693 & 0.2658 & $\checkmark$ & $\checkmark$ & $0.766 \pm 0.054$ \\
0.01   & 0.01 & 1.246 & 0.691 & 0.001 & 0.689 & 0.2893 & $\checkmark$ & $\checkmark$ & $0.965 \pm 0.015$ \\
0.1    & 0.08 & 0.667 & 0.566 & 0.050 & 0.496 & 0.3589 & $\checkmark$ & $\checkmark$ & $0.901 \pm 0.019$ \\
0.3117 & 0.25 & 0.110 & 0.108 & 0.308 & 0.075 & 0.4489 & $\checkmark$ &              & $0.174 \pm 0.214$ \\
0.6234 & 0.50 & 0.000 & 0.000 & 0.502 & 0.000 & 0.4608 &              &              & $0.000 \pm 0.000$ \\
1      & 0.80 & 0.000 & 0.000 & 0.502 & 0.000 & 0.4584 & $\checkmark$ &              & $0.000 \pm 0.000$ \\
1.2468 & 1.00 & 0.000 & 0.000 & 0.504 & 0.000 & 0.4596 &              &              & $0.000 \pm 0.000$ \\
2.4935 & 2.00 & 0.000 & 0.000 & 0.495 & 0.000 & 0.4591 &              &              & $0.000 \pm 0.000$ \\
4.987  & 4.00 & 0.000 & 0.000 & 0.499 & 0.000 & 0.4591 &              &              & $0.000 \pm 0.000$ \\
\bottomrule
\end{tabular}%
}
\end{subtable}

\vspace{0.8em}

\begin{subtable}[t]{\linewidth}
\centering
\caption{Sequential}
\label{tab:sweepbeta-sequential}

\resizebox{\linewidth}{!}{%
\begin{tabular}{rrrrrrrccc}
\toprule
$\beta$
& $\beta/\beta_{\mathrm{max}}$
& $D_{\mathrm{KL}}$
& $I(M;Z\mid S)$
& $\hat\rho$
& $B_{\hat\rho}$
& $L_\beta$
& $L_\beta \le C$
& $\hat\rho \le 0.1$
& Success \\
\midrule
0.001  & 0.00 & 6.321 & 1.386 & 0.000 & 1.385 & 0.2598 & $\checkmark$ & $\checkmark$ & $0.918 \pm 0.018$ \\
0.01   & 0.03 & 1.437 & 1.268 & 0.039 & 1.179 & 0.3218 & $\checkmark$ & $\checkmark$ & $0.938 \pm 0.024$ \\
0.02   & 0.05 & 0.890 & 0.856 & 0.217 & 0.626 & 0.3385 & $\checkmark$ &              & $0.728 \pm 0.095$ \\
0.04   & 0.11 & 0.000 & 0.000 & 0.749 & 0.000 & 0.3477 &              &              & $0.000 \pm 0.000$ \\
0.0912 & 0.25 & 0.000 & 0.000 & 0.751 & 0.000 & 0.3481 &              &              & $0.000 \pm 0.000$ \\
0.1    & 0.27 & 0.000 & 0.000 & 0.747 & 0.000 & 0.3471 &              &              & $0.000 \pm 0.000$ \\
0.1824 & 0.50 & 0.000 & 0.000 & 0.747 & 0.000 & 0.3480 &              &              & $0.000 \pm 0.000$ \\
0.3648 & 1.00 & 0.000 & 0.000 & 0.752 & 0.000 & 0.3481 &              &              & $0.000 \pm 0.000$ \\
0.7296 & 2.00 & 0.000 & 0.000 & 0.749 & 0.000 & 0.3481 &              &              & $0.000 \pm 0.000$ \\
1      & 2.74 & 0.000 & 0.000 & 0.748 & 0.000 & 0.3471 &              &              & $0.000 \pm 0.000$ \\
1.4592 & 4.00 & 0.000 & 0.000 & 0.748 & 0.000 & 0.3480 &              &              & $0.000 \pm 0.000$ \\
\bottomrule
\end{tabular}%
}
\end{subtable}

\end{table}

\subsection{Aggregate Matching and Prior Coverage}
\label{app:results-agg-matching}
As discussed in Appendix~\ref{app:proofs-aggregate}, the critical-\(\beta\) collapse criterion is specific to pointwise KL and does not apply to aggregate matching. For aggregate-matched objectives, the relevant deployment question is whether prior samples cover the decoder-valid latent regions learned from posterior samples.

\paragraph{Posterior--prior valid-mass gap.}
We test the deployment caveat from Appendix~\ref{app:proofs-aggregate}:
aggregate matching can preserve mode information in the posterior, but
successful inference also requires the prior to place mass on
decoder-valid latent regions. We therefore compare posterior-sampled and
prior-sampled valid mass. Posterior evaluation samples
\(Z\sim q_\phi(\cdot\mid A,S)\) after drawing
\((S,A)\sim p_{\mathcal D}\), whereas prior evaluation samples
\(Z\sim p_\psi(\cdot\mid S)\), as at deployment. Since each synthetic
action chunk is a complete trajectory, valid mass is the probability that
the decoded trajectory satisfies the synthetic success criterion. We
report
\[
\begin{aligned}
\Delta_{\rm valid}
&=
\E_{(S,A)\sim p_{\mathcal D}}
\Pr_{Z\sim q_\phi(\cdot\mid A,S)}
\!\left[
f_\theta(Z,S)\in
\bigcup_{k=1}^{\mathsf K(S)}C_k(S)
\right] \\
&\quad -
\E_{S\sim p_{\mathcal D,S}}
\Pr_{Z\sim p_\psi(\cdot\mid S)}
\!\left[
f_\theta(Z,S)\in
\bigcup_{k=1}^{\mathsf K(S)}C_k(S)
\right].
\end{aligned}
\]
Table~\ref{tab:valid-mass-gap} shows that aggregate matching changes the
failure mode rather than removing it. KL-CVAE has the smallest gap,
consistent with pointwise KL keeping posterior latents close to the
prior, but it also has lower posterior valid mass. MMD-CWAE attains the
highest posterior valid mass, indicating that its posterior latents
contain useful decoder-valid regions, but it has the largest
posterior--prior gap. LAT-Flow gives the highest prior valid mass and a
smaller gap than MMD/Sinkhorn, suggesting that learning the prior
improves coverage of decoder-valid regions. We also find that this gap is
sensitive to latent-space regularization during training: removing these
regularizers substantially increases \(\Delta_{\rm valid}\), widening the
gap.

\begin{table}[t]
\centering
\small
\setlength{\tabcolsep}{5pt}
\caption{
Posterior--prior valid-mass gap on synthetic tasks. We report macro
averages across all four synthetic tasks and averages on the two hardest
\(\mathsf K=16\) tasks. Posterior evaluation samples
\(Z\sim q_\phi(\cdot\mid A,S)\) after drawing
\((S,A)\sim p_{\mathcal D}\); prior evaluation samples
\(Z\sim p_\psi(\cdot\mid S)\), as at deployment. A larger
\(\Delta_{\rm valid}\) indicates that posterior latents decode to valid
trajectories more often than prior samples.}
\label{tab:valid-mass-gap}
\begin{tabular*}{\linewidth}{@{\extracolsep{\fill}}lcccc@{}}
\toprule
Method
& Post. valid
& Prior valid
& Avg. \(\Delta_{\rm valid}\)
& Hard \(\Delta_{\rm valid}\) \\
\midrule
KL-CVAE        & $0.65$ & $0.61$ & $\mathbf{0.03}$ & $\mathbf{0.03}$ \\
MMD-CWAE       & $\mathbf{0.72}$ & $0.57$ & $0.14$ & $0.21$ \\
Sinkhorn-CWAE  & $0.69$ & $0.57$ & $0.12$ & $0.14$ \\
LAT-Flow       & $0.70$ & $\mathbf{0.64}$ & $0.06$ & $0.09$ \\
\bottomrule
\end{tabular*}
\end{table}

\begin{figure}[t]
    \centering

    \begin{minipage}[t]{0.40\linewidth}
        \centering
        \vspace{0pt}

        \resizebox{\linewidth}{!}{%
            \begin{tabular}{lcc}
                \toprule
                Method
                & Avg. \(\Delta_{\rm valid}\)
                & Hard \(\Delta_{\rm valid}\) \\
                \midrule
                KL-CVAE        & $0.04$ & $0.05$ \\
                MMD-CWAE       & $0.32$ & $0.57$ \\
                Sinkhorn-CWAE  & $0.32$ & $0.58$ \\
                LAT-Flow       & $0.18$ & $0.30$ \\
                \bottomrule
            \end{tabular}%
        }

        \captionof{table}{
            Effect of removing latent-space regularizers. We remove dropout,
            decoder latent jitter, and the posterior geometry regularizer, and
            report the posterior--prior valid-mass gap. Larger values indicate
            that posterior latents decode to valid trajectories more often than
            prior samples, revealing a prior-coverage failure.
        }
        \label{tab:posterior-geometry-ablation}
    \end{minipage}
    \hfill
    \begin{subfigure}[t]{0.27\linewidth}
        \centering
        \vspace{0pt}
        \includegraphics[width=\linewidth]{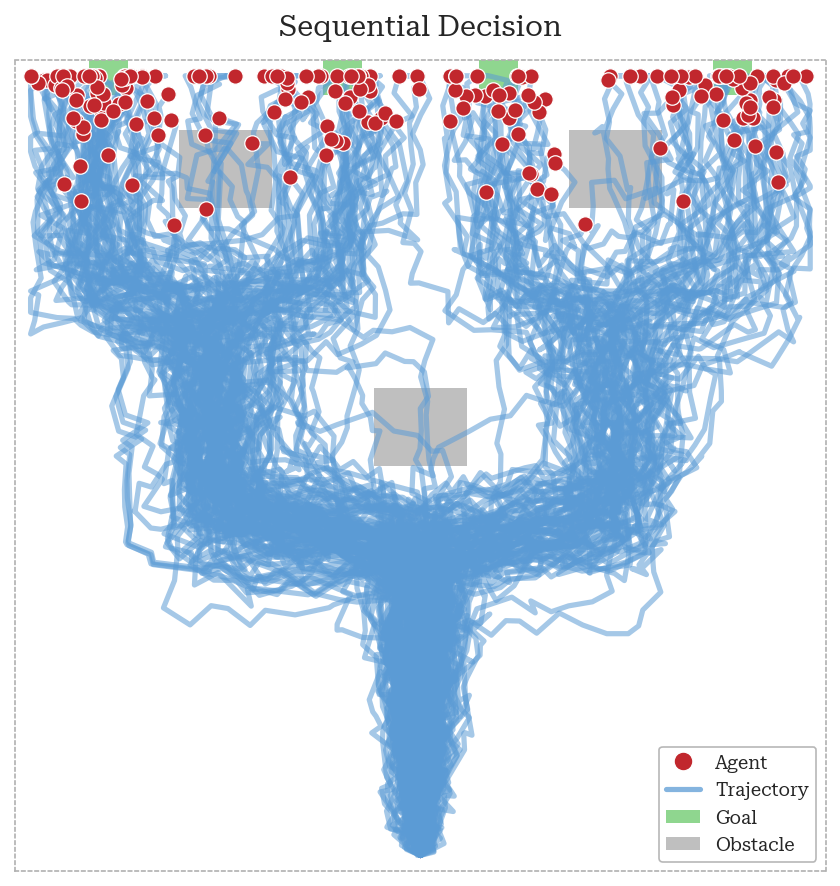}
        \caption{Without latent-space regularizers}
    \end{subfigure}
    \hfill
    \begin{subfigure}[t]{0.27\linewidth}
        \centering
        \vspace{0pt}
        \includegraphics[width=\linewidth]{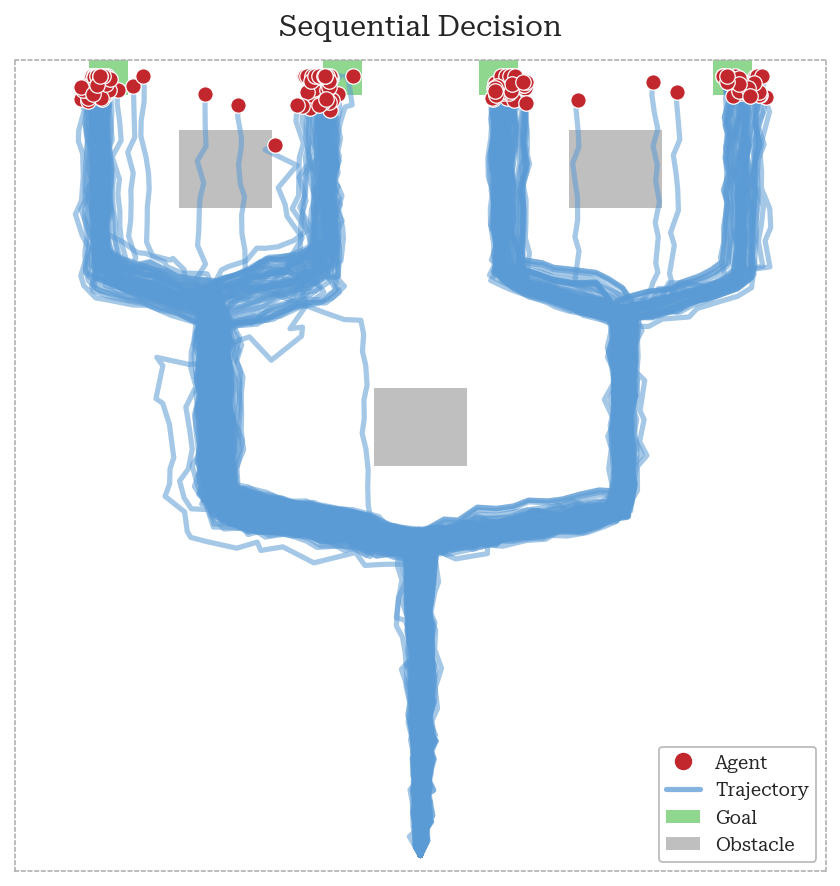}
        \caption{With latent-space regularizers}
    \end{subfigure}

    \caption{
        \textbf{Effect of latent-space regularization on Sinkhorn-CWAE.}
        Prior-sampled trajectories on the \textit{Sequential} task. Each policy
        is rolled out 200 times. Without dropout, decoder latent jitter, and
        posterior geometry regularization, prior samples generate many dispersed
        and invalid trajectories. With the regularizers enabled, prior samples
        concentrate on valid trajectory branches and reach the goal regions more
        consistently. This qualitative behavior is consistent with
        Table~\ref{tab:posterior-geometry-ablation}: removing the regularizers
        increases the posterior--prior valid-mass gap, revealing that the
        posterior can define useful latent regions while the deployment prior
        fails to sample them reliably.
    }
    \label{fig:sinkhorn-cwae-regularizers}

\end{figure}

\paragraph{Posterior-geometry ablation.}
Aggregate matching constrains a batch-level latent distribution, but it does not by itself control whether the deterministic posterior codes have a geometry that is easy for the prior to sample. We therefore use dropout, decoder latent jitter, and posterior geometry regularization mentioned in \ref{app:method-posterior-geometry} to stabilize the posterior regions seen by the decoder. Table~\ref{tab:posterior-geometry-ablation} ablates these regularizers. Removing them increases the posterior--prior valid-mass gap, especially on the \(\mathsf K=16\) tasks. The effect is largest for MMD-CWAE and Sinkhorn-CWAE: their posterior latents can still decode to valid trajectories, but prior samples miss the same decoder-valid
regions much more often. Figure~\ref{fig:sinkhorn-cwae-regularizers} shows the same failure mode qualitatively for Sinkhorn-CWAE on the synthetic \textit{Sequential} task. Without posterior regularization, prior-sampled reconstructions degrade
quickly and become highly noisy. With these regularizers, prior-sampled
trajectories become substantially sharper and remain concentrated on the
valid branches.

\begin{table}[t]
\centering
\small
\setlength{\tabcolsep}{5pt}
\caption{
Conditional aggregate-matching ablation for Sinkhorn-CWAE on PushT Image.
Posterior $L_1$ loss evaluates chunks decoded from
\(Z\sim q_\phi(\cdot\mid A,S)\); prior-sample $L_1$ loss evaluates chunks decoded
from \(Z\sim p_\psi(\cdot\mid S)\), as at deployment. We report
\(\Delta_{\rm L_1}=\mathrm{L_1}_{\rm prior}-\mathrm{L_1}_{\rm post}\).
}
\label{tab:conditional-ot-ablation}
\begin{tabular*}{\linewidth}{@{\extracolsep{\fill}}lcccc@{}}
\toprule
OT variant
& Final IoU \(\uparrow\)
& Post. $L_1$ \(\downarrow\)
& Prior $L_1$ \(\downarrow\)
& \(\Delta_{\rm L_1}\downarrow\) \\
\midrule
Fixed Gaussian
& $0.0004$ & $1.9829$ & $38.8889$ & $36.9060$ \\
Learned prior
& $0.4385$ & $2.0476$ & $\mathbf{16.8767}$ & $\mathbf{14.8290}$ \\
Conditional learned prior
& $\mathbf{0.7517}$ & $2.0422$ & $20.1505$ & $18.1084$ \\
\bottomrule
\end{tabular*}
\end{table}

\paragraph{Conditional aggregate matching}
As mentioned in \ref{app:method-conditional-cwae}, in simulation benchmarks the observation \(s\) changes the valid action
distribution, so marginal latent matching can mix posterior and prior
samples across incompatible conditioning states. 
Table~\ref{tab:conditional-ot-ablation} ablates the prior parameterization used by Sinkhorn-CWAE on PushT Image environment. All variants obtain
similar posterior-sampled $L_1$ loss, indicating that the posterior encoder
and decoder can reconstruct the demonstrated chunks. The difference
appears at deployment: the fixed Gaussian prior produces very poor
prior-sampled chunks and near-zero rollout performance. Learning the
prior substantially reduces the posterior--prior sample gap and improves
final IoU. Adding conditional aggregate matching gives the best closed-loop
performance, suggesting that matching the state--latent coupling is more important for deployment than marginal latent matching alone.

\subsection{Action-Space Bridge--Smoothness Trade-off}

\begin{figure}[t]
    \centering
    \includegraphics[width=\linewidth]{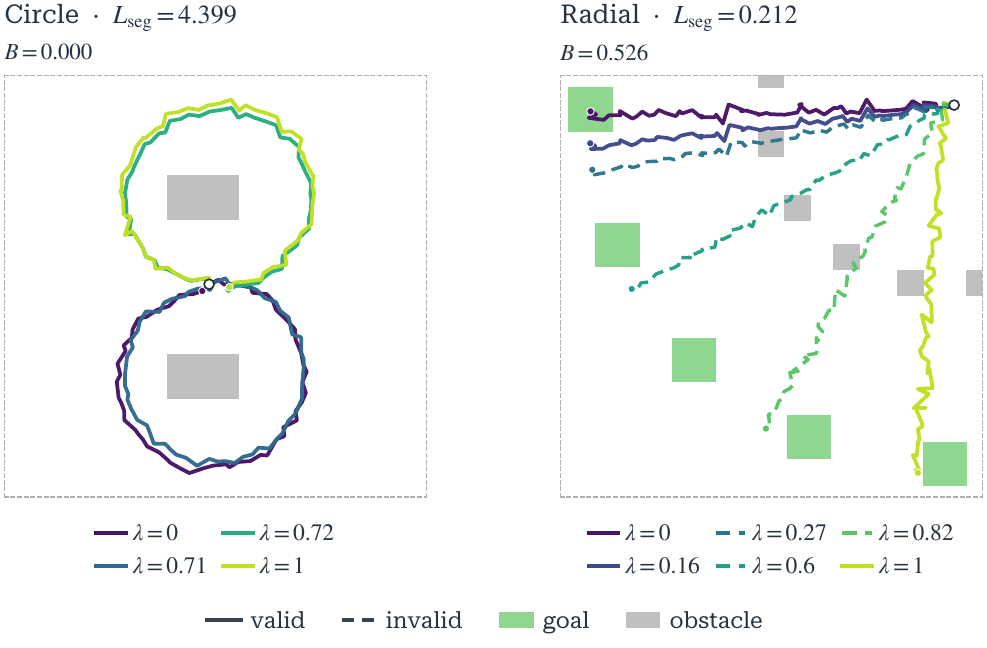}
\caption{ \textbf{Action-space bridge--smoothness trade-off.} Cross-mode interpolations in base space. Colors denote \(\lambda\); dashed curves are invalid trajectories and therefore part of the bridge mass.
}
\label{fig:action-bridge-bound}
\end{figure}

\begin{table}[t]
  \centering
  \caption{
  Bridge-smoothness trade-off for action-space samplers on synthetic tasks.
  }
  \label{tab:action-bridge-diagnostic}
  \begin{tabular*}{\linewidth}{@{\extracolsep{\fill}}llccccc@{}}
  \toprule
  Method & Task & $\mathsf K$ & $B$
  & $\|u_{\lambda_+}-u_{\lambda_-}\|$ & $L_{\rm seg}$ & $L_{\rm seg}/\Delta$ \\
  \midrule
  Act-Flow & Circle     & 2  & $0.0000$ & $0.155$  & $4.3985$ & $7.92$  \\
  Act-Flow & Sequential & 4  & $0.0013$ & $1.910$  & $2.3893$ & $17.65$ \\
  Act-Flow & Corridor   & 16 & $0.5851$ & $12.687$ & $0.0405$ & $1.67$  \\
  Act-Flow & Radial     & 16 & $0.5259$ & $10.558$ & $0.2119$ & $2.63$  \\
  \midrule
  Act-Diff & Circle     & 2  & $0.0000$ & $0.155$  & $4.2850$ & $7.71$  \\
  Act-Diff & Sequential & 4  & $0.0005$ & $1.833$  & $2.7834$ & $20.56$ \\
  Act-Diff & Corridor   & 16 & $0.5779$ & $11.908$ & $0.0520$ & $2.14$  \\
  Act-Diff & Radial     & 16 & $0.5475$ & $10.730$ & $0.2786$ & $3.46$  \\
  \bottomrule
  \end{tabular*}
  \end{table}
\label{app:action-bridge-diagnostic}
\paragraph{Estimating the Lipschitz constant.}
We estimate the Lipschitz constant using finite differences. From a fixed \(s\), we sample a pool of base noises \(u\sim \mathcal N(0,\mathbf I)\), and decode trajectories $G_{\theta, s}(u)$. We group the input noises of the valid decoded trajectories by output mode, and from each mode pair we randomly select a pair of noises $(u_1,u_2)$, $G(u_1)\in C_i,G(u_2)\in C_j$. 
For each sampled cross-mode pair $(u_1,u_2)$, we evaluate
$G_{\theta,s}(u_\lambda)$ on a fixed grid of $N=101$ interpolation
coefficients $\lambda\in[0,1]$, where
$u_\lambda=(1-\lambda)u_1+\lambda u_2$.
We identify $\lambda_+$ as the first grid point whose decoded
trajectory is valid in the second mode, and $\lambda_-$ as
the last preceding grid point whose decoded trajectory is valid
in the first mode.
We compute the finite-difference sensitivity
\begin{equation}\label{eq:sseg}
    S_{\rm seg}
=
\frac{
d_{\rm traj}\!\left(G_{\theta,s}(u_{\lambda_+}),
G_{\theta,s}(u_{\lambda_-})\right)}
{\|u_{\lambda_+}-u_{\lambda_-}\|}, 
\end{equation}
where \(d_{\rm traj}\) is the root-mean-square pointwise Euclidean
distance between trajectories.
The maximum $L_{seg}:=\max(S_{seg})$ is a lower bound of the Lipschitz constant $L_{\theta, s}$. 
\paragraph{Full table results}
For each action-space
sampler, we use its native inference setting: 10 solver steps for
Act-Flow and 100 DDIM steps for Act-Diff. For each task, we fix the
initial observation, sample \(4096\) base noises, select \(100\)
cross-mode valid pairs, and evaluate \(101\) interpolation points per
pair. We obtain
the mode separation $\Delta_{ij}$ as the
$d_{\rm traj}$ from the synthetic ground-truth trajectories
for the two modes, generated from the known task geometry. Table~\ref{tab:action-bridge-diagnostic} reports the results and the Act-Flow values used for Figure~\ref{fig:action-bridge-bound}.

\subsection{Measuring Multimodality in Robot Benchmark Datasets}
\label{app:results-data-ambiguity}
Estimating conditional action multimodality requires distinguishing alternative actions at the same semantic state from action variation caused by differences between nearby but semantically different states. Directly clustering action
chunks collected across a neighborhood can conflate these two
sources of variation. We therefore fit a local joint state--action Gaussian mixture model (GMM) and condition it analytically on a query state to account for local state--action dependence. We then apply modal clustering to the resulting conditional density, identifying modes by their attraction basins~\citep{chacon2019mixture}.
This method allows several GMM components to describe a single action mode and mitigates over-segmentation caused by treating each Gaussian component as a distinct mode.

For each query proprioceptive state $s_i$, we retrieve nearby states from other demonstrations (with the same task or language instruction, where
applicable). Both the query and its candidate neighbors must have a complete future chunk of $\mathsf H_{\mathrm{diag}}$ recorded actions contained within their respective demonstrations.
We denote the set of eligible state indices by
$\mathcal I_{\mathsf H_{\mathrm{diag}}}$.
State coordinates are standardized using dataset-level moments within each task group. We use the dimension-normalized distance
\[
d_{\mathrm{state}}(i,j)
=
\frac{\|\widetilde s_i-\widetilde s_j\|_2}{\sqrt{d_s}},
\]
where $d_s$ is the state dimension.
Each other demonstration contributes at most its nearest eligible
state; the query demonstration is excluded.
Let $\mathcal J_i$ denote the indices retained by this selection.
For a radius $\varepsilon$, the neighborhood is
\[
\mathcal N_\varepsilon(i)
=
\left\{
j\in\mathcal J_i:
d_{\mathrm{state}}(i,j)\le\varepsilon
\right\},
\qquad
n_i=|\mathcal N_\varepsilon(i)|.
\]
The main results use $\varepsilon=0.5$.
Sensitivity checks use $\varepsilon\in\{0.3,0.5,1.0\}$.
\paragraph{Local joint density and conditioning.}
Within each neighborhood, we flatten the future action chunks
and fit a local principal component analysis (PCA) projection
with $d_y=2$ components. For $n_i=2$, we retain one PCA component;
neighborhoods with fewer than two demonstrations are excluded
from density fitting.
State coordinates retain their global standardization and are
only centered locally.
Let $x$ denote the resulting state coordinates and $y$ the PCA
coordinates of the corresponding future action chunk.
We fit a GMM to the concatenated state--action vectors
$[x^\top,y^\top]^\top$, obtaining the local joint density
\[
\widehat p_i(x,y)
=
\sum_{g=1}^{G_i}\pi_g\,
\mathcal N\!\left(
\begin{bmatrix}x\\y\end{bmatrix};
\begin{bmatrix}\mu_g^x\\\mu_g^y\end{bmatrix},
\begin{bmatrix}
\Sigma_g^{xx}&\Sigma_g^{xy}\\
\Sigma_g^{yx}&\Sigma_g^{yy}
\end{bmatrix}
\right),
\]
where $G_i$ is the number of Gaussian components and $\pi_g$
are their mixture weights.

Candidate GMMs are fitted by expectation--maximization (EM).
We compare $G=1,\ldots,\min(20,n_i)$ components and two covariance
structures: a separate full covariance matrix for each component,
or a single full covariance matrix shared across components.
For each configuration, we compare two initialization procedures.
The first uses three $k$-means starts in joint state--action space.
The second first fits an action-only full-covariance GMM with three
starts, uses its component-membership probabilities
(soft responsibilities) to initialize the joint mixture parameters,
and performs one joint EM fit.

We select the candidate with the smallest Bayesian information
criterion (BIC),
\[
-2\sum_{j\in\mathcal N_\varepsilon(i)}
\log\widehat p_i(x_j,y_j)
+
p\log n_i,
\]
where $p$ is the number of free parameters.
This selects both the number of Gaussian components and the
covariance structure for each neighborhood.

We then analytically condition the selected joint GMM at the query
coordinate $x_i$, obtaining
\[
\widehat p_i(y\mid x_i)
=
\sum_g w_g(x_i)\,
\mathcal N\!\left(y;\mu_{g\mid i},\Sigma_{g\mid i}\right),
\]
where
\begin{align*}
w_g(x_i)
&=
\frac{\pi_g\mathcal N(x_i;\mu_g^x,\Sigma_g^{xx})}
{\sum_h\pi_h\mathcal N(x_i;\mu_h^x,\Sigma_h^{xx})},\\
\mu_{g\mid i}
&=
\mu_g^y+
\Sigma_g^{yx}(\Sigma_g^{xx})^{-1}(x_i-\mu_g^x),\\
\Sigma_{g\mid i}
&=
\Sigma_g^{yy}-
\Sigma_g^{yx}(\Sigma_g^{xx})^{-1}\Sigma_g^{xy}.
\end{align*}

\paragraph{Modal clusters and entropy.}
We identify modes of the conditional action density using
~\cite{chacon2019mixture}'s Algorithm~2.
A modal cluster consists of points that converge to the same
density maximum and may contain probability mass contributed
by several Gaussian components.

We draw $R=4096$ samples from $\widehat p_i(y\mid x_i)$.
For each sample, we first select a component according to
$w_g(x_i)$ and then sample from its conditional Gaussian.
Gaussian-mixture mean shift iteratively moves each sample toward
a density maximum; samples reaching the same maximum are assigned
to the same modal cluster. Endpoints are checked for negative curvature, and numerically
identical maxima are merged.
The probability mass of each modal cluster is estimated by the
fraction of conditional samples assigned to it:
\[
\widehat m_{i\ell}
=
\frac{1}{R}
\sum_{r=1}^{R}
\mathbf 1\{
y_i^{(r)}\text{ converges to mode }\ell
\}.
\]
We retain modal clusters with estimated mass at least $2.5\%$,
\[
\mathcal R_i
=
\{\ell:\widehat m_{i\ell}\ge0.025\},
\]
and report their count and entropy:
\[
\widehat K(s_i)
=
|\mathcal R_i|,
\qquad
\widehat H_i
=
-\sum_{\ell\in\mathcal R_i}
\widetilde m_{i\ell}\log_2\widetilde m_{i\ell},
\qquad
\widetilde m_{i\ell}
=
\frac{
\widehat m_{i\ell}
}{
\sum_{j\in\mathcal R_i}\widehat m_{ij}
}.
\]The probability cutoff is applied to modal-cluster mass after
conditioning and clustering; retained masses are renormalized for entropy.

\paragraph{Dataset protocols.}
For the synthetic validation, we use 100 demonstration-initial query states
per benchmark and $H=60$. 
The robot dataset fits use query states across demonstration trajectories,
with $H\in\{10,30\}$.
We sample 1000 query starts per dataset, without replacement and stratified by task. We allocate at least 16 queries per language task
when available, then distribute remaining queries according to the task's
state population. 

\begin{table}[t]
    \centering
    \small
    \setlength{\tabcolsep}{4pt}
    \caption{Dataset-wide conditional-mode $\hat K$ and conditional entropy $\hat H$ estimates.
    Entries correspond to radii $\varepsilon=0.3/0.5/1.0$.
    The modal-mass cutoff is $2.5\%$; entropy is in bits.}
    \label{tab:sim-mbg-dataset-modes}
    \begin{tabular*}{\linewidth}{@{\extracolsep{\fill}}crr@{}}
    \toprule
    Horizon & $\widehat K$ & $\widehat H$ \\
    \midrule
    \multicolumn{3}{@{}l}{\textbf{UR3 BlockPush}} \\
    10 & $1.128/1.136/1.110$ & $0.080/0.082/0.068$ \\
    30 & $1.159/1.164/1.098$ & $0.098/0.103/0.055$ \\
    \midrule
    \multicolumn{3}{@{}l}{\textbf{Kitchen}} \\
    10 & $1.000/1.143/1.111$ & $0.000/0.083/0.071$ \\
    30 & $1.000/1.231/1.145$ & $0.000/0.147/0.099$ \\
    \midrule
    \multicolumn{3}{@{}l}{\textbf{PushT}} \\
    10 & $1.022/1.068/1.144$ & $0.013/0.038/0.087$ \\
    30 & $1.000/1.005/1.104$ & $0.000/0.004/0.059$ \\
    \midrule
    \multicolumn{3}{@{}l}{\textbf{LIBERO}} \\
    10 & $1.003/1.014/1.028$ & $0.003/0.007/0.016$ \\
    30 & $1.023/1.027/1.033$ & $0.015/0.017/0.020$ \\
    \midrule
    \multicolumn{3}{@{}l}{\textbf{MetaWorld}} \\
    10 & $1.009/1.037/1.069$ & $0.004/0.020/0.040$ \\
    30 & $1.022/1.067/1.123$ & $0.016/0.043/0.073$ \\
    \midrule
    \multicolumn{3}{@{}l}{\textbf{Bimodal Tissue Grasping}} \\
    10 & $1.486/1.468/1.454$ & $0.449/0.432/0.419$ \\
    30 & $2.267/2.267/2.267$ & $0.760/0.760/0.760$ \\
    \bottomrule
    \end{tabular*}
  \end{table}
\paragraph{Results.}
Table \ref{tab:sim-mbg-dataset-modes} shows the complete results with varying neighborhood radius. The simulation estimates are generally close to one, while Bimodal Tissue Grasping, given its proof-of-concept nature, exhibits stronger conditional multimodality. Ambiguity
  may be concentrated at initial or branching states, with subsequent
  states admitting largely unambiguous continuations. Once a policy
  commits to a branch, potentially through small prediction or execution
  deviations, replanning can guide it along a coherent expert strategy.

\clearpage

\subsection{Simulation and Real-World Robot Results}
\label{app:benchmark-results}
We provide the complete results from Section~\ref{sec:sim-results}.
Table~\ref{tab:sim_reeval} reports performance at the final and
best checkpoints, allowing the best performance achieved during
training to be distinguished from subsequent degradation.
Table~\ref{tab:mode_entropy} reports diversity across completed
behavior sequences for simulation benchmarks with available
ground-truth behavior information or reference entropy values.
Table~\ref{tab:libero_metaworld_subtasks} reports performance on
the language-conditioned LIBERO and Meta-World tasks.

Table~\ref{tab:rollout_metrics} reports the complete results for
the bimodal tissue-grasping robot task, including task success,
conditional mode entropy, inter-mode trajectory separation, and
invalid-rollout counts. A rollout is classified as invalid if
the policy grasps the delicate region of interest that should
be avoided. A rollout is classified as a failure if it is invalid
or does not complete a successful grasp within a reasonable time.
Finally, Table~\ref{tab:throughput} reports the inference throughput
of the different policy parametrizations under the evaluation
settings used in simulation.

\begin{table*}[h]
\centering
\fontsize{9}{11}\selectfont
\caption{Simulation evaluation. 'LP' stands for 'learned prior' variant. When not specified, prior is a fixed standard Gaussian. Values are mean $\pm$ standard deviation
across three evaluation seeds, with 200 rollouts per seed. Final denotes
epoch 1000; Best is selected among checkpoints evaluated every 100 epochs,
with its epoch in superscript. A dash denotes Best coinciding with Final.
Bold indicates the highest reported mean for each benchmark.}
\label{tab:sim_reeval}
\setlength{\tabcolsep}{3pt}
\renewcommand{\arraystretch}{1.08}

\begin{tabularx}{\linewidth}{@{}>{\raggedright\arraybackslash}p{0.27\linewidth} l *{3}{>{\centering\arraybackslash}X}@{}}
\toprule
\textbf{Policy} &
& \shortstack{\textbf{PushT State}\\$(/1)$}
& \shortstack{\textbf{PushT Image}\\$(/1)$}
& \shortstack{\textbf{Kitchen State}\\$(/4)$} \\
\midrule
BCAT & Final
& $0.723{\pm}0.046$ & $0.663{\pm}0.012$ & $2.76{\pm}0.01$ \\
& Best
& -- & $(0.692{\pm}0.011)^{700}$ & $(2.78{\pm}0.01)^{500}$ \\
\addlinespace[3pt]
KL-CVAE & Final
& $0.625{\pm}0.024$ & $0.662{\pm}0.016$ & $1.78{\pm}0.08$ \\
& Best
& $(0.635{\pm}0.021)^{900}$ & -- & -- \\
\addlinespace[3pt]
KL-CVAE-LP & Final
& $0.765{\pm}0.013$ & $0.749{\pm}0.022$ & $3.06{\pm}0.07$ \\
& Best
& $(0.796{\pm}0.019)^{400}$ & $\mathbf{(0.793{\pm}0.021)}^{900}$ & -- \\
\addlinespace[3pt]
MMD-CWAE-LP & Final
& $0.745{\pm}0.003$ & $0.647{\pm}0.016$ & $2.97{\pm}0.04$ \\
& Best
& -- & -- & $(3.07{\pm}0.07)^{400}$ \\
\addlinespace[3pt]
Sinkhorn-CWAE-LP & Final
& $0.694{\pm}0.034$ & $0.744{\pm}0.029$ & $3.21{\pm}0.10$ \\
& Best
& $(0.725{\pm}0.024)^{500}$ & -- & $\mathbf{(3.26{\pm}0.02)}^{300}$ \\
\addlinespace[3pt]
Act-Flow & Final
& $\mathbf{0.825{\pm}0.014}$ & $0.778{\pm}0.014$ & $2.53{\pm}0.07$ \\
& Best
& -- & $(0.781{\pm}0.014)^{800}$ & $(2.83{\pm}0.15)^{200}$ \\
\addlinespace[3pt]
LAT-Flow & Final
& $0.682{\pm}0.027$ & $0.724{\pm}0.023$ & $3.06{\pm}0.08$ \\
& Best
& $(0.703{\pm}0.006)^{500}$ & $(0.747{\pm}0.026)^{400}$ & -- \\
\bottomrule
\end{tabularx}

\par\vspace{8pt}

\begin{tabularx}{\linewidth}{@{}>{\raggedright\arraybackslash}p{0.27\linewidth} l *{2}{>{\centering\arraybackslash}X}@{}}
\toprule
\textbf{Policy} &
& \shortstack{\textbf{Kitchen Image}\\$(/4)$}
& \shortstack{\textbf{UR3 BlockPush}\\$(/2)$} \\
\midrule
BCAT & Final
& $0.53{\pm}0.05$ & $\mathbf{1.99{\pm}0.01}$ \\
& Best
& $(1.86{\pm}0.04)^{600}$ & -- \\
\addlinespace[3pt]
KL-CVAE & Final
& $1.12{\pm}0.05$ & $1.92{\pm}0.02$ \\
& Best
& -- & $(1.96{\pm}0.03)^{100}$ \\
\addlinespace[3pt]
KL-CVAE-LP & Final
& $0.94{\pm}0.01$ & $1.65{\pm}0.03$ \\
& Best
& $(1.70{\pm}0.04)^{600}$ & -- \\
\addlinespace[3pt]
MMD-CWAE-LP & Final
& $1.28{\pm}0.01$ & $1.88{\pm}0.01$ \\
& Best
& $\mathbf{(2.67{\pm}0.03)}^{500}$ & -- \\
\addlinespace[3pt]
Sinkhorn-CWAE-LP & Final
& $1.31{\pm}0.02$ & $1.63{\pm}0.02$ \\
& Best
& $(2.04{\pm}0.06)^{800}$ & $(1.92{\pm}0.01)^{200}$ \\
\addlinespace[3pt]
Act-Flow & Final
& $2.33{\pm}0.04$ & $0.47{\pm}0.01$ \\
& Best
& $(2.58{\pm}0.06)^{900}$ & $(1.61{\pm}0.02)^{200}$ \\
\addlinespace[3pt]
LAT-Flow & Final
& $1.09{\pm}0.05$ & $1.49{\pm}0.02$ \\
& Best
& $(1.86{\pm}0.03)^{900}$ & $(1.53{\pm}0.01)^{300}$ \\
\bottomrule
\end{tabularx}
\end{table*}

\begin{table}[h]
\centering
\caption{Behavioral entropy with seed-level uncertainty, following the QFAT/VQ-BeT
convention: Shannon entropy (bits) of the empirical distribution over completed
behavior sequences, $H = -\sum_{m}\hat p(m)\log_2 \hat p(m)$. Mean $\pm$ std over 3 evaluation seeds
$\times$ 200 rollouts; Final and Best as in Table 3.}
\label{tab:mode_entropy}
\setlength{\tabcolsep}{3pt}
\resizebox{\textwidth}{!}{%
\begin{tabular}{l cc cc cc}
\toprule
& \multicolumn{2}{c}{\textbf{Kitchen State}} & \multicolumn{2}{c}{\textbf{Kitchen Image}} & \multicolumn{2}{c}{\textbf{UR3 BlockPush}} \\
\cmidrule(lr){2-3} \cmidrule(lr){4-5} \cmidrule(lr){6-7}
\textbf{Policy} & Final & Best & Final & Best & Final & Best \\
\midrule
BCAT             & $3.43 \pm 0.05$          & $3.04 \pm 0.00^{(500)}$          & $1.81 \pm 0.26$          & $1.28 \pm 0.62^{(600)}$          & $0.98 \pm 0.00$          & -- \\
KL-CVAE          & $3.09 \pm 0.14$          & --          & $1.52 \pm 0.52$          & --         & $\mathbf{0.99 \pm 0.00}$ & $0.93 \pm 0.01^{(100)}$ \\
KL-CVAE-LP       & $\mathbf{3.76 \pm 0.09}$ & --                               & $0.00 \pm 0.00$          & $1.71 \pm 0.26^{(600)}$          & $0.97 \pm 0.02$          & -- \\
MMD-CWAE-LP      & $2.48 \pm 0.07$          & $2.48 \pm 0.10^{(400)}$          & $0.00 \pm 0.00$          & $1.98 \pm 0.12^{(500)}$          & $0.98 \pm 0.00$          & -- \\
Sinkhorn-CWAE-LP & $2.66 \pm 0.16$          & $2.62 \pm 0.04^{(300)}$          & $0.00 \pm 0.00$          & $1.51 \pm 0.11^{(800)}$          & $0.71 \pm 0.04$          & $\mathbf{0.99 \pm 0.00}^{(200)}$ \\
Act-Flow         & $3.08 \pm 0.37$          & $2.84 \pm 0.17^{(200)}$          & $\mathbf{2.60 \pm 0.20}$ & ${2.36 \pm 0.27}^{(900)}$ & $0.00 \pm 0.00$          & $\mathbf{0.99 \pm 0.01}^{(200)}$ \\
LAT-Flow         & $3.28 \pm 0.15$          & -- & $0.00 \pm 0.00$          & $2.17 \pm 0.30^{(900)}$          & $0.07 \pm 0.07$          & $0.03 \pm 0.05^{(300)}$ \\
\bottomrule
\end{tabular}}
\end{table}

    \begin{table*}[t]
  \centering
  \caption{
  Success rates (\%) by LIBERO subtask suite and MetaWorld task difficulty. Policies are all evaluated on the last epoch checkpoint.
  Values are mean $\pm$ standard deviation over three seeds, 400 rollouts.
  }
  \label{tab:libero_metaworld_subtasks}

  \setlength{\tabcolsep}{3pt}
  \renewcommand{\arraystretch}{1.12}

  \resizebox{\linewidth}{!}{%
  \begin{tabular}{@{}lccccccc@{}}
  \toprule
  Benchmark
  & BCAT
  & \shortstack{KL-\\CVAE}
  & \shortstack{KL-CVAE-\\LP}
  & \shortstack{MMD-CWAE-\\LP}
  & \shortstack{Sinkhorn-CWAE-\\LP}
  & \shortstack{Act-\\Flow}
  & \shortstack{LAT-\\Flow} \\
  \midrule

  \multicolumn{8}{l}{\textit{LIBERO}} \\

  Long
  & $86.33{\pm}1.25$
  & $\mathbf{92.33{\pm}0.94}$
  & $83.67{\pm}0.47$
  & $89.00{\pm}0.00$
  & $90.00{\pm}0.82$
  & $87.00{\pm}3.74$
  & $88.67{\pm}0.47$ \\

  Spatial
  & $92.33{\pm}1.25$
  & $\mathbf{98.67{\pm}1.25}$
  & $96.00{\pm}0.82$
  & $\mathbf{98.67{\pm}0.94}$
  & $96.00{\pm}0.00$
  & $93.67{\pm}0.47$
  & $94.00{\pm}2.16$ \\

  Goal
  & $96.00{\pm}0.82$
  & $\mathbf{97.33{\pm}1.25}$
  & $95.67{\pm}0.47$
  & $96.00{\pm}0.82$
  & $96.00{\pm}0.82$
  & $91.00{\pm}2.16$
  & $94.33{\pm}1.25$ \\

  Object
  & $99.67{\pm}0.47$
  & $\mathbf{100.00{\pm}0.00}$
  & $98.33{\pm}0.47$
  & $\mathbf{100.00{\pm}0.00}$
  & $97.67{\pm}0.47$
  & $97.67{\pm}0.94$
  & $99.00{\pm}0.82$ \\

  \cmidrule(lr){1-8}
  \textbf{Total}
  & $93.58{\pm}0.66$
  & $\mathbf{97.08{\pm}0.82}$
  & $93.42{\pm}0.31$
  & $95.92{\pm}0.31$
  & $94.92{\pm}0.24$
  & $92.33{\pm}0.85$
  & $94.00{\pm}0.35$ \\

  \addlinespace[3pt]
  \multicolumn{8}{l}{\textit{MetaWorld}} \\

  Easy
  & $61.55{\pm}0.44$
  & $71.78{\pm}1.62$
  & $\mathbf{81.31{\pm}1.22}$
  & $68.45{\pm}1.38$
  & $72.26{\pm}0.73$
  & $39.28{\pm}1.27$
  & $69.41{\pm}2.19$ \\

  Medium
  & $46.97{\pm}2.14$
  & $55.15{\pm}0.42$
  & $\mathbf{58.49{\pm}2.81}$
  & $43.33{\pm}2.14$
  & $37.27{\pm}3.86$
  & $18.79{\pm}4.35$
  & $37.27{\pm}1.97$ \\

  Hard
  & $69.45{\pm}2.08$
  & $75.00{\pm}1.36$
  & $71.67{\pm}2.72$
  & $57.22{\pm}6.98$
  & $\mathbf{78.33{\pm}6.81}$
  & $34.45{\pm}3.93$
  & $73.33{\pm}2.72$ \\

  Very Hard
  & $46.00{\pm}6.53$
  & $\mathbf{73.33{\pm}3.77}$
  & $65.33{\pm}1.89$
  & $68.67{\pm}2.49$
  & $58.00{\pm}4.32$
  & $38.00{\pm}4.32$
  & $49.33{\pm}5.25$ \\

  \cmidrule(lr){1-8}
  \textbf{Total}
  & $57.73{\pm}0.52$
  & $68.67{\pm}1.41$
  & $\mathbf{73.53{\pm}0.34}$
  & $61.60{\pm}1.18$
  & $63.87{\pm}1.52$
  & $34.07{\pm}2.39$
  & $60.80{\pm}0.99$ \\

  \bottomrule
  \end{tabular}%
  }
  \end{table*}

\begin{table}[h]
\centering
\small
\caption{On-robot rollout metrics on the bimodal
bowel-grasping task, with 20 recorded rollouts per policy. Mode entropy $H(M\mid S=s_0)$ is conditioned on the initial state. Inter-mode width is the
Euclidean distance between the two per-mode mean trajectories. Invalid denotes a mode-average trajectory into the invalid region of interest, while general failure in reaching/grasping is incorporated in the success rate.}
\label{tab:rollout_metrics}
\begin{tabular}{lrrrrrr}
\toprule
Policy
& \multicolumn{1}{c}{$H(M \mid S)$}
& \multicolumn{1}{c}{Max width (mm)}
& \multicolumn{1}{c}{Mean width (mm)}
& \multicolumn{1}{c}{Invalid}
& \multicolumn{1}{c}{Success rate} \\
\midrule
BCAT           & 0.000 & \textemdash & \textemdash  & 20 & 0/20 (0.0) \\
ACT-Flow       & 1.000 & 40.0        & 19.6        &  1& 19/20(0.95) \\
KL-CVAE-LP        & 0.881 & 37.9        & 15.4        &  2&17/20(0.85) \\
MMD-CWAE-LP  & 0.982 & 33.1        & 13.3        &   3 & 13/20(0.65) \\
LAT-Flow       & 0.764 & 41.3        & 16.6        & 0 & 16/20(0.8)\\
Sinkhorn-CWAE-LP  & 0.000 & \textemdash & \textemdash &  2 & 17/20(0.85)\\
\bottomrule
\end{tabular}
\end{table}

These results suggest four observations:
\begin{enumerate}
    \item BCAT performs reasonably well across the simulation
    benchmarks, consistent with limited conditional action
    multimodality in their demonstration data. It also achieves
    higher average inference throughput than the flow-matching
    and CVAE policies evaluated here. The choice of a
    multimodal-capable policy should therefore be informed by
    whether the demonstration data actually require this
    representational capacity.

    \item The performance of multimodal-capable policies varies
    with training dynamics. Latent-variable policies involve
    many hyperparameters, resulting in a large search space
    for an effective configuration. Action-space flow or
    diffusion policies may therefore be preferable when
    reducing this tuning burden is a priority.

    \item Some models collapse late in training, including
    Act-Flow on UR3 BlockPush and KL-CVAE-LP on Kitchen Image.
    These observations highlight the importance of checkpoint
    selection and motivate further investigation of the role
    of covariate shift \cite{ross2011reduction, laskey2017dart} in performance degradation. 
    
    \item As emphasized in the manuscript, the bimodal
    tissue-grasping task illustrates the consequences of
    failing to represent certified multimodal demonstrations.
    BCAT collapses toward an average between modes, whereas
    Sinkhorn-CWAE-LP collapses to a single demonstrated mode.
    In this case, single-mode collapse retains a demonstrated
    strategy rather than interpolating into a region outside
    the expert support. Although Sinkhorn-CWAE-LP does not
    recover the full multimodal behavior, it still achieves
    an $85\%$ success rate.
\end{enumerate}
\begin{table}[h]
\centering
\scriptsize
\setlength{\tabcolsep}{3pt}
\caption{
Inference throughput in policy queries per second, measured as
the inverse wall-clock time required to generate one action chunk. All
methods are evaluated with the same hardware, batch size, and native inference setting used in the simulation
experiments. Higher is better.
}
\label{tab:throughput}

\resizebox{\linewidth}{!}{%
\begin{tabular}{lcccccc}
\toprule
Method
& PushT Img.
& PushT State
& Kitchen Img.
& Kitchen State
& UR3 BlockPush
& Avg. \\
\midrule
Act-Flow   & 6.24  & 6.59  & 4.61  & 5.79  & 6.24  & 5.89 \\
LAT-Flow   & 13.78 & 20.58 & 8.79  & 12.66 & 15.19 & 14.20 \\
KL-CVAE   & 30.63 & 69.02 & 17.50 & 21.62 & 19.73 & 31.70 \\
KL-CVAE-LP & 31.97 & 58.94 & 12.11 & 14.75 & 30.18 & 29.59 \\
MMD-CWAE-LP   & 29.89 & 59.10 & 13.31 & 27.05 & 15.72 & 29.02 \\
Sinkhorn-CWAE-LP    & 27.06 & 33.76 & 12.69 & 14.24 & 29.72 & 23.49 \\
BCAT       & 43.86 & 63.72 & 18.95 & 16.86 & 25.57 & 33.79 \\
\bottomrule
\end{tabular}%
}
\end{table}

\subsection{Relationship between policy performance and action horizon}
\label{app:action-horizon}
This ablation examines how policy performance varies with the training and execution action horizons. As noted in the manuscript, the
multimodality of the demonstration data is horizon-dependent.
We investigate how the choice of training horizon affects policy
performance and how this interacts with the execution horizon,
which determines the frequency of replanning.

The results in Table~\ref{tab:execution_horizon_ablation} show that longer open-loop execution amplifies failure but does not fully explain it. Performance generally deteriorates at longer execution horizons, particularly for long training chunks, consistent with the accumulation of open-loop errors. However, longer training chunks can also reduce performance at a fixed, short execution horizon, suggesting that modeling longer, higher-dimensional, and potentially more multimodal action distributions introduces additional difficulty. Training horizons of around $10$ actions and execution horizons of $2$--$5$ actions typically yield the best performance.
This suggests a double trade-off: during training, short chunks provide limited temporal consistency but are easier to model, while long chunks teach policies temporal coherence of the actions but need to learn a higher-dimensional conditional distribution. During execution, short horizons enable frequent replanning, improving reactivity and limiting the accumulation of deployment errors, but reduce sustained commitment to a global strategy. Longer execution horizons instead enforce commitment to a longer-term plan, at the cost of accumulating open-loop errors. Future work should investigate how policy modeling interacts with the execution horizon, and how these choices jointly affect covariate shift, multimodality representation, and generalization.

\begin{table}[h]
\centering
\caption{Execution-horizon ablation. For each benchmark, method, and training chunk, the best checkpoint is selected once from each training chunk horizon, then held fixed across all execution horizons. Superscripts give the selected training epoch.}
\label{tab:execution_horizon_ablation}
\setlength{\tabcolsep}{2.5pt}
\begin{tabular*}{\linewidth}{@{\extracolsep{\fill}}lccccc@{}}
\toprule
Benchmark
& \shortstack{Training\\horizon}
& \shortstack{Execution\\horizon}
& BCAT & KL-CVAE-LP & Act-Flow \\
\midrule
PushT State ($\cdot/1$) & 5 & 2 & $0.597^{(200)}$ & $0.743^{(500)}$ & $0.778^{(300)}$ \\
 &  & 5 & $0.724^{(200)}$ & $0.800^{(500)}$ & $0.816^{(300)}$ \\
 & 10 & 2 & $0.723^{(1000)}$ & $0.796^{(400)}$ & $0.825^{(1000)}$ \\
 &  & 5 & $0.734^{(1000)}$ & $0.827^{(400)}$ & $\mathbf{0.854}^{(1000)}$ \\
 &  & 10 & $0.611^{(1000)}$ & $\mathbf{0.829}^{(400)}$ & $0.813^{(1000)}$ \\
 & 30 & 2 & $0.643^{(1000)}$ & $0.727^{(800)}$ & $0.701^{(800)}$ \\
 &  & 5 & $\mathbf{0.749}^{(1000)}$ & $0.809^{(800)}$ & $0.808^{(800)}$ \\
 &  & 10 & $0.662^{(1000)}$ & $0.723^{(800)}$ & $0.774^{(800)}$ \\
 &  & 30 & $0.326^{(1000)}$ & $0.473^{(800)}$ & $0.615^{(800)}$ \\
\midrule
Kitchen State ($\cdot/4$) & 5 & 2 & $3.19^{(600)}$ & $3.04^{(300)}$ & $2.56^{(900)}$ \\
 &  & 5 & $3.02^{(600)}$ & $2.93^{(300)}$ & $2.23^{(900)}$ \\
 & 10 & 2 & $\mathbf{3.29}^{(500)}$ & $3.21^{(1000)}$ & $\mathbf{2.83}^{(200)}$ \\
 &  & 5 & $3.18^{(500)}$ & $\mathbf{3.27}^{(1000)}$ & $2.65^{(200)}$ \\
 &  & 10 & $2.94^{(500)}$ & $2.92^{(1000)}$ & $2.38^{(200)}$ \\
 & 30 & 2 & $3.00^{(1000)}$ & $2.74^{(900)}$ & $2.42^{(300)}$ \\
 &  & 5 & $2.96^{(1000)}$ & $2.84^{(900)}$ & $2.21^{(300)}$ \\
 &  & 10 & $2.83^{(1000)}$ & $2.56^{(900)}$ & $1.98^{(300)}$ \\
 &  & 30 & $1.03^{(1000)}$ & $1.12^{(900)}$ & $1.00^{(300)}$ \\
\midrule
UR3 BlockPush  ($\cdot/2$) & 5 & 2 & $1.94^{(700)}$ & $1.67^{(900)}$ & $\mathbf{1.68}^{(100)}$ \\
 &  & 5 & $1.93^{(700)}$ & $1.52^{(900)}$ & $1.57^{(100)}$ \\
 & 10 & 2 & $\mathbf{1.99}^{(1000)}$ & $1.65^{(1000)}$ & $1.48^{(200)}$ \\
 &  & 5 & $1.97^{(1000)}$ & $1.53^{(1000)}$ & $1.43^{(200)}$ \\
 &  & 10 & $1.97^{(1000)}$ & $1.47^{(1000)}$ & $1.45^{(200)}$ \\
 & 30 & 2 & $1.79^{(500)}$ & $\mathbf{1.72}^{(400)}$ & $1.35^{(100)}$ \\
 &  & 5 & $1.81^{(500)}$ & $1.69^{(400)}$ & $1.35^{(100)}$ \\
 &  & 10 & $1.87^{(500)}$ & $1.63^{(400)}$ & $1.24^{(100)}$ \\
 &  & 30 & $1.77^{(500)}$ & $1.44^{(400)}$ & $1.18^{(100)}$ \\
\bottomrule
\end{tabular*}
\end{table}

\clearpage

\newpage
\section{Experiment Hyperparameters}
\label{app:hyperparameters}
We report the hyperparameters used across synthetic and simulation
experiments. A dash denotes a setting that is not applicable or not run.
Unless stated otherwise, all models use AdamW with learning rate
\(5\times10^{-4}\), weight decay \(0.02\), observation horizon
\(\mathsf T_{\rm obs}=1\), and action horizon \(\mathsf H=10\) for simulation and robot tasks.
Synthetic tasks use \(\mathsf H=60\). Image encoders are fine-tuned jointly with
the policy. All transformer layers use ReLU as activation function.

\paragraph{Latent and generative objectives.}
For KL-CVAE, the pointwise KL weight is
\(\beta_{\rm KL}=0.01\). Learned-prior KL variants use a conditional
Gaussian prior with fixed unit variance. For VQ-VAE, the commitment
weight is \(\beta_{\rm vq}=0.25\) and the number of residual
quantization layers is \(\mathsf L_{\rm vq}=2\).

For aggregate-matched latent models, posterior geometry regularization
uses decoder-side latent jitter \(\sigma_{\rm dec}=0.075\), mean penalty
weight \(\beta_{\rm mean}=0.01\), standard-deviation penalty weight
\(\beta_{\rm std}=0.05\), covariance penalty weight
\(\beta_{\rm cov}=0.01\), target standard deviation \(\sigma_\star=1\),
and maximum standard deviation \(\sigma_{\max}=2\). Learned latent priors
are matched to posterior means.

For MMD-CWAE and its conditional variant, the latent aggregate matching
weight is \(\beta=1.0\). We use the inverse multiquadratic kernel
~\citep{tolstikhin2018wae}
\[
k_{\rm IMQ}(z,z')
=
\sum_{C\in\mathcal C_z}
\frac{C}{C+\|z-z'\|_2^2}.
\]
For synthetic experiments,
\(\mathcal C_z=\{0.5\mathsf d_z,\mathsf d_z,2\mathsf d_z,4\mathsf d_z\}\).
For simulation and robot experiments,
\(\mathcal C_z=\{\mathsf d_z,2\mathsf d_z,4\mathsf d_z,8\mathsf d_z\}\).

For Sinkhorn-CWAE and its conditional variant, the latent aggregate
matching weight is \(\beta=0.01\), the blur parameter is
\(\varepsilon_{\rm Sink}=0.1\), and the transport cost is squared
Euclidean distance.

For both conditional aggregate matching models
(Appendix~\ref{app:method-conditional-cwae}) used in the simulation
experiments, the state-conditioning weight is \(\lambda_s=1.0\).

For LAT-Flow, the latent flow-prior loss weight is \(\beta_{\rm flow}=0.01\).
LAT-Flow and Act-Flow both use an Euler solver with 10 inference steps
and sample training times from \(\operatorname{Beta}(1.5,1)\).

For Act-Diff, we use a DDIM scheduler with 100 training timesteps and
100 inference steps, the squared-cosine capped noise schedule, sample
prediction, sample clipping enabled, \(\alpha_0=1\), and step offset 0.

\paragraph{Dropout.}
Synthetic action decoders use dropout \(0.2\) and attention dropout
\(0.05\). Simulation and robot action decoders use dropout \(0.1\) and attention
dropout \(0.01\). Posterior encoders use dropout \(0.2\) and attention
dropout \(0.05\). Learned-prior encoders use dropout \(0.1\) and
attention dropout \(0.01\).

\paragraph{Data augmentation and normalization.}
We normalize observations and actions to \([-1,1]\) using min--max
scaling. For image-based simulation experiments, we apply data augmentation with Albumentations~\citep{buslaev2020albumentations}. The
photometric augmentation pipeline contains
ColorJitter with brightness \(0.3\), contrast \(0.4\), saturation
\(0.5\), hue \(0.1\), and probability \(0.5\);
RandomSunFlare with flare region \((0,0,1,0.5)\), white source color,
and probability \(0.6\); RandomBrightnessContrast with brightness and
contrast limits \(0.4\) and probability \(0.6\); RandomGamma with gamma
range \((80,120)\) and probability \(0.3\); CLAHE with clip limit \(4.0\)
and probability \(0.3\); RandomShadow with probability \(0.4\); and JPEG
ImageCompression with quality range \([50,100]\) and probability \(0.2\). The spatial and occlusion augmentation pipeline contains GaussianBlur
with blur limit \((3,7)\) and probability \(0.5\), CoarseDropout with at
most \(8\) holes of size \(8\times 8\) and probability \(0.3\), and
ShiftScaleRotate with shift limit \(0.0625\), scale limit
\([-0.5,0.6]\), zero rotation, and probability \(0.5\).

\label{app:hp}
\begin{table}[p]
\centering
\scriptsize
\setlength{\tabcolsep}{3pt}
\caption{Common benchmark-level training settings.}
\label{tab:common-training-hparams}

\begin{tabularx}{\linewidth}{@{}p{0.18\linewidth}*{7}{>{\centering\arraybackslash}X}@{}}
\toprule
Setting & Synthetic & PushT-S & PushT-I & Kitchen-S & Kitchen-I & UR3-S & MBG \\
\midrule
Observation Dimension & 3x64x64 & 24 & 3x96x96 & 30 & 3x112x112 & 6 & 4 \\
Observation horizon & 1 & 1 & 1 & 1 & 1 & 1 & 1 \\
Action horizon \(\mathsf H\) & 60 & 10 & 10 & 10 & 10 & 10 & 10 \\
Batch size & 256 & 256 & 64 & 256 & 64 & 256 & 256 \\
Validation Ratio & 0.05 & 0.05 & 0.05 & 0.05 & 0.05 & 0.05 & 0.05 \\
Epochs & 600 (Circle) / 1000 & 1000 & 1000 & 1000 & 1000 & 1000 & 1000 \\
Rollout every & 100 & 100 & 100 & 100 & 100 & 100 & -- \\
Optimizer & AdamW & AdamW & AdamW & AdamW & AdamW & AdamW & AdamW \\
Policy LR & \(5{\times}10^{-4}\) & \(5{\times}10^{-4}\) & \(5{\times}10^{-4}\) & \(5{\times}10^{-4}\) & \(5{\times}10^{-4}\) & \(5{\times}10^{-4}\) & \(5{\times}10^{-4}\) \\
Weight decay & 0.02 & 0.02 & 0.02 & 0.02 & 0.02 & 0.02 & 0.02 \\
Image-encoder LR & \(10^{-5}\) & -- & \(10^{-5}\) & -- & \(10^{-5}\) & -- & -- \\
Image-encoder WD & 0.001 & -- & 0.001 & -- & 0.001 & -- & -- \\
EMA & -- & true & true & true & true & true & true \\
\bottomrule
\end{tabularx}

\end{table}
  \begin{table}[t]
  \centering
  \small
  \setlength{\tabcolsep}{5pt}
  \caption{Common training settings for LIBERO and Meta-World.}
  \label{tab:libero-metaworld-training-hparams}
  \begin{tabularx}{\linewidth}{@{}l*{2}{>{\centering\arraybackslash}X}@{}}
  \toprule
  Setting & LIBERO & Meta-World \\
  \midrule
  Image observation dimension
  & \(2\times(3\times256\times256)\)
  & \(3\times480\times480\) \\
  Language encoder & all-MiniLM-L6-v2 & all-MiniLM-L6-v2 \\
  Language encoding dimension & \(31\times384\) & \(31\times384\) \\
  Language encoder frozen & true & true \\
  Observation horizon & 1 & 1 \\
  Action horizon \(\mathsf H\) & 10 & 10 \\
  Batch size & 64 & 64 \\
  Validation ratio & 0.05 & 0.05 \\
  Epochs & 100 & 100 \\
  Optimizer & AdamW & AdamW \\
  Policy LR & \(5\times10^{-5}\) & \(5\times10^{-5}\) \\
  Weight decay & 0.02 & 0.02 \\
  Image-encoder LR & \(10^{-5}\) & \(10^{-5}\) \\
  Image-encoder WD & 0.001 & 0.001 \\
  EMA & false & false \\
  \bottomrule
  \end{tabularx}
  \end{table}

\begin{table}[p]
\centering
\scriptsize
\setlength{\tabcolsep}{4pt}
\caption{Model sizes used in the experiments. \(L\) denotes transformer
layers, \(d\) hidden width, \(h\) attention heads, and FF the feedforward
width. All transformers use ReLU as activation function.}
\label{tab:model-sizes}
\begin{tabular}{@{}p{0.21\linewidth}p{0.34\linewidth}p{0.34\linewidth}@{}}
\toprule
Module / family & Synthetic & Simulation/Robot \\
\midrule
Image encoder
& MobileNetV4 Conv Small 0.50
& EfficientNet-B0 \\

State encoder
& --
& Linear projection (state\_dim, embedding\_dim) \\

BCAT decoder
& \(L=8,\ d=256,\ h=8,\ \mathrm{FF}=512\)
& \(L=8,\ d=256,\ h=8,\ \mathrm{FF}=1024\) \\

Latent decoder
& \(L=8,\ d=256,\ h=8,\ \mathrm{FF}=512\)
& \(L=8,\ d=256,\ h=8,\ \mathrm{FF}=1024\) \\
MMDiT decoder
& \(L=8,\ d=256,\ h=8,\ \mathrm{FF}=512,\ \text{timestep\_emb=32}\)
& \(L=8,\ d=256,\ h=8,\ \mathrm{FF}=1024,\ \text{timestep\_emb=128}\) \\

Posterior encoder
& \(L=8,\ d=256,\ h=8,\ \mathrm{FF}=512\)
& \(L=8,\ d=256,\ h=8,\ \mathrm{FF}=1024\) \\

Learned Gaussian prior
& --
& \(L=4,\ d=256,\ h=8,\ \mathrm{FF}=1024\) \\

Conditional aggregate prior
& --
& \(L=8,\ d=256,\ h=8,\ \mathrm{FF}=1024\) \\

Latent flow prior
& \(L=8,\ d=256,\ h=8,\ \mathrm{FF}=512\)
& \(L=8,\ d=256,\ h=8,\ \mathrm{FF}=1024\) \\

\bottomrule
\end{tabular}
\end{table}

\begin{table}[p]
\centering
\scriptsize
\setlength{\tabcolsep}{4pt}
\caption{Latent dimensions and codebook sizes.}
\label{tab:latent-dims-codebooks}
\begin{tabular}{@{}p{0.28\linewidth}p{0.20\linewidth}p{0.20\linewidth}p{0.20\linewidth}@{}}
\toprule
Benchmark / task & Modes \(\mathsf K\) & Latent dim. \(d_z\) & VQ codebook size \(\mathsf C\) \\
\midrule
Circle & 2 & 1 & 2 \\
Sequential & 4 & 2 & 4 \\
Radial & 16 & 4 & 16 \\
Corridor & 16 & 4 & 16 \\
Simulation benchmarks & -- & 16 & -- \\
\bottomrule
\end{tabular}
\end{table}

%% file: checklist.tex
\section*{NeurIPS Paper Checklist}

\begin{enumerate}

\item {\bf Claims}
    \item[] Question: Do the main claims made in the abstract and introduction accurately reflect the paper's contributions and scope?
    \item[] Answer: \answerYes{}
    \item[] Justification: The abstract and Section~\ref{sec:intro} state the paper's scope: an analysis of multimodality in behavioral cloning across latent-variable and action-space generative action-chunking policies. The theoretical claims are formalized in Section~\ref{sec:theory} and Appendix \ref{app:proofs}, and the empirical claims are evaluated in Section~\ref{sec:experiments} and additionally in Appendix~\ref{app:results}.

\item {\bf Limitations}
    \item[] Question: Does the paper discuss the limitations of the work performed by the authors?
    \item[] Answer: \answerYes{}
    \item[] Justification: The conclusion highlights the main limitations: the strongest mode-level diagnostics rely on controlled synthetic tasks with known mode labels and simulation benchmarks. It also identifies extensions to richer real-world mode annotations and broader policy classes as future work.

\item {\bf Theory assumptions and proofs}
    \item[] Question: For each theoretical result, does the paper provide the full set of assumptions and a complete (and correct) proof?
    \item[] Answer: \answerYes{}
    \item[] Justification: Section~\ref{sec:theory} states the assumptions and results for the latent-information bounds, aggregate matching, and action-space geometric cost. Full proofs and supporting notation and definitions are provided in Appendix~\ref{app:notation} and \ref{app:proofs}.

    \item {\bf Experimental result reproducibility}
    \item[] Question: Does the paper fully disclose all the information needed to reproduce the main experimental results of the paper to the extent that it affects the main claims and/or conclusions of the paper (regardless of whether the code and data are provided or not)?
    \item[] Answer: \answerYes{}
    \item[] Justification: Section~\ref{sec:experiments} describes the benchmark structure, evaluated policy families, and main metrics. Appendices ~\ref{app:exps}, ~\ref{app:results} and ~\ref{app:hyperparameters} provide additional architecture, training, diagnostic, and evaluation details. Exact bitwise reproducibility is not guaranteed because some GPU operations may be nondeterministic, but the reported protocols and aggregate metrics are reproducible in expectation.

\item {\bf Open access to data and code}
    \item[] Question: Does the paper provide open access to the data and code, with sufficient instructions to faithfully reproduce the main experimental results, as described in supplemental material?
    \item[] Answer: \answerYes{}
    \item[] Justification: We provide as supplementary material a downloadable ZIP archive for the code, synthetic data-generation
    scripts, simulation datasets, simulation benchmark evaluation servers, diagnostic scripts, README and reproduction instructions. The archive is anonymized and contains no author-identifying metadata. Upon acceptance, we will release the final public version.

\item {\bf Experimental setting/details}
    \item[] Question: Does the paper specify all the training and test details (e.g., data splits, hyperparameters, how they were chosen, type of optimizer) necessary to understand the results?
    \item[] Answer: \answerYes{}
    \item[] Justification: The experimental setup is summarized in Section~\ref{sec:experiments}. Full details on policy architectures, training settings, hyperparameters, datasets, inference settings, and evaluation metrics are provided in Appendix~\ref{app:exps} and ~\ref{app:hyperparameters}.

\item {\bf Experiment statistical significance}
    \item[] Question: Does the paper report error bars suitably and correctly defined or other appropriate information about the statistical significance of the experiments?
    \item[] Answer: \answerNo{}
    \item[] Justification: The paper reports averages over fixed evaluation protocols and, where applicable, multiple rollout samples, but does not provide confidence intervals or formal statistical significance tests for all experiments. Because the study compares several policy classes across multiple benchmarks and diagnostics, repeated multi-seed evaluation is computationally expensive. We therefore emphasize controlled diagnostics and report detailed per-task and aggregate results in Appendix~\ref{app:results}.

\item {\bf Experiments compute resources}
    \item[] Question: For each experiment, does the paper provide sufficient information on the computer resources (type of compute workers, memory, time of execution) needed to reproduce the experiments?
    \item[] Answer: \answerYes{}
    \item[] Justification: The paper provides accounting of GPU type, memory, and runtime for the reported experiment in Appendix \ref{app:training-details}.
    
\item {\bf Code of ethics}
    \item[] Question: Does the research conducted in the paper conform, in every respect, with the NeurIPS Code of Ethics \url{https://neurips.cc/public/EthicsGuidelines}?
    \item[] Answer: \answerYes{}
    \item[] Justification: The work is a methodological study of imitation-learning policy classes using synthetic and simulation benchmarks. We do not release high-risk models, collect sensitive data, or conduct human-subject experiments. We will release the code with the appropriate public licenses.

\item {\bf Broader impacts}
    \item[] Question: Does the paper discuss both potential positive societal impacts and negative societal impacts of the work performed?
    \item[] Answer: \answerNA{}
    \item[] Justification: The paper is primarily a foundational and methodological study of multimodal behavioral cloning. It discusses technical limitations, but does not include a dedicated broader-impact analysis of positive and negative societal consequences.
    
\item {\bf Safeguards}
    \item[] Question: Does the paper describe safeguards that have been put in place for responsible release of data or models that have a high risk for misuse (e.g., pre-trained language models, image generators, or scraped datasets)?
    \item[] Answer: \answerNA{}
    \item[] Justification: The paper does not release high-risk pretrained models, scraped datasets, language models, or image generators requiring misuse safeguards.

\item {\bf Licenses for existing assets}
    \item[] Question: Are the creators or original owners of assets (e.g., code, data, models), used in the paper, properly credited and are the license and terms of use explicitly mentioned and properly respected?
    \item[] Answer: \answerYes{}
    \item[] Justification: The paper cites the original works for existing benchmarks and policy families used as related works, including Push-T, Block Push, Kitchen, ACT, Diffusion Policy, VQ-BeT, and related methods. Existing assets are used only for research benchmarking and are credited in the corresponding sections.

\item {\bf New assets}
    \item[] Question: Are new assets introduced in the paper well documented and is the documentation provided alongside the assets?
    \item[] Answer: \answerYes{}
    \item[] Justification: The synthetic multimodal benchmark tasks introduced in the paper are specified in Section~\ref{sec:experiments} and Appendix~\ref{app:exps}, including task geometry, mode definitions, success criteria, and diagnostic metrics. The code to generate the data is released, as previously mentioned.

\item {\bf Crowdsourcing and research with human subjects}
    \item[] Question: For crowdsourcing experiments and research with human subjects, does the paper include the full text of instructions given to participants and screenshots, if applicable, as well as details about compensation (if any)? 
    \item[] Answer: \answerNA{}
    \item[] Justification: The paper does not involve crowdsourcing experiments or human-subject studies.

\item {\bf Institutional review board (IRB) approvals or equivalent for research with human subjects}
    \item[] Question: Does the paper describe potential risks incurred by study participants, whether such risks were disclosed to the subjects, and whether Institutional Review Board (IRB) approvals (or an equivalent approval/review based on the requirements of your country or institution) were obtained?
    \item[] Answer: \answerNA{}
    \item[] Justification: The paper does not involve human-subject research, so IRB approval or equivalent review is not applicable.

\item {\bf Declaration of LLM usage}
    \item[] Question: Does the paper describe the usage of LLMs if it is an important, original, or non-standard component of the core methods in this research? Note that if the LLM is used only for writing, editing, or formatting purposes and does \emph{not} impact the core methodology, scientific rigor, or originality of the research, declaration is not required.
    \item[] Answer: \answerNA{}
    \item[] Justification: LLMs are not part of the core methodology, experiments, theory, or scientific contribution of the paper.

\end{enumerate}